\documentclass[runningheads]{llncs}

\usepackage{eccv}

\usepackage{eccvabbrv}

\usepackage{graphicx}
\usepackage{booktabs}
\usepackage{tikz} 
\usepackage{graphicx}
\usepackage[accsupp]{axessibility}  %
\usepackage{comment}
\usepackage{tabularx}
\usepackage{bm}
\usepackage{caption} 
\usepackage{subcaption}
\usepackage{pifont} 
\usepackage{colortbl}
\usepackage{multirow}
\usepackage{makecell}
\usepackage{float}
\usepackage{enumitem} %
\usepackage{arydshln}
\usepackage[table,dvipsnames]{xcolor}
\usepackage{grffile}
\usepackage{siunitx} %
\usepackage[table,dvipsnames]{xcolor}

\usepackage{hyperref}

\usepackage{orcidlink}

\newcommand{\figref}[1]{Fig.~\ref{#1}}
\newcommand{\secref}[1]{Sec.~\ref{#1}}

\newcommand{\eqnref}[1]{Eq.~\eqref{#1}}
\newcommand{\tabref}[1]{Table~\ref{#1}}
\newcommand{\secrefn}[1]{Sec.\textcolor[rgb]{0,0.55,0.85}{~\ref*{#1}}}
\newcommand{\tabrefn}[1]{Table\textcolor[rgb]{0,0.55,0.85}{~\ref*{#1}}}
\newcommand{\figrefn}[1]{Fig.\textcolor[rgb]{0,0.55,0.85}{~\ref*{#1}}}
\newcommand{\eqnrefn}[1]{Eq.\textcolor[rgb]{0,0.55,0.85}{~\eqref{#1}}}

\makeatletter
\DeclareRobustCommand\onedot{\futurelet\@let@token\@onedot}
\def\@onedot{\ifx\@let@token.\else.\null\fi\xspace}
 
\def\ie{i.e\onedot}

\makeatother

\newcommand{\boldparagraph}[1]{\vspace{0.2em}\noindent{\bf #1.}}

\renewcommand{\paragraph}[1]{\boldparagraph{#1}}

\definecolor{darkgreen}{rgb}{0,0.7,0}
\definecolor{newyellow}{rgb}{1,0.8,0.05}
\definecolor{newgreen}{rgb}{0.2,0.8,0.2}

\definecolor{lightyellow}{RGB}{255,255,180} %

\colorlet{colorFst}{Green!25}       %
\colorlet{colorSnd}{SpringGreen!45} %
\colorlet{colorTrd}{Yellow!30}      %
\colorlet{colorLow}{darkgray!30}    %
\newcommand{\fs}{\cellcolor{colorFst}\bf}   %
\newcommand{\nd}{\cellcolor{colorSnd}}      %
\newcommand{\rd}{\cellcolor{colorTrd}}      %

\makeatletter
\def\adl@drawiv#1#2#3{%
        \hskip.5\tabcolsep
        \xleaders#3{#2.5\@tempdimb #1{1}#2.5\@tempdimb}%
                #2\z@ plus1fil minus1fil\relax
        \hskip.5\tabcolsep}
\newcommand{\cdashlinelr}[1]{%
  \noalign{\vskip\aboverulesep
           \global\let\@dashdrawstore\adl@draw
           \global\let\adl@draw\adl@drawiv}
  \cdashline{#1}
  \noalign{\global\let\adl@draw\@dashdrawstore
           \vskip\belowrulesep}}
\makeatother

\begin{document}

\title{VIGS-SLAM: \\ Visual Inertial Gaussian Splatting SLAM} 

\titlerunning{VIGS-SLAM}

\author{
Zihan Zhu\inst{1} \and
Wei Zhang\inst{2} \and
Moyang Li\inst{1} \and
Norbert Haala\inst{2} \and \\
Marc Pollefeys\inst{1,3} \and
Daniel Barath\inst{1}
}

\authorrunning{Zihan Zhu et al.}

\institute{
ETH Zurich, Zurich, Switzerland \and
University of Stuttgart, Stuttgart, Germany \and
Microsoft
}

\maketitle

\vspace{-1mm}
\begin{abstract}
We present VIGS-SLAM, a visual-inertial 3D Gaussian Splatting SLAM system that achieves robust real-time tracking and high-fidelity reconstruction. Although recent 3DGS-based SLAM methods achieve dense and photorealistic mapping, their purely visual design degrades under challenging conditions such as motion blur, low texture, and exposure variations. Our method tightly couples visual and inertial cues within a unified optimization framework, jointly optimizing camera poses, depths, and IMU states. It features robust IMU initialization, time-varying bias modeling, and loop closure with consistent Gaussian updates. Experiments on five challenging datasets demonstrate our superiority over state-of-the-art methods.
The code will be made public.
\keywords{SLAM \and IMU \and Gaussian Splatting}
\end{abstract}
    
\begin{figure}[ht]
\vspace{-12mm}
    \includegraphics[width=0.99\textwidth]{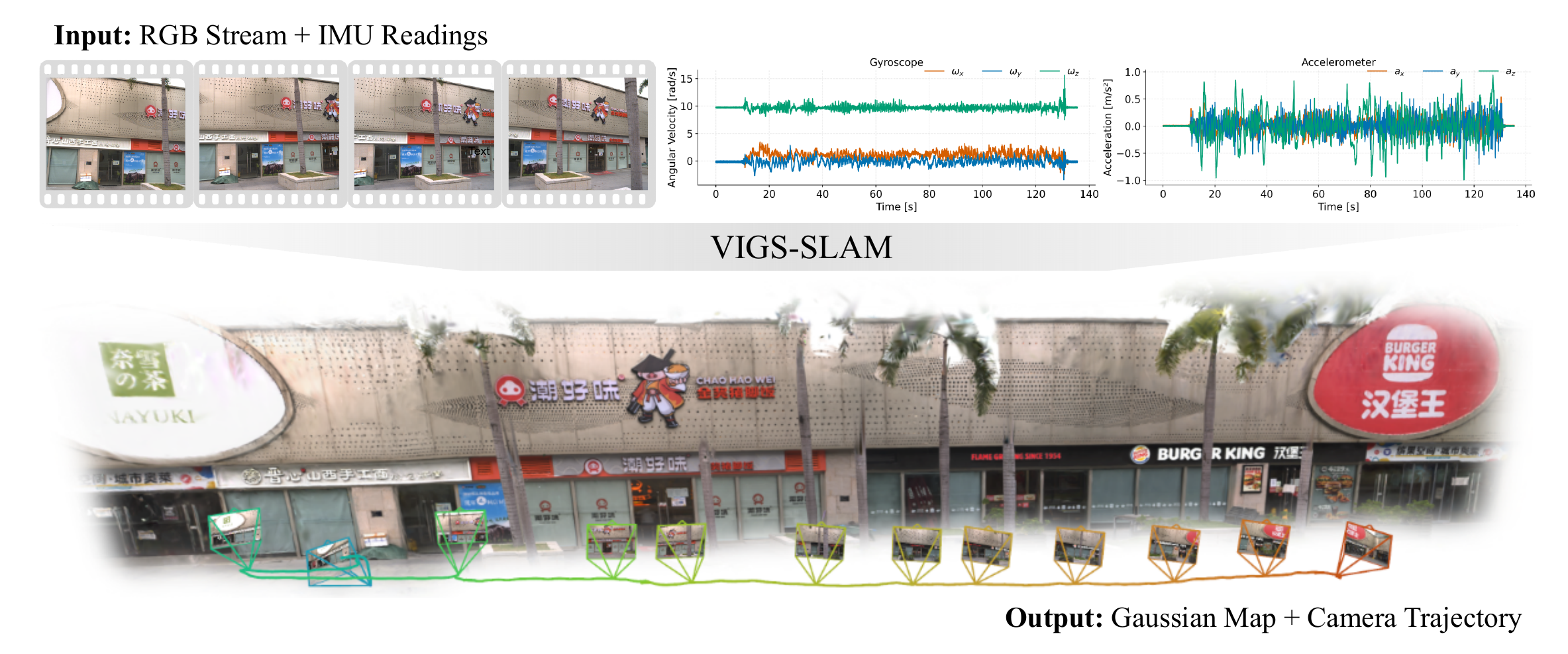}
    \vspace{-2mm}
    \captionof{figure}{
    \textbf{VIGS-SLAM.} Given a sequence of RGB frames and IMU readings, our method robustly tracks the camera trajectory while reconstructing a high-quality Gaussian map. Above is the visualization of \textit{Retail} sequence in FAST-LIVO2~\cite{zheng2024fast} dataset.
    }
\vspace{-12mm}
\label{fig:teaser}
\end{figure}
\section{Introduction}
\label{sec:intro}

Simultaneous Localization and Mapping (SLAM) is a key problem in robotics and computer vision, enabling autonomous navigation, augmented (AR) and mixed reality (VR) applications. Recent advances in neural implicit representations~\cite{Mildenhall2020ECCV} and 3D Gaussian Splatting (3DGS)~\cite{kerbl3Dgaussians} have transformed SLAM mapping from sparse point clouds to dense, photorealistic scene reconstructions. In particular, 3DGS SLAM methods~\cite{sandstrom2024splat,zhang2024hislam2} demonstrate that 3D Gaussian primitives provide a compact yet expressive scene representation, enabling real-time dense mapping and photorealistic novel view synthesis. 
However, most existing 3DGS-based SLAM systems remain purely visual: they rely on visual correspondences for tracking and often degrade under motion blur, textureless regions, low frame rate, or transient occlusions -- conditions common in real-world scenarios. 

Meanwhile, inertial measurement units (IMUs) have become ubiquitous -- integrated in virtually every modern smartphone~\cite{Google_Pixel_9_2024, Samsung_Galaxy_S24_2024}, AR headset~\cite{Meta_Quest_3_2023,Apple_Vision_Pro_2024}, and consumer camera~\cite{StereolabsZED2_2024, Insta360X3_2022,DJIOsmoAction_Gyroflow2024}. These sensors are low-cost (MEMS often under \$1 per chip) yet provide high-frequency measurements of acceleration and angular velocity, complementing vision by stabilizing tracking, recovering metric scale, and maintaining robustness under visually degraded conditions.

Traditional visual-inertial odometry (VIO) and SLAM such as OKVIS \cite{leutenegger2015okvis}, MSCKF \cite{mourikis2007msckf}, VINS-Mono~\cite{qin2018vins}, and ORB-SLAM3~\cite{campos2021orb3} are typically formulated as either filtering-based or optimization-based frameworks, that fuse visual observations with inertial measurements. Most systems rely on sparse geometric features such as BRISK~\cite{leutenegger2011brisk} or ORB~\cite{rublee2011orb}, while others adopt direct photometric alignment such as VI-DSO~\cite{von2018direct} to jointly optimize image intensities. Despite their accuracy, these representations produce only sparse or semi-dense point maps.
The recent state-of-the-art visual SLAM system DROID-SLAM~\cite{teed2021droid} reformulates dense correspondence estimation and bundle adjustment within a learned, differentiable framework, enabling iterative refinement of correspondences and joint optimization over camera poses and per-pixel disparities. This formulation is particularly well-suited for Gaussian initialization, as the 3D Gaussians can be directly initialized from the unprojected point cloud derived from the estimated disparities. However, DROID-SLAM~\cite{teed2021droid} and subsequent Gaussian-Splatting extensions~\cite{zhang2024hislam2, sandstrom2024splat} remain purely visual, without exploiting inertial measurements that could further enhance robustness.

The very few existing VIO 3DGS-SLAM approaches~\cite{sun2024mm3dgs, wu2025vings, ndoye2025vigs} exhibit several limitations, such as reliance on depth sensors, naive fusion of IMU data through pairwise (non-windowed) constraints, fixed IMU bias, or decoupled alternating optimization. Therefore, they suffer from reduced accuracy, limited robustness, and blurry renderings.
In contrast, we develop \textbf{VIGS-SLAM}, a highly robust SLAM system that achieves both \emph{real-time accurate tracking} and \emph{high-fidelity Gaussian reconstruction}. Our method tightly couples visual and inertial energy terms into a unified optimization framework, incorporates robust multi-stage IMU initialization and efficient loop closure with consistent Gaussian updates. This design improves tracking accuracy, reduces drift, enhances robustness under challenging conditions (e.g., motion blur, low texture, exposure variations, dynamic objects), and yields higher-fidelity renderings.
Extensive experiments on five diverse datasets, including a challenging self-captured visual-inertial dataset, demonstrate consistent improvements over state-of-the-art methods.
\section{Related Work}
\label{sec:related_work}
\paragraph{Visual-Inertial Odometry and SLAM} 
An early and seminal visual-inertial odometry approach is MSCKF~\cite{mourikis2007msckf}, which introduced a feature-marginalizing EKF that maintains a sliding window of cloned states, achieving real-time operation with a bounded state size.
Follow-ups~\cite{li2013high, huang2010observability} improved consistency, observability, and robustness, and inspired open implementations~\cite{bloesch2015robust, geneva2020openvins}. 

In parallel, optimization-based methods jointly minimize reprojection and preintegrated IMU residuals via a sliding-window nonlinear least-squares formulation. OKVIS~\cite{leutenegger2015okvis} tightly couples reprojection and IMU residuals in a keyframe bundle-adjustment backend. 
VINS-Mono~\cite{qin2018vins} adds robust initialization, loop closure, and relocalization. 
ICE-BA~\cite{liu2018ice} streamlines efficient incremental BA. 
ORB-SLAM3~\cite{campos2021orb3} unifies visual and visual-inertial modes with multi-map management and loop closing. 
Despite strong accuracy, these pipelines typically rely on sparse features or direct pixel intensities, which makes them brittle under low texture, motion blur, repetitive patterns, strong illumination/exposure changes. 
They also produce sparse (or semi-dense) maps that limit downstream tasks such as dense reconstruction, semantics, and photorealistic rendering.

DBA-Fusion~\cite{zhou2024dba} is the first to leverage dense correspondences from DROID-SLAM~\cite{teed2021droid} for visual-inertial SLAM. Our VIGS-SLAM also utilizes the same dense correspondences, but incorporates and optimizes visual-inertial constraints in a fundamentally different manner. 
DBA-Fusion~\cite{zhou2024dba} employs a cascaded two-stage framework: visual bundle adjustment is first solved independently, and the resulting Schur-complement Hessian is injected as a linearized factor into GTSAM~\cite{dellaert2012factor} inertial optimizer. This separation introduces a linearization gap that can only be partially mitigated through outer re-linearization iterations, resulting in limited robustness due to the lack of coherent visual-inertial fusion.
In contrast, VIGS-SLAM implements custom CUDA kernels for inertial bundle adjustment that directly fuse visual and inertial Hessian contributions within a single normal-equation solve per iteration. This formulation enables true tightly coupled joint optimization over poses, velocities, and biases, leading to improved numerical consistency and robustness. 

Furthermore, beyond tightly coupled optimization, our system integrates robust staged IMU initialization, efficient loop closure and Gaussian Splatting-based dense mapping, resulting in a unified visual-inertial Gaussian Splatting SLAM framework that improves tracking accuracy and robustness while enabling high-fidelity dense reconstruction and novel view synthesis.

\begin{figure*}[t]
\centering
 \includegraphics[width=\linewidth]{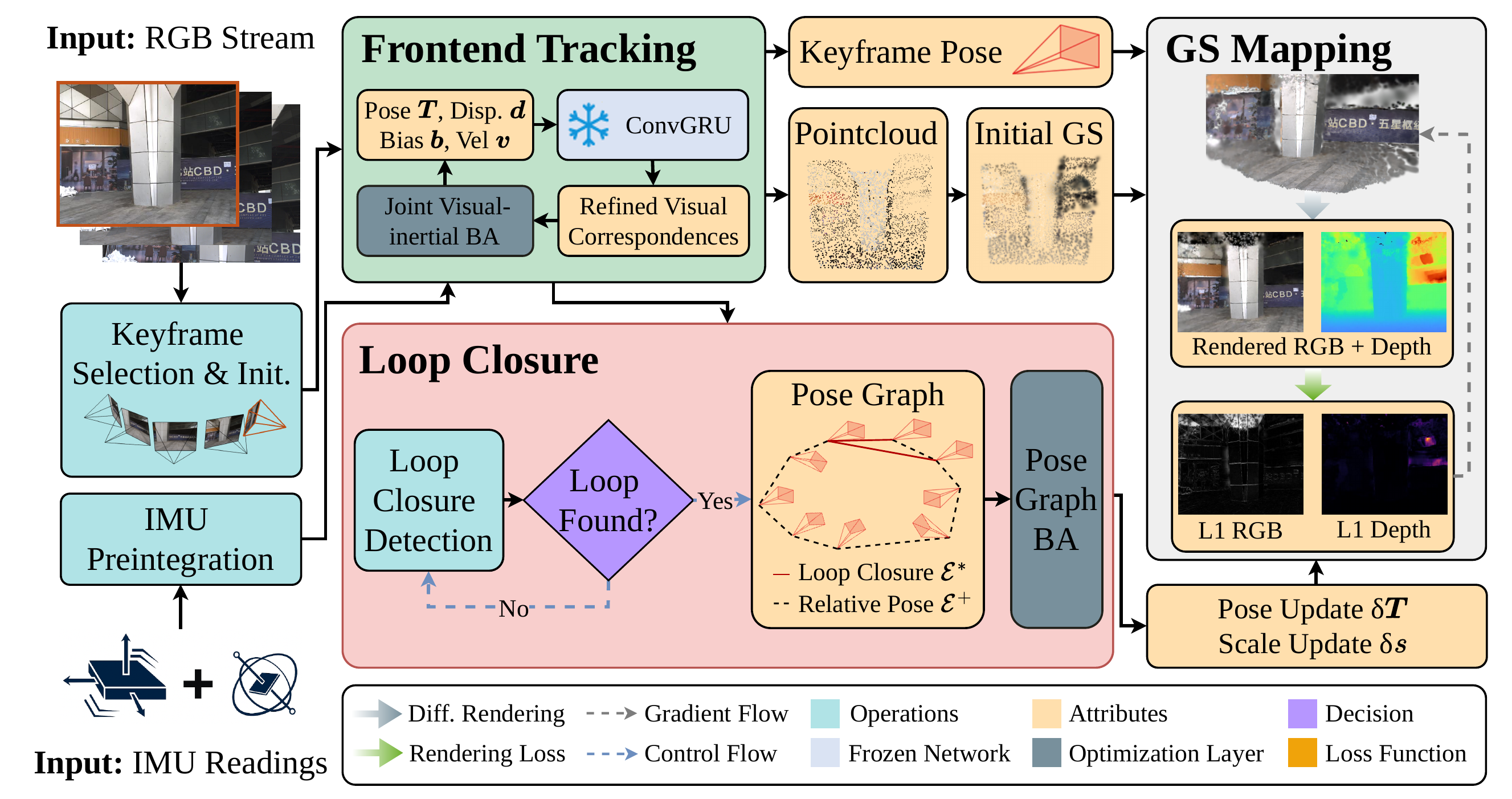}
\caption{\textbf{System Overview.} VIGS-SLAM takes a sequence of RGB frames and IMU readings as input, and simultaneously estimates camera poses while building a 3D Gaussian map~$\mathcal{G}$. Keyframes are selected based on optical flow, and each new keyframe is initialized using the IMU pre-integration from the previous keyframe. This keyframe is then added to the local frame graph, where visual-inertial bundle adjustment jointly optimizes camera poses, depths, and IMU parameters. Visual correspondences are iteratively refined using a recurrent ConvGRU module. In parallel, a global pose graph is maintained using relative pose constraints from the frontend tracking. For Gaussian mapping, the depth of each new keyframe is unprojected into 3D using the estimated pose, converted into initial Gaussians, and fused into the global map. Both color and depth re-rendering losses are used to refine the Gaussians. Loop closure detection is performed based on optical flow differences between the new keyframe and all previous ones. When a loop is detected, pose graph bundle adjustment is performed, followed by an efficient Gaussian update to maintain global consistency.}
\vspace{-11pt}
\label{fig:pipeline}
\end{figure*}

\paragraph{Neural Implicit and 3DGS SLAM}
Neural implicit representations and 3D Gaussian Splatting (3DGS) have gained significant attention in SLAM for accurate dense reconstruction and realistic novel view synthesis. Early systems such as iMAP~\cite{Sucar2021ICCV} and NICE-SLAM~\cite{Zhu2022CVPR} established unified pipelines that jointly perform mapping and tracking with neural fields. Subsequent work improves efficiency and scalability through compact encodings~\cite{Johari2022ESLAM, Kruzhkov2022MESLAM, wang2023co}, extends to monocular settings~\cite{zhu2024nicer, zhang2023hi, belos2022mod, chung2022orbeez, rosinol2023nerf}, and integrates semantics~\cite{zhu2024sni, li2023dns, zhai2024nis}.
3DGS~\cite{kerbl3Dgaussians} provides an explicit and efficient alternative, which sparked a wave of SLAM systems~\cite{matsuki2024gaussian, yan2024gs, huang2024photo, keetha2024splatam, li2025sgs, ha2024rgbd, hu2025cg, peng2024rtg, li2024gs3lam, zhu2024loopsplat, zheng2025wildgs, cao2025mcgs, deng2025gigaslam, zhang2024hislam2, sandstrom2024splat}. 
Among them, MonoGS~\cite{matsuki2024gaussian} is the first to achieve near real-time monocular SLAM using 3DGS as the sole map representation, while Splat-SLAM~\cite{sandstrom2024splat} and HI-SLAM2~\cite{zhang2024hislam2} further improve pose accuracy and map fidelity.

Integrating IMU sensor with monocular or RGB-D neural-implicit/3DGS SLAM remains underexplored. NeRF-VINS~\cite{katragadda2024nerf} and NeRF-VIO~\cite{zhang2025nerf} address map-based visual-inertial localization on a prior NeRF representation, rather than performing full SLAM. 
MM3DGS-SLAM~\cite{sun2024mm3dgs} and VIGS-Fusion~\cite{ndoye2025vigs} primarily target RGB-D+IMU settings. 
Only MM3DGS-SLAM supports RGB input by treating monocular depth as sensor depth. 
GI-SLAM~\cite{liu2025gi} reports stereo+IMU results on EuRoC~\cite{burri2016euroc} and a monocular+IMU variant on TUM~\cite{sturm2012benchmark} using accelerometer measurements only, without considering the gyroscope. VINGS-Mono~\cite{wu2025vings} adopts the interleaved tracking scheme of DBA-Fusion~\cite{zhou2024dba}, and thus suffers from similar robustness limitations and severe drift under challenging environments. For mapping, while it enables Gaussian reconstruction at kilometer-scale scenes, the resulting maps often miss fine details.

In contrast to prior work, we jointly optimize visual and inertial terms, incorporate robust IMU initialization, and perform efficient loop closure with consistent Gaussian map updates, resulting in improved accuracy and robustness.
\section{Method}
We provide an overview of our pipeline in \figref{fig:pipeline}. 
Our system takes as input a sequence of RGB frames $\{I_i\}_{i=1}^{N}$ and raw IMU measurements 
$\{\bm{a}_k, \bm{\omega}_k\}_{k=1}^{M}$, where $\bm{a}_k$ and $\bm{\omega}_k$ denote the angular velocity and linear acceleration. It simultaneously performs camera tracking and Gaussian Splatting Mapping. In detail, we first perform a staged IMU initialization to recover the metric scale, gravity direction, and IMU parameters. 
After initialization, incoming frames are processed by the frontend tracking module, which selects keyframes and forms a local frame graph $\mathcal{E}$. Camera motion as well as IMU parameters are then optimized within this graph by jointly minimizing both visual and inertial residuals.
Parallel to tracking, we incrementally build and refine a 3D Gaussian Splatting Map for high-quality rendering. Loop closures are detected and performed efficiently, with corresponding Gaussian updates to ensure global consistency.

\vspace{-3pt}
\subsection{Tracking}
The tracking module processes sequential images online, creating a new keyframe when the optical-flow magnitude to the last keyframe exceeds a threshold. To limit drift from IMU pre-integration over long time intervals, we also force a new keyframe at least every $t_{\text{kf}}$ seconds.
We formalize our tracking as an optimization problem where, for each keyframe $i$, we optimize camera pose $\bm{T}_i=(\bm{R}_i,\bm{p}_i)$, disparity $\bm{d}_i$, velocity $\bm{v}_i$, and IMU bias $\bm{b}_i=[\bm{b}_{g_i}^T,\bm{b}_{a_i}^T]^T$ (gyroscope $\bm{b}_{g_i}$ and accelerometer bias $\bm{b}_{a_i}$). In the following, we first introduce the visual and inertial residuals, followed by the optimization modules that build upon these residuals.

\paragraph{Vision Residual}
Following DROID-SLAM~\cite{teed2021droid}, we utilize a learned GRU-based update operator to predict and iteratively update the correspondence $\bm{u^{*}}_{ij}$ and its associated confidence map $\bm{w}_{ij}$ between a keyframe pair $(i, j)$.
We jointly optimize per-keyframe camera pose $\bm{T}$ and disparity $\bm{d}$ by minimizing a reprojection error between the refined correspondence $\bm{u^{*}}_{ij}$ and that computed from the current pose and disparity.
The vision residuals are formulated as:
{
\begin{equation}
\footnotesize
    \begin{aligned}
        \label{eq:vision}
        E_{\mathrm{vis}} (\bm{T}, \bm{d}) =
        \sum_{(i,j) \in \mathcal E} 
        \left\| 
            \bm{u^{*}}_{ij} - 
            \Pi(\bm{T}_{ij} \Pi^{-1}(\bm{u}_i, \bm{d}_i)) \right\|^2_{\bm{\Sigma}_{ij}}, 
    \end{aligned}
\end{equation}
}%
where $\bm{\Sigma}_{ij}  = \operatorname{diag}(\bm{w}^{\mathrm{vis}}_{ij})$.
Here, we loop over edges $(i, j)$ within a frame graph $\mathcal E$ connecting co-visible keyframes, and $\bm{T}_{ij}$ represents the relative transform between frames $i$ and $j$. We denote the projection and back-projection functions as $\Pi$ and $\Pi^{-1}$. Parameter $\bm{u}_i$ is the 2D image coordinate in frame $i$. Function $\|\cdot\|_{\bm{\Sigma}}$  denotes the Mahalanobis distance, weighting the residuals according to the confidence $\bm{w}^{\mathrm{vis}}_{ij}$.

\paragraph{Inertial Residual}
For each consecutive keyframe pair $(i,j)$, we first perform IMU pre-integration~\cite{forster2015imu} between timestamps $t_i$ and $t_j$ 
to efficiently fuse the high-frequency IMU data.
The pre-integration yields relative rotation $\Delta \bm{R}_{ij}$,
position $\Delta \bm{p}_{ij}$, and velocity $\Delta \bm{v}_{ij}$,
together with the corresponding Jacobians 
$(\bm{J}^{\mathrm{rot}}_{ij}, \bm{J}^{\mathrm{pos}}_{ij}, \bm{J}^{\mathrm{vel}}_{ij})$ and covariance $\bm{\Sigma}_{ij}^{\mathrm{iner}}$. 

In the inertial residual, we jointly optimize camera pose $\bm{T}_i$, velocity $\bm{v}_i$ and IMU bias $\bm{b}_i$ by minimizing the discrepancy between the current relative camera motion and the corresponding preintegrated IMU measurements. Additionally, we include a temporal bias smoothness term.
The inertial residuals are formulated as follows:
{
\begin{equation}
\footnotesize
\label{eq:inertial}
E_{\mathrm{iner}}(\bm{T}, \bm{v}, \bm{b}) = \sum_{(i, j=i+1) \in \mathcal{E}} 
\left\| 
\begin{bmatrix}
(\bm{r}^{\mathrm{rot}}_{i,j})^{\top},
(\bm{r}^{\mathrm{pos}}_{i,j})^{\top},
(\bm{r}^{\mathrm{vel}}_{i,j})^{\top},
(\bm{r}^{\mathrm{bias}}_{i,j})^{\top}
\end{bmatrix}^{\!\top}
\right\|^2_{\bm{\Sigma}_{ij}^{\mathrm{iner}}}. 
\end{equation}
}%
Here, the residual terms are defined as:
\begin{equation}
\footnotesize
\bm r_{ij} =
\begin{bmatrix}
\bm{r}^{\mathrm{rot}}_{i,j} \\[9pt]
\bm{r}^{\mathrm{pos}}_{i,j} \\[9pt]
\bm{r}^{\mathrm{vel}}_{i,j} \\[9pt]
\bm{r}^{\mathrm{bias}}_{i,j}
\end{bmatrix}
=
\begin{bmatrix}
\operatorname{Log} \!\left( 
(\Delta \bm{R}_{ij} 
\operatorname{Exp}(\bm{J}_{ij}^{\mathrm{rot}} 
(\bm{b}_{g_i} - \hat{\bm{b}}_{g_i})))^{\!\top} 
\bm{R}_i^{\top} \bm{R}_j 
\right)
\\[6pt]
\bm{R}_i^{\top} \!\left(
\bm{p}_j - \bm{p}_i - \bm{v}_i \Delta t_{ij}
- \tfrac{1}{2} \bm{g} \Delta t_{ij}^2
\right)
-
\left(
\Delta \bm{p}_{ij} +
\bm{J}_{ij}^{\mathrm{pos}}
(\bm{b}_i - \hat{\bm{b}}_i)
\right)
\\[6pt]
\bm{R}_i^{\top} \!\left(
\bm{v}_j - \bm{v}_i -
\bm{g} \Delta t_{ij}
\right)
-
\left(
\Delta \bm{v}_{ij} +
\bm{J}_{ij}^{\mathrm{vel}}
(\bm{b}_{g_i} - \hat{\bm{b}}_{g_i})
\right)
\\[6pt]
\bm{b}_{j} - \bm{b}_{i}
\end{bmatrix},
\end{equation}
where $\operatorname{Log}(\cdot)$ and $\operatorname{Exp}(\cdot)$ are the Lie algebra logarithm and exponential maps. Parameter  $\hat{\bm{b}}$ denotes the initial IMU bias and \(\bm{g} = \bm{R}_{\mathrm{wg}}\bm{g}_{\mathrm{I}}\) is the gravity direction in the world frame, where \(\bm{g}_{\mathrm{I}} = (0,0,G)^{\top}\) denotes the gravity in a gravity-aligned inertial frame, \(G\) denotes its magnitude, and \(\bm{R}_{\mathrm{wg}} \in \mathrm{SO}(3)\) is the rotation from the inertial frame to the world frame. For simplicity, we omit the transformation $T_{cb}$ between the camera and IMU coordinate frames in the equations; however, it is taken into account in the implementation.

\paragraph{Frontend Tracking: Local Bundle Adjustment}
We maintain a sliding-window local frame graph $\mathcal{E}$, following DROID-SLAM~\cite{teed2021droid}.
For pose initialization, we use IMU pre-integration to initialize each new keyframe.
To prevent unreliable inertial cues, we fall back to the previous-pose initialization whenever the pre-integration uncertainty is high, \ie, when the covariance trace exceeds a predefined threshold, $\mathrm{tr}(\bm{\Sigma}_{ij}^{\text{iner}})\!>\!\tau_{\Sigma}^{\text{init}}$. 
Using the estimated pose of the new keyframe, we initialize correspondences to earlier frames via geometric warping, providing a stronger starting point for the GRU-based update operator. 
We then perform joint visual-inertial optimization by minimizing the sum of the vision \eqnref{eq:vision} and inertial \eqnref{eq:inertial} residuals on the local frame graph $\mathcal E$.
The pose $\bm{T}$, velocity $\bm{v}$, bias $\bm{b}$ and disparity $\bm{d}$ are jointly optimized using the Levenberg-Marquardt algorithm through custom CUDA kernels as follows:
\begin{equation}
\footnotesize
\begin{aligned}
{\renewcommand{\arraystretch}{1.3}
\begin{bmatrix}
\bm{B} & \bm{E} \\
\bm{E}^{\mathsf{T}} & \bm{C}
\end{bmatrix}
\begin{bmatrix}
\Delta \boldsymbol{\xi} \\
\Delta \bm{d}
\end{bmatrix}
=
\begin{bmatrix}
\bm{w}_{\xi} \\
\bm{w}_{d}
\end{bmatrix}}
\end{aligned}
\qquad
\begin{aligned}
\Delta \boldsymbol{\xi} &=
[\bm{B} - \bm{E}\bm{C}^{-1}\bm{E}^{\mathsf{T}}]^{-1}
(\bm{w}_{\xi} - \bm{E}\bm{C}^{-1}\bm{w}_{d})\\
\Delta \bm{d} &=
\bm{C}^{-1}(\bm{w}_{d} - \bm{E}^{\mathsf{T}}\Delta \boldsymbol{\xi})
\end{aligned}
\label{eq:schur_update}
\end{equation}
where $\boldsymbol{\Delta \xi}$ represents the update of $[\bm{T}, \bm{v}, \bm{b}]$, and $\bm{\Delta d}$ denotes the depth update. Matrix $\bm{C}$ is diagonal as each term in \eqnref{eq:vision} depends only on a single depth value, thus it can be inverted by $\bm{C}^{-1} = 1 / \bm{C}$.

\paragraph{Loop Closure: Pose Graph Bundle Adjustment (PGBA)}
To handle loop closure efficiently, inspired by~\cite{zhang2024hislam2}, we adopt PGBA rather than a full global BA, trading a small amount of accuracy for substantial speed up. A parallel loop-detector builds loop edges $\mathcal{E}^{*}$ based on optical flow differences. We dynamically grow the relative pose graph $\mathcal E^{+}$ by adding the relative pose from local frame graph $\mathcal E$. We restrict heavy vision updates to only loop closure pairs $\mathcal{E}^{*}$ and add lightweight relative-pose constraints over the whole pose graph $\mathcal{E}^{+}$ as follows:
{
\begin{equation}
\footnotesize
\begin{aligned}
E_{\mathrm{PGBA}}({\bm T,\bm d}) = \sum_{(i,j)\in\mathcal E^{*}}
\left\|
\bm{u^{*}}_{ij}-
\Pi\!\big(\bm T_{ij}\,\Pi^{-1}(\bm u_i,\bm d_i)\big)
\right\|_{\bm\Sigma_{ij}}^{2} + \sum_{(i,j)\in\mathcal E^{+}}
\left\|
\log\!\big(\tilde{\bm T}_{ij}\,\bm T_i\,\bm T_j^{-1}\big)
\right\|_{\bm\Sigma^{\mathrm{rel}}_{ij}}^{2} \nonumber,
\label{eq:pgba}
\end{aligned}
\end{equation}
}%
where $\tilde{\bm T}_{ij}$ is the relative pose in the pose graph, and $\bm\Sigma^{\mathrm{rel}}_{ij}$ are relative-pose covariance from pairwise dense two-view correspondences as in~\cite{zhang2024hislam2}. The graph is optimized in $Sim(3)$ to correct long-term scale drift. 

\subsection{IMU Initialization}
At the beginning of each sequence, we initialize the IMU using a carefully designed three-stage procedure to ensure stable visual-inertial coupling.

\paragraph{Stage 1: Pure Vision Initialization}
Using the first $N^{\mathrm{vis}}_{\mathrm{init}}$ keyframes, we minimize \eqnref{eq:vision} to estimate poses ${\bm{T}_i}$ and disparities ${\bm{d}_i}$ up to a single global scale.

\paragraph{Stage 2: Inertial-Only Optimization}
We continue visual tracking until $N^{\mathrm{iner}}_{\mathrm{init}}$ keyframes are available. We solve the inertial objective \eqnref{eq:inertial} restricted to only the $\bm{R}_{\mathrm{wg}}$, thereby aligning the gravity direction.
Keeping ${\bm{T}_i}$ fixed, we augment the optimization variables with per-keyframe velocities $\bm{v}_i$, and IMU bias $\bm{b}_i$ together with a global log-scale parameter to recover metric scale.
Details are in the supplementary material.

\paragraph{Stage 3: Visual-Inertial Optimization} 
We further refine the estimates by jointly minimizing the visual and inertial objectives in Eqs.~\ref{eq:vision} and \ref{eq:inertial}, respectively.

\vspace{1mm}
\noindent Our staged initialization enhances robustness by postponing visual-inertial coupling until the IMU parameters can be reliably estimated.

\subsection{Gaussian Splatting Mapping}

\paragraph{Preliminary} We utilize the 3D Gaussian representation \cite{kerbl3Dgaussians} to model scene appearance and geometry. The scene is represented by a set of anisotropic Gaussians $\mathcal{G}=\{ g_{i}\}^{K}_{i=1}$. Each Gaussian $g_{i}$ contains color $\bm{c}_{i} \in \mathbb{R}^{3}$, opacity $o_{i} \in [0,1]$, mean $\boldsymbol{\mu}_{i} \in \mathbb{R}^{3}$, and covariance matrix $\boldsymbol{\Sigma}_{i} \in \mathbb{R}^{3 \times 3}$. 
The color of each pixel in the rendered image is calculated by alpha-blending the visible Gaussians.
Following prior work~\cite{matsuki2024gaussian,yugay2023gaussian}, we replace spherical-harmonic colors with direct RGB value, reducing optimization complexity.

\paragraph{Map Management}
We initialize the Gaussian map by unprojecting the new keyframe’s depth $\bm{D}$ (converted from disparity $\bm{d}$) into 3D, setting each Gaussian’s color $\bm{c}$ to the corresponding pixel color and its opacity $o$ to $0.5$. For each new keyframe, we run 10 mapping iterations. In each iteration, we randomly sample keyframes from the frontend tracking frame graph $\mathcal{E}$ along with two global keyframes to render color $\hat{I}$ and depth $\hat{D}$ from the Gaussian map. We calculate color loss $\mathcal{L}_c = \lVert \hat{I} - I \rVert_{1}$ and depth loss $\mathcal{L}_d = \lVert \hat{D} - D \rVert_{1}$, as well as isotropic regularization loss $\mathcal{L}_{iso}$~\cite{matsuki2024gaussian} to prevent
excessive elongation in sparsely observed regions. The optimization minimizes the following weighted total loss:
{
\begin{equation}
\label{eq:render}
\mathcal{L} = \lambda_d \mathcal{L}_d + \lambda_c \mathcal{L}_c + \lambda_{iso} \mathcal{L}_{iso} \, .
\end{equation}
}

\paragraph{Loop Closure Gaussian Update}
After loop closure, the poses and scales of all keyframes in the relative pose graph are updated. To keep the map consistent without reinitializing and reoptimizing all Gaussians, we propagate each keyframe’s update to the Gaussians anchored to it. 
For keyframe $k$, let pre-/post-PGBA poses be $(\bm R_k^{-}, \bm p_k^{-})$ and $(\bm R_k^{+}, \bm p_k^{+})$, and let the scale change be $\delta s_k$. For any Gaussian $g_{i}$ initialized from keyframe $k$ with mean $\mu_i^{-}$ and covariance $\boldsymbol{\Sigma}_i^{-}$, we update by first transforming into the 
old camera frame, applying the scale, then mapping to the new world frame as follows:
{
\begin{equation}
\footnotesize
    \begin{aligned}
    \bm x_{\text{loc}}^{-} = (\bm R_k^{-})^{\top}\big(\boldsymbol{\mu}_i^{-}-\bm t_k^{-}\big), \qquad
    \bm {x}_{\text{loc}}^{+} = \delta s_k\bm x_{\text{loc}}^{-}, \qquad
    \boldsymbol{\mu}_i^{+} = \bm R_k^{+}\bm x_{\text{loc}}^{+}+\bm t_k^{+}.
    \end{aligned}
\end{equation}
}%
Covariances are updated analogously:
{
\begin{equation}
\footnotesize
\begin{aligned}
\boldsymbol{\Sigma}_i^{+}
= \bm R_k^{+}\left(\delta s_k^{2}(\bm R_k^{-})^{\top}\boldsymbol{\Sigma}_i^{-}\bm R_k^{-}\right)(\bm R_k^{+})^{\top},
\end{aligned}
\end{equation}
}%
while $o_{i}$ and $\bm{c}_{i}$ remain unchanged. This entire update is applied in batch operations for high efficiency.

\section{Experiments}\label{sec:experiments}

\paragraph{Datasets} 
We evaluate on the EuRoC~\cite{burri2016euroc}, RPNG AR Table~\cite{Chen2023rpng}, UTMM~\cite{sun2024mm3dgs}, FAST-LIVO2~\cite{zheng2024fast} datasets, as well as a self-captured dataset.
The EuRoC dataset provides grayscale images, while the others offer RGB ones. %
For FAST-LIVO2~\cite{zheng2024fast} dataset, we use the poses from FAST-LIVO2~\cite{zheng2024fast} as ground truth since it leverages LiDAR measurements. For our self-captured dataset, we use the Manifold Odin 1~\cite{manifoldtech2025odin1}, whose offline processing tool MindCloud~\cite{manifold_mindcloud} provides ground-truth poses via LiDAR-visual-inertial fusion.
The remaining datasets provide motion-capture ground truth. 
Additional details are provided in the supp.\ material.

\begin{table*}[!t]
\centering
\setlength{\tabcolsep}{3.0pt}
\caption{\textbf{Tracking Performance on EuRoC Dataset~\cite{burri2016euroc}} (ATE RMSE $\downarrow$ [cm]). Best results are highlighted as \colorbox{colorFst}{\bf first},\colorbox{colorSnd}{second}, and\colorbox{colorTrd}{third}. `F' indicates failure. Results for  SVO~\cite{forster2014svo}, TartanVO~\cite{wang2021tartanvo}, DSO~\cite{Engel2017DSO}, MSCKF~\cite{mourikis2007msckf}, OKVIS~\cite{leutenegger2015okvis}, VINS-Mono~\cite{qin2018vins}, and ORB-SLAM3~\cite{campos2021orb3} are as reported by the ORB-SLAM3 paper~\cite{campos2021orb3}; DROID-SLAM~\cite{teed2021droid} numbers are from its paper. All other results are reproduced from their official code.}
\resizebox{\textwidth}{!}{
\begin{tabular}{
    lrrrrrrrrrrr
    !{\smash{\tikz[baseline]{\draw[densely dashed, gray!80, line width=0.8pt] (0pt,-2pt)--(0pt,8pt);}}}
    r
}
\toprule
Method & \tt{MH\_01} & \tt{MH\_02} & \tt{MH\_03} & \tt{MH\_04} & \tt{MH\_05} & \tt{V1\_01} & \tt{V1\_02} & \tt{V1\_03} & \tt{V2\_01} & \tt{V2\_02} & \tt{V2\_03} & \textbf{Avg.} \\
\midrule
\multicolumn{13}{l}{\cellcolor[HTML]{EEEEEE}{\textit{RGB}}} \\
SVO~\cite{forster2014svo}            & 10.00  & 12.00  & 41.00  & 43.00  & 30.00  & 7.00   & 21.00  & \multicolumn{1}{r}{F} & 11.00  & 11.00  & 108.00 & \multicolumn{1}{r}{N/A} \\
Splat-SLAM~\cite{sandstrom2024splat} & 257.64 & 266.02 & 312.58 & 458.14 & 360.86 & 168.99 & 166.65 & 128.67 & 198.84 & 195.85 & 190.86 & 245.01 \\
TTT3R~\cite{chen2025ttt3r}           & 421.47 & 385.94 & 293.01 & 414.59 & 381.05 & 155.43 & 128.70 & 118.06 & 141.96 & 93.56  & 101.81 & 239.60 \\
TartanVO~\cite{wang2021tartanvo}     & 63.90  & 32.50  & 55.00  & 115.30 & 102.10 & 44.70  & 38.90  & 62.20  & 43.30  & 74.90  & 115.20 & 68.00 \\
DSO~\cite{Engel2017DSO}              & 4.60   & 4.60   & 17.20  & 381.00 & 11.00  & 8.90   & 10.70  & 90.30  & 4.40   & 13.20  & 115.20 & 60.10 \\
DROID-SLAM~\cite{teed2021droid}      & 16.30  & 12.10  & 24.20  & 39.90  & 27.00  & 10.30  & 16.50  & 15.80  & 10.20  & 11.50  & 20.40  & 18.60 \\
HI-SLAM2~\cite{zhang2024hislam2}     & \nd2.66 & \nd1.44 & \nd2.71 & \nd6.86 & \fs5.07 & \fs3.55 & \nd1.32 & \fs2.49 & \nd2.56 & \nd1.77 & \fs1.92 & \nd2.94 \\
\hdashline
\noalign{\vskip 1pt}
\multicolumn{13}{l}{\cellcolor[HTML]{EEEEEE}{\textit{RGB+IMU}}} \\
MSCKF~\cite{mourikis2007msckf}       & 42.00  & 45.00  & 23.00  & 37.00  & 48.00  & 34.00  & 20.00  & 67.00  & 10.00  & 16.00  & 113.00 & 41.40 \\
VINGS-Mono~\cite{wu2025vings} & 21.03 & 16.47 & 25.46 & 25.03 & 36.01 & 6.54 & 9.79 & 11.46 & 11.51 & 93.44 & 12.39 & 24.47 \\
OKVIS~\cite{leutenegger2015okvis}    & 16.00  & 22.00  & 24.00  & 34.00  & 47.00  & 9.00   & 20.00  & 24.00  & 13.00  & 16.00  & 29.00  & 23.10 \\
DBA-Fusion~\cite{zhou2024dba} & 17.88 & 16.72 & 24.03 & 23.57 & 27.81 & 20.11 & 9.51 & 8.89 & 10.37 & 11.51 & 16.22 & 16.97 \\
OPEN-VINS~\cite{geneva2020openvins}  & 50.61  & 5.61   & 7.16   & 6.34   & 12.78  & 17.46  & 9.48   & 5.96   & 11.68  & 10.47  & 14.83  & 13.85 \\
VINS-Mono~\cite{qin2018vins}         & 8.40   & 10.50  & 7.40   & 12.20  & 14.70  & 4.70   & 6.60   & 18.00  & 5.60   & 9.00   & 24.40  & 11.00 \\
ORB-SLAM3~\cite{campos2021orb3}      & \rd6.20 & \rd3.70 & \rd4.60 & \rd7.50 & \rd5.70 & \rd4.90 & \rd1.50 & \rd3.70 & \rd4.20 & \rd2.10 & \nd2.70 & \rd4.30 \\
\textbf{VIGS-SLAM (Ours)}                         & \fs1.42 & \fs1.29 & \fs2.55 & \fs5.16 & \nd5.64 & \nd3.67 & \fs1.15 & \nd2.68 & \fs2.34 & \fs1.53 & \rd3.27 & \fs2.79 \\
\bottomrule
\end{tabular}
}

\label{tab:tracking_euroc}
\end{table*}

\begin{table*}[t]
\centering
\footnotesize
\setlength{\tabcolsep}{5.5pt}
\caption{\textbf{Tracking Performance on RPNG AR Table Dataset~\cite{Chen2023rpng}} (ATE RMSE $\downarrow$ [cm]). Best results are highlighted as \colorbox{colorFst}{\bf first},\colorbox{colorSnd}{second}, and\colorbox{colorTrd}{third}. `F' indicates failure.}
\resizebox{\textwidth}{!}{
\begin{tabular}{ l rrrrrrrr
!{\smash{\tikz[baseline]{\draw[densely dashed, gray!80, line width=0.8pt] (0pt,-2pt)--(0pt,8pt);}}}
r}
\toprule
Method & \tt{table\_01} & \tt{table\_02} & \tt{table\_03} & \tt{table\_04} & \tt{table\_05} & \tt{table\_06} & \tt{table\_07} & \tt{table\_08} & {\textbf{Avg.}} \\
\midrule
\multicolumn{10}{l}{\cellcolor[HTML]{EEEEEE}{\textit{RGB}}} \\
HI-SLAM2~\cite{zhang2024hislam2} & \nd1.43 & \nd1.66 & \nd1.23 & 2.59 & \multicolumn{1}{r}{F} & \nd1.47 & \fs0.97 & \fs2.67 & {N/A} \\
Splat-SLAM~\cite{sandstrom2024splat} & 22.46 & 41.50 & 37.93 & \fs 1.09 & \fs 1.19 & 1.53 & 1.76 & 4.37 & 13.98 \\
DROID-SLAM~\cite{teed2021droid} & 9.59 & 6.58 & 7.28 & 11.82 & 6.24 & 4.22 & 4.27 & 46.59 & 12.07 \\
\hdashline
\noalign{\vskip 1pt}
\multicolumn{10}{l}{\cellcolor[HTML]{EEEEEE}{\textit{RGB+IMU}}} \\
OKVIS~\cite{leutenegger2015okvis} & 9.00 & 7.70 & 15.30 & 16.20 & 24.50 & 10.20 & 13.60 & 19.80 & 14.60 \\
VINGS-Mono~\cite{wu2025vings} & 4.57 & 2.90 & 4.04 & 6.51 & \rd 1.90 & 5.16 & 5.08 & 23.05 & 6.65 \\
DBA-Fusion~\cite{zhou2024dba} & 4.73 & \rd 2.77 & 4.85 & 6.64 & 2.27 & 4.28 & 5.64 & 20.71 & 6.49 \\
OPEN-VINS~\cite{geneva2020openvins} & 4.29 & 3.03 & 3.11 & 6.20 & 3.85 & 4.45 & 6.58 & 9.20 & 5.09 \\
ORB-SLAM3~\cite{campos2021orb3} & 2.52 & 15.79 & \rd 1.57 & \nd 1.22 & 7.34 & \rd 1.49 & \rd 1.24 & \nd 3.43 & \nd 4.33 \\
VINS-Mono~\cite{qin2018vins} & \rd 2.72 & 5.98 & 3.30 & 4.01 &  2.18 &  1.87 & 2.05 & 5.54 & \rd 3.46 \\
\textbf{VIGS-SLAM (Ours)} & \fs 1.31 & \fs 1.57 & \fs 1.22 & \rd 1.75 & \nd 1.28 & \fs 1.38 & \nd 1.08 & \rd 3.86 & \fs 1.68 \\
\bottomrule
\end{tabular}
}

\label{tab:tracking_rpng}
\vspace{-4mm}
\end{table*}

\begin{figure*}[!b]
\centering
\scriptsize
\setlength{\tabcolsep}{1pt}
\renewcommand{\arraystretch}{0.5}

\newcommand{\sz}{0.094} %
\setlength{\arrayrulewidth}{0.8pt}
\begin{tabular}{lcccccc}
\raisebox{-0.3\normalbaselineskip}[0pt][0pt]{\rotatebox[origin=c]{90}{VINGS-Mono~\cite{wu2025vings}}}  &
\includegraphics[height=\sz\linewidth]{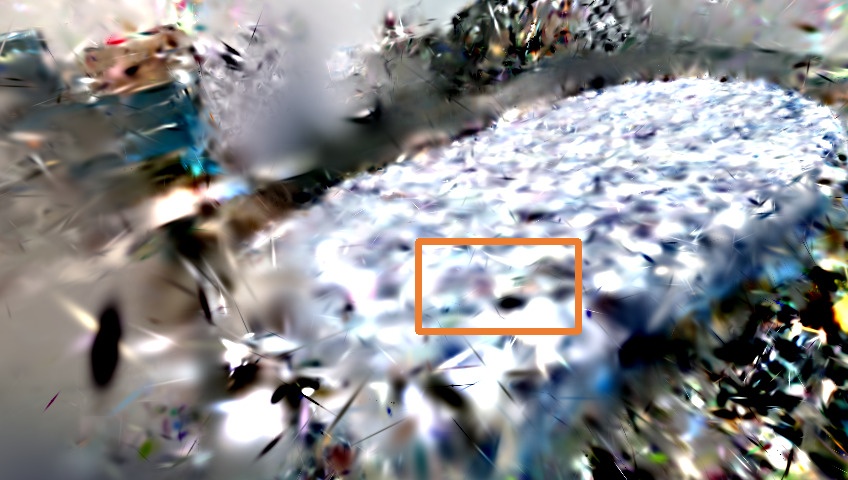} &
\includegraphics[height=\sz\linewidth]{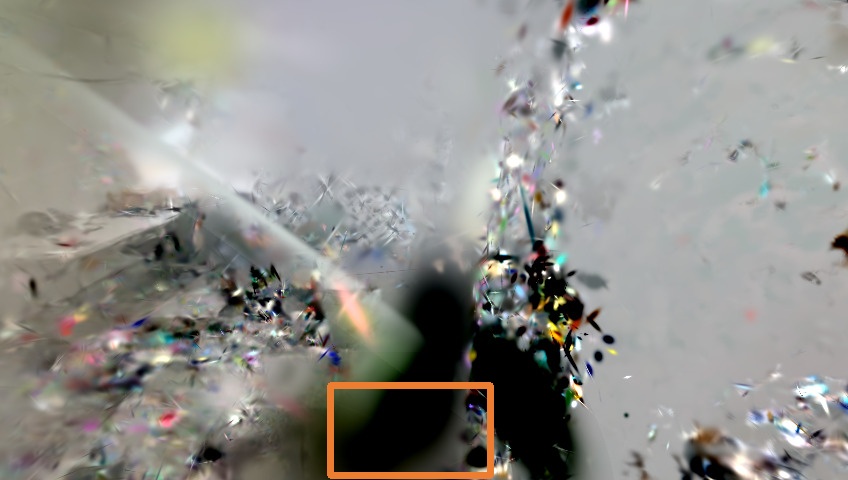} &
\includegraphics[height=\sz\linewidth]{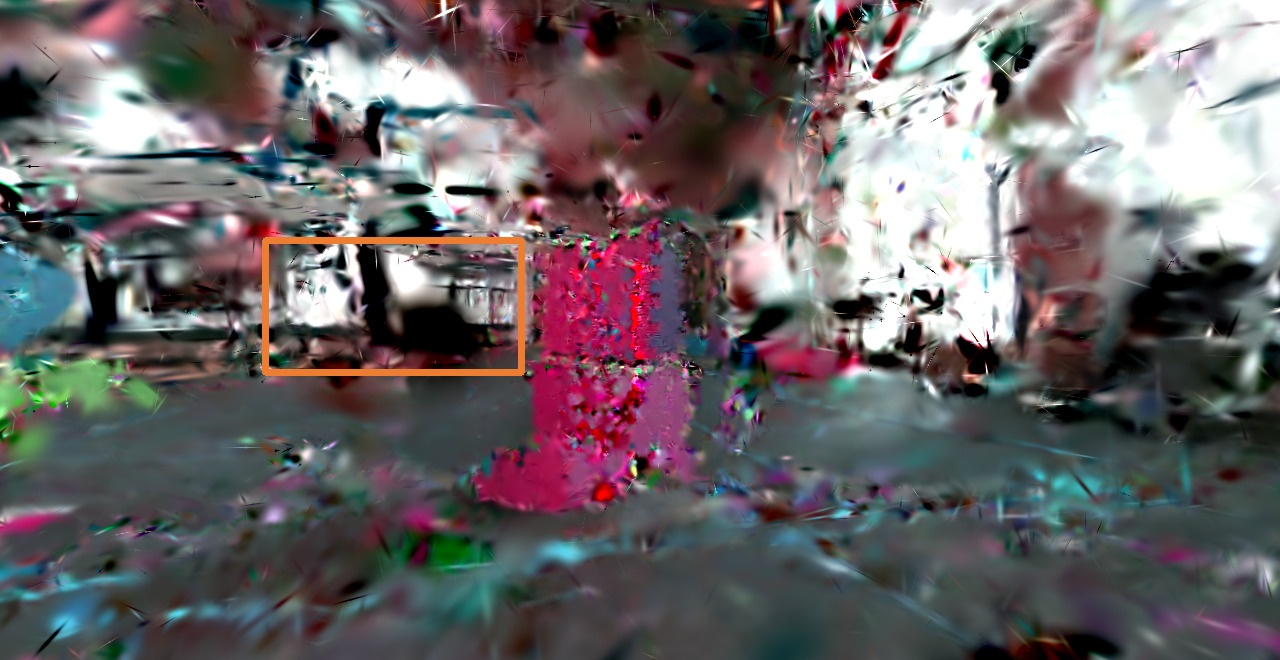} &
\includegraphics[height=\sz\linewidth]{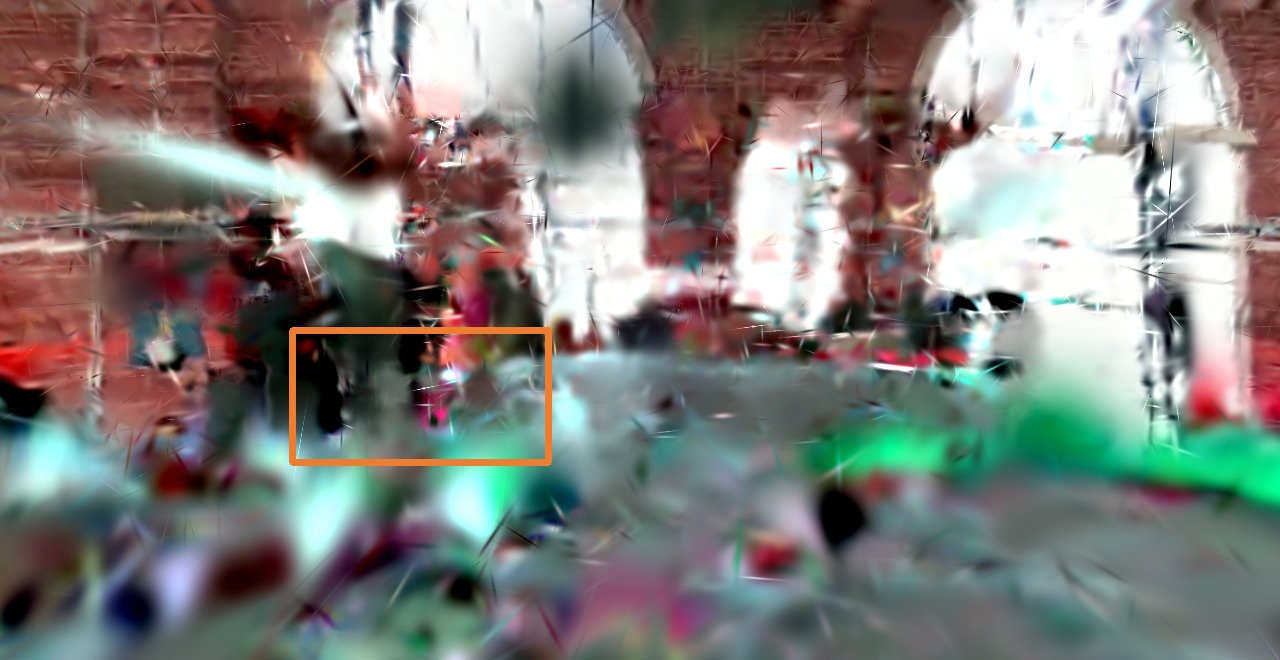} &
\includegraphics[height=\sz\linewidth]{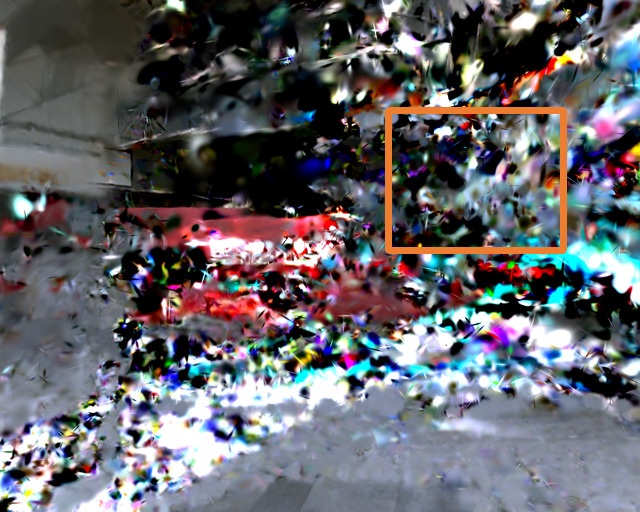} &
\includegraphics[height=\sz\linewidth]{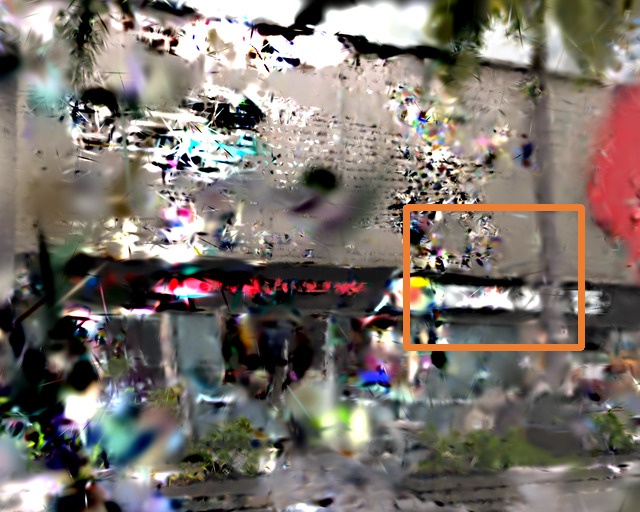} \\
&
\includegraphics[height=\sz\linewidth]{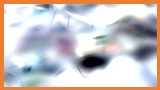} &
\includegraphics[height=\sz\linewidth]{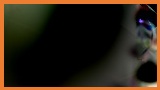} &
\includegraphics[height=\sz\linewidth]{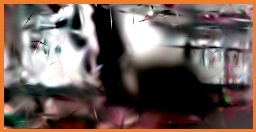} &
\includegraphics[height=\sz\linewidth]{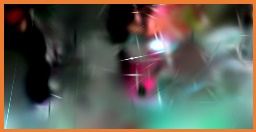} &
\includegraphics[height=\sz\linewidth]{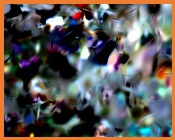} &
\includegraphics[height=\sz\linewidth]{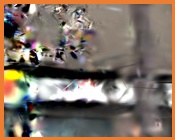} \\
\noalign{\vskip 0.5pt}
\hdashline
\noalign{\vskip 1.5pt}

\raisebox{-0.1\height}[0pt][0pt]{\rotatebox[origin=c]{90}{HI-SLAM2~\cite{zhang2024hislam2}}}  &
\includegraphics[height=\sz\linewidth]{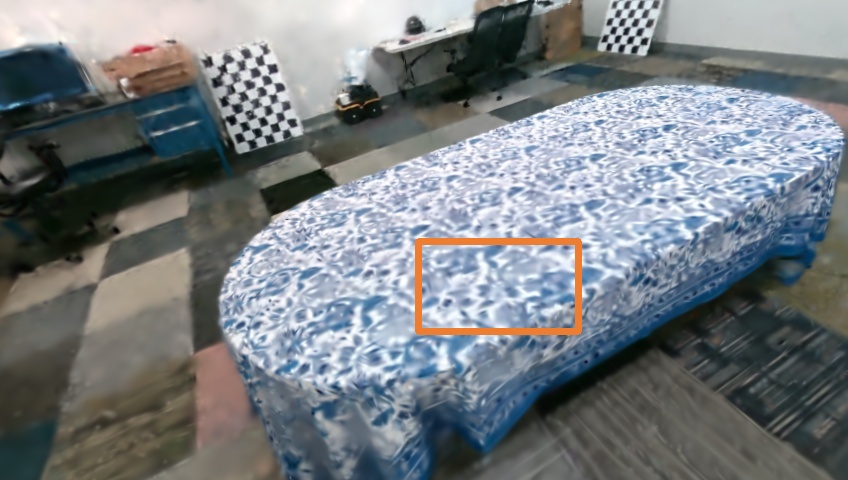} &
\includegraphics[height=\sz\linewidth]{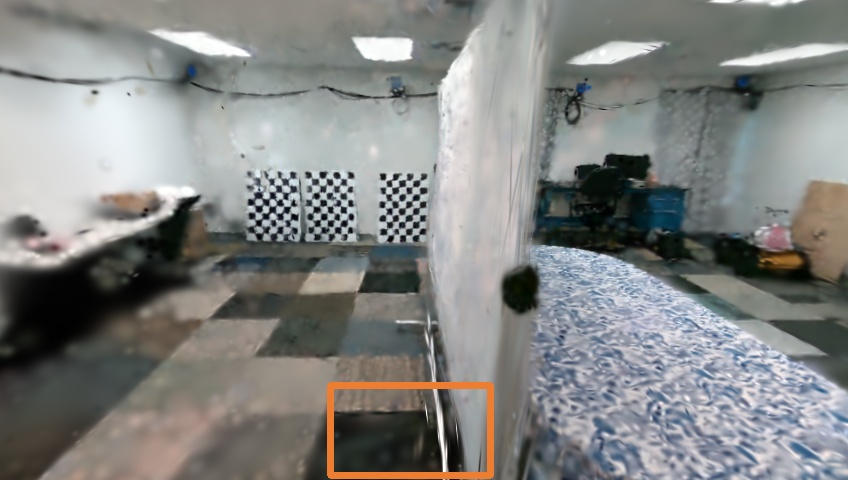} &
\includegraphics[height=\sz\linewidth]{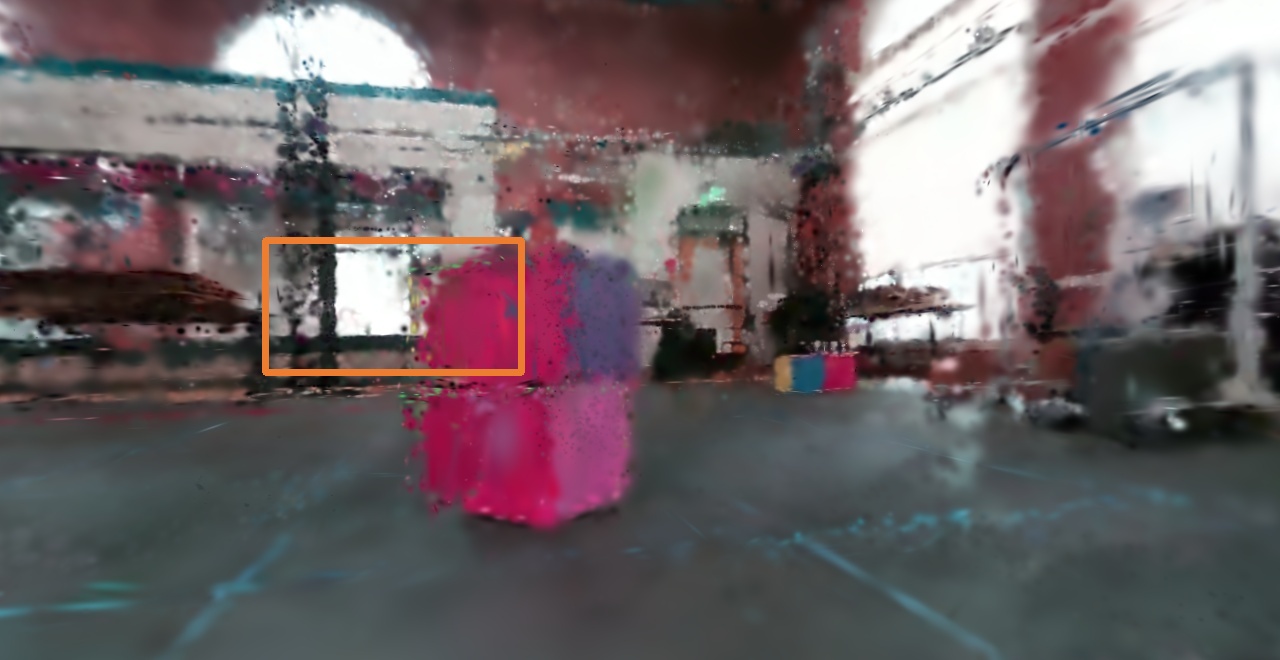} &
\includegraphics[height=\sz\linewidth]{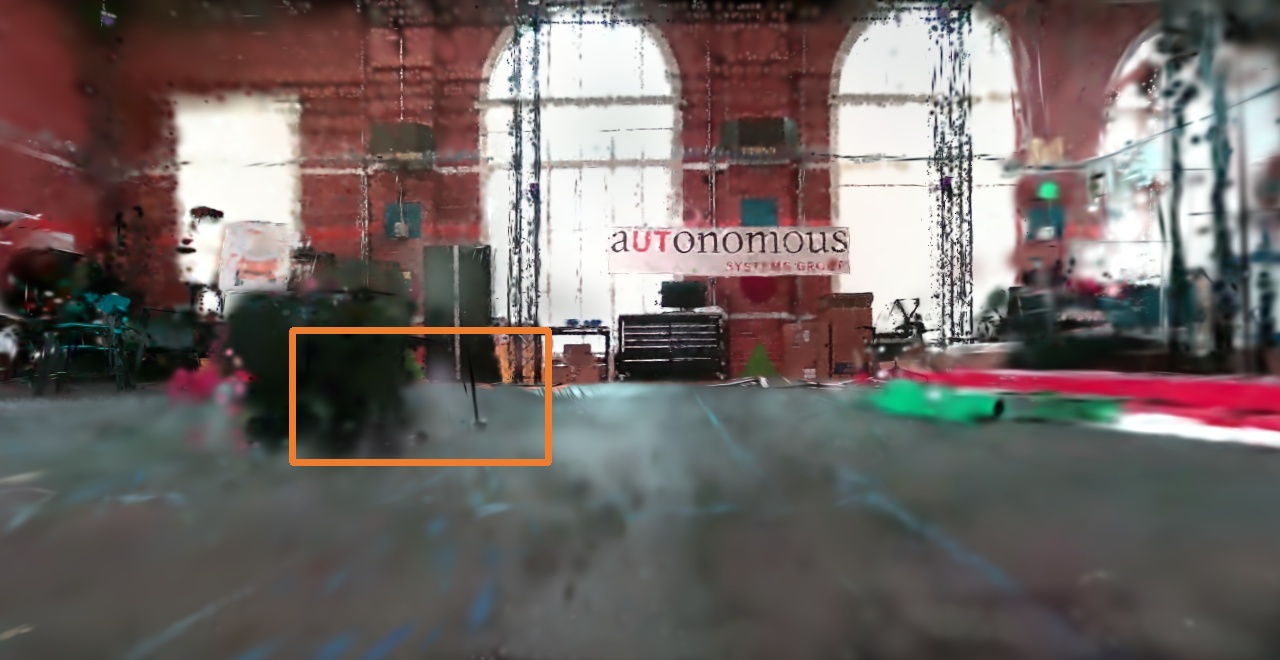} &
\includegraphics[height=\sz\linewidth]{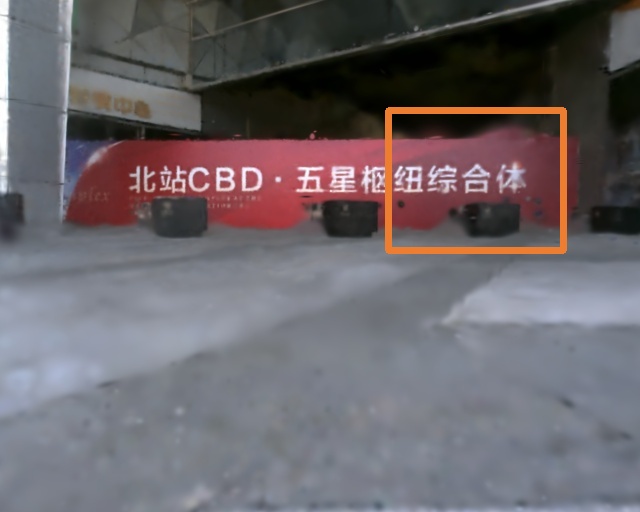} &
\includegraphics[height=\sz\linewidth]{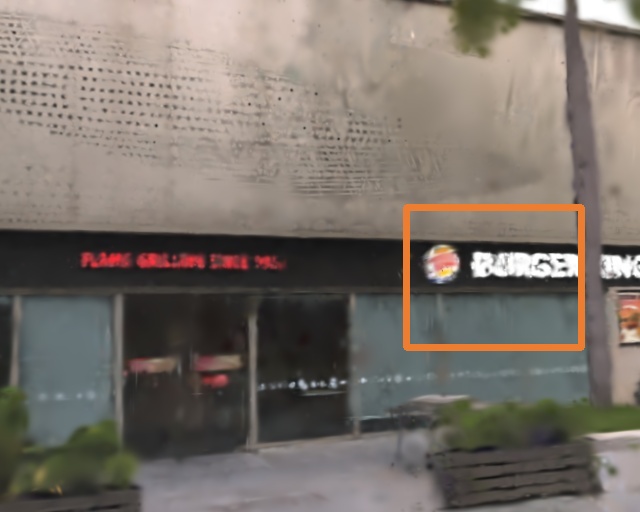} \\
&
\includegraphics[height=\sz\linewidth]{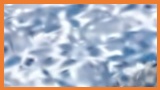} &
\includegraphics[height=\sz\linewidth]{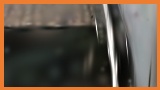} &
\includegraphics[height=\sz\linewidth]{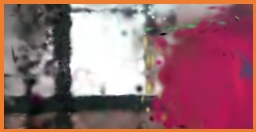} &
\includegraphics[height=\sz\linewidth]{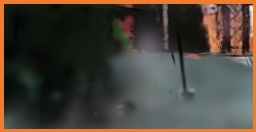} &
\includegraphics[height=\sz\linewidth]{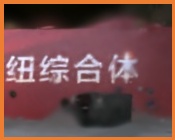} &
\includegraphics[height=\sz\linewidth]{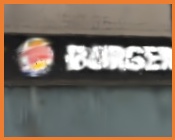} \\
\noalign{\vskip 0.5pt}
\hdashline
\noalign{\vskip 1.5pt}

\raisebox{-0.2\height}[0pt][0pt]{\rotatebox[origin=c]{90}{\textbf{Ours}}}  &
\includegraphics[height=\sz\linewidth]{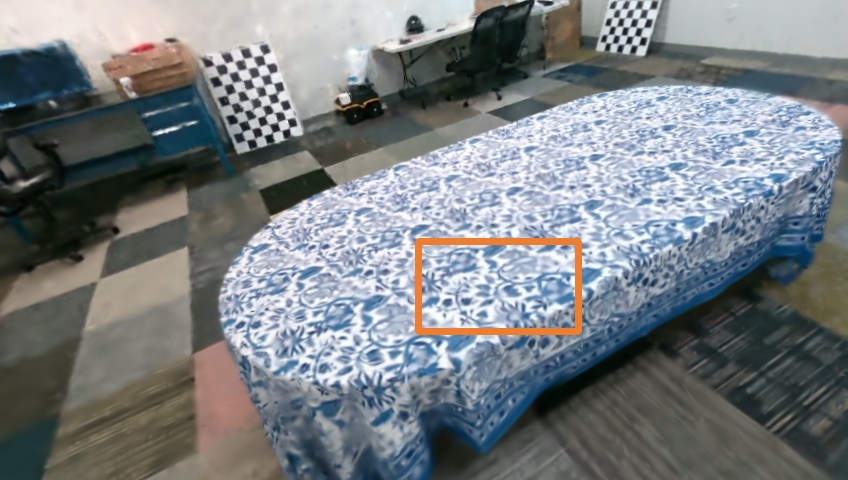} &
\includegraphics[height=\sz\linewidth]{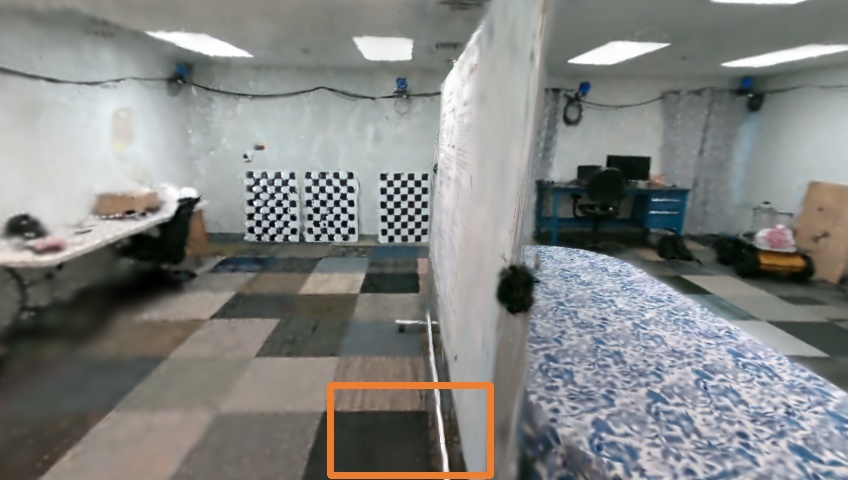} &
\includegraphics[height=\sz\linewidth]{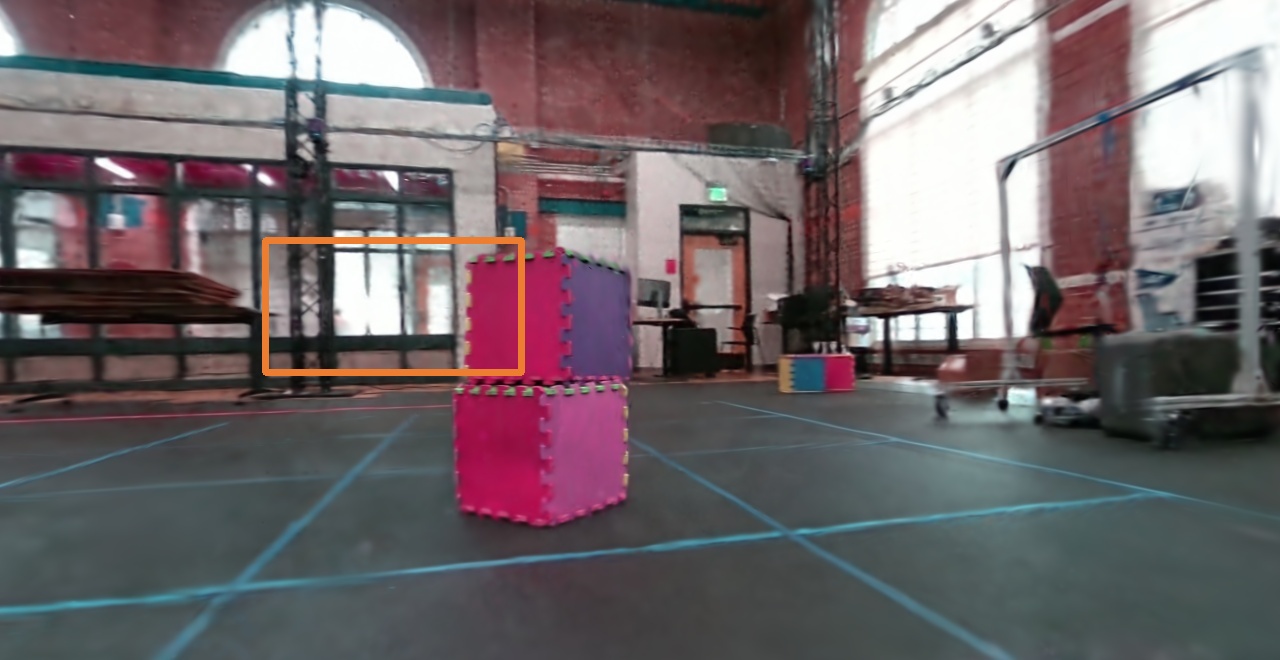} &
\includegraphics[height=\sz\linewidth]{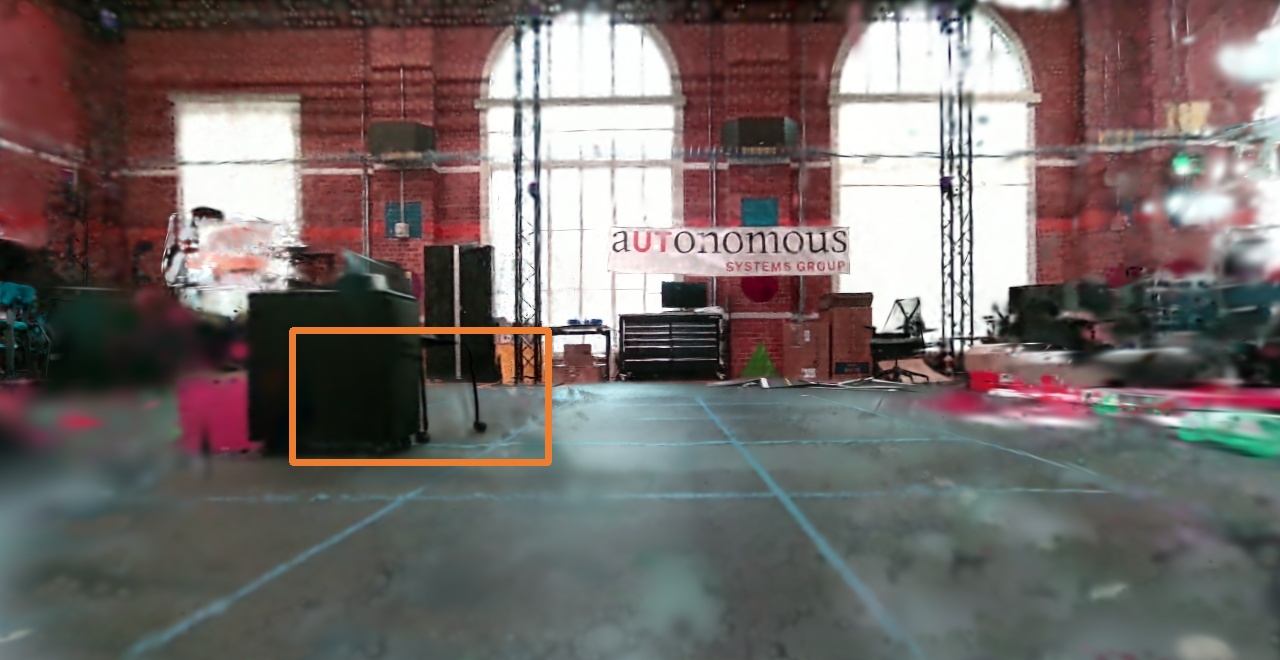} &
\includegraphics[height=\sz\linewidth]{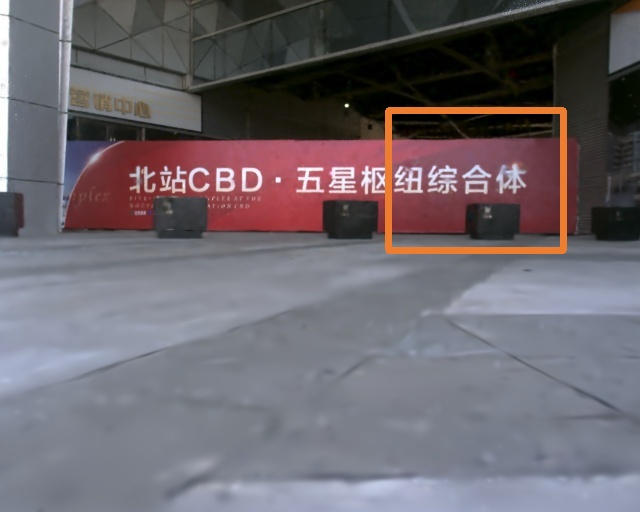} &
\includegraphics[height=\sz\linewidth]{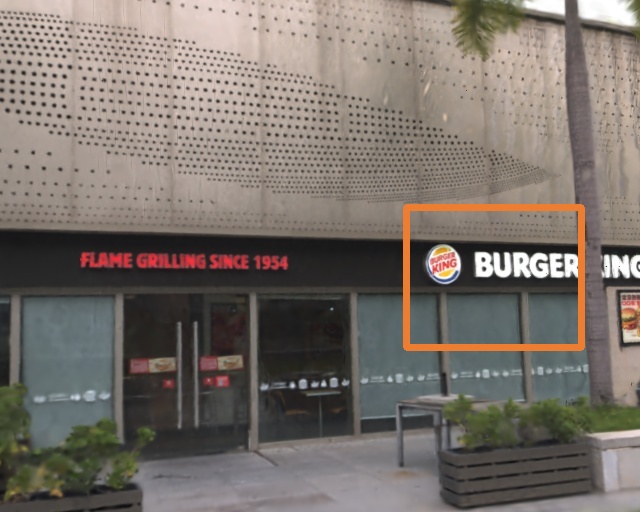} \\
&
\includegraphics[height=\sz\linewidth]{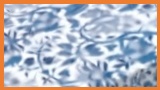} &
\includegraphics[height=\sz\linewidth]{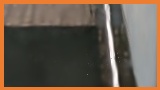} &
\includegraphics[height=\sz\linewidth]{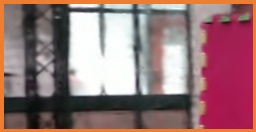} &
\includegraphics[height=\sz\linewidth]{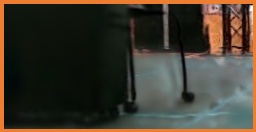} &
\includegraphics[height=\sz\linewidth]{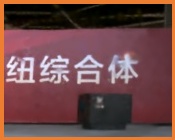} &
\includegraphics[height=\sz\linewidth]{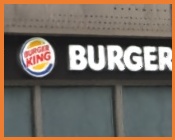} \\

\noalign{\vskip 0.5pt}
\hdashline
\noalign{\vskip 1.5pt}
\raisebox{-0.2\height}[0pt][0pt]{\rotatebox[origin=c]{90}{GT}} &
\includegraphics[height=\sz\linewidth]{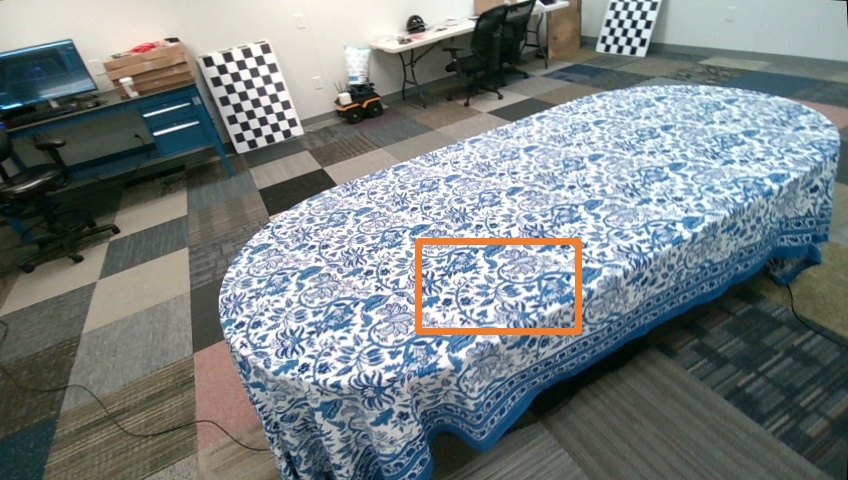} &
\includegraphics[height=\sz\linewidth]{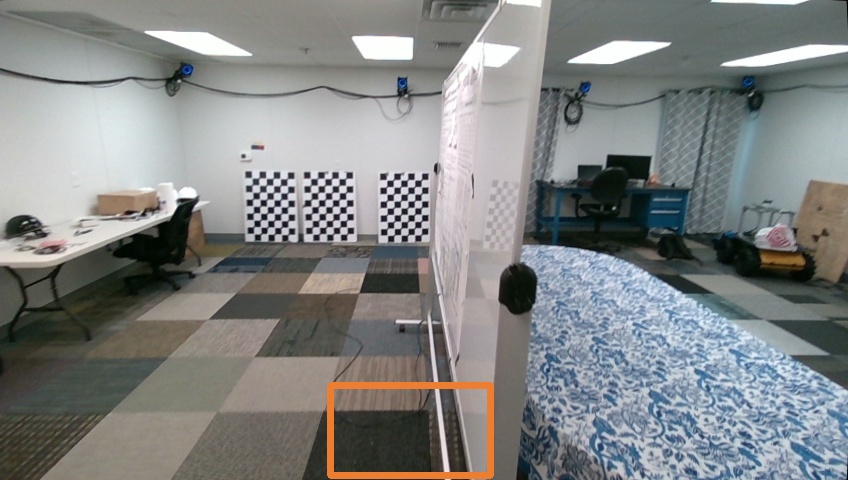} &
\includegraphics[height=\sz\linewidth]{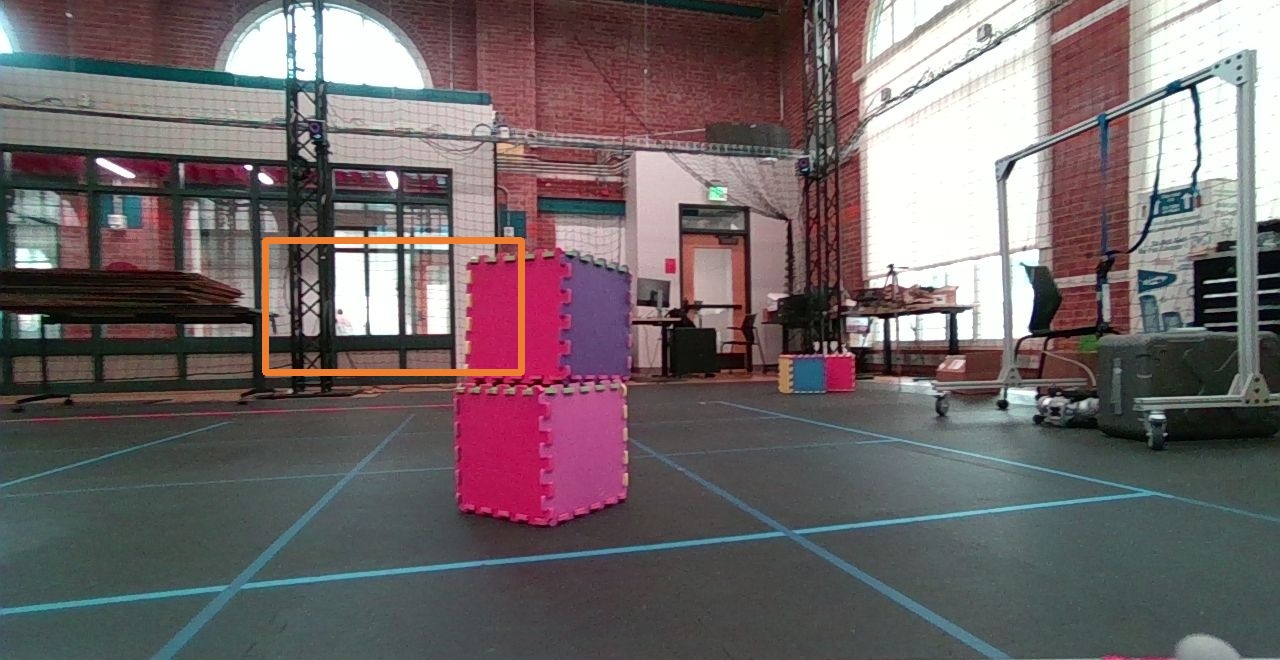} &
\includegraphics[height=\sz\linewidth]{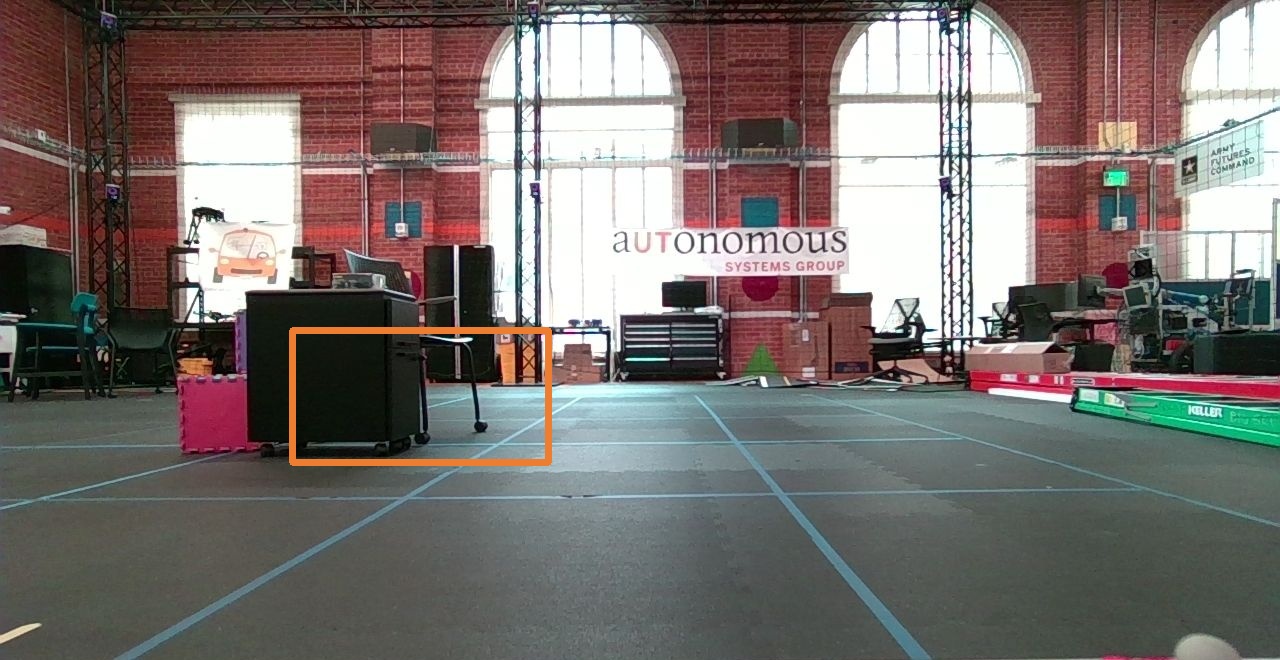} &
\includegraphics[height=\sz\linewidth]{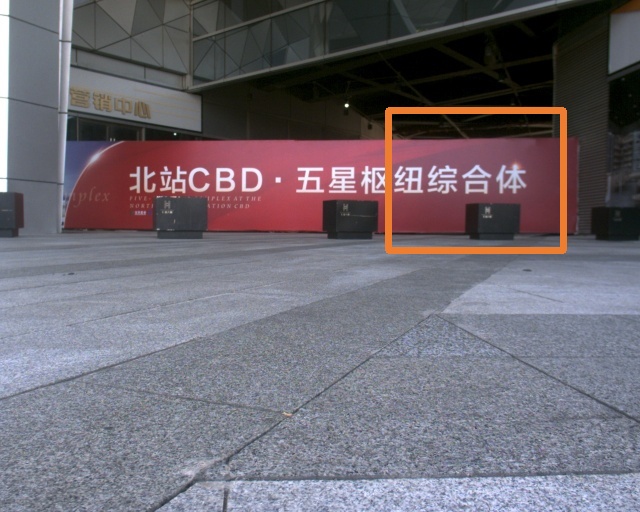} &
\includegraphics[height=\sz\linewidth]{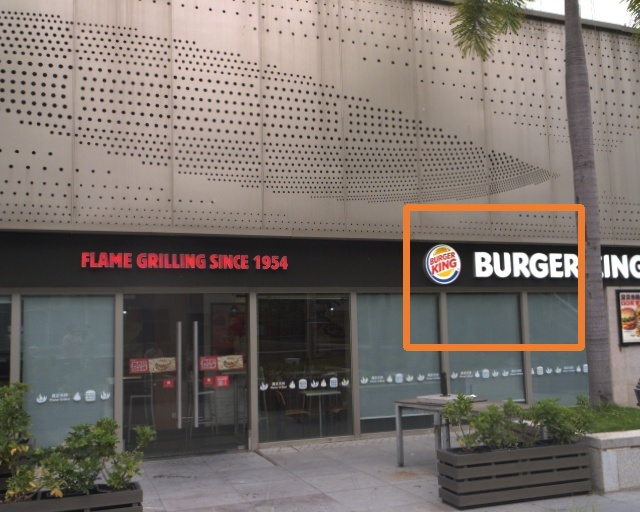} \\
&
\includegraphics[height=\sz\linewidth]{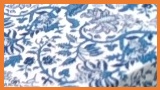} &
\includegraphics[height=\sz\linewidth]{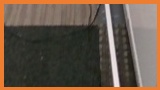} &
\includegraphics[height=\sz\linewidth]{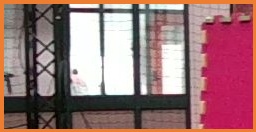} &
\includegraphics[height=\sz\linewidth]{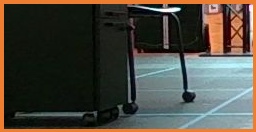} &
\includegraphics[height=\sz\linewidth]{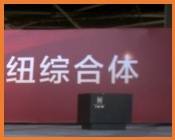} &
\includegraphics[height=\sz\linewidth]{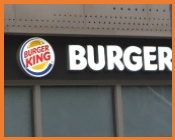} \\
& \tt{table\_01} & \tt{table\_06} & \tt{EgoDrv} & \tt{Sq-2} & \tt{CBD2} & \tt{Retail} \\
\end{tabular}
\caption{\textbf{Novel View Synthesis Results across Datasets.} Sequences are sampled from RPNG~\cite{Chen2023rpng} (\texttt{table\_01}, \texttt{table\_06}), UTMM~\cite{sun2024mm3dgs} (\texttt{EgoDrv}, \texttt{Sq-2}), and FAST-LIVO2~\cite{zheng2024fast} (\texttt{CBD2}, \texttt{Retail}) datasets.}
\label{fig:rendering_all}
\vspace{-4mm}
\end{figure*}

\begin{table}[t]
\centering
\setlength{\tabcolsep}{5pt}
\scriptsize
\caption{\textbf{Rendering Evaluation across Datasets.} As HI-SLAM2 \cite{zhang2024hislam2} fails on one sequence in the RPNG dataset~\cite{Chen2023rpng}, we compute averages over the remaining sequences. Detailed per-sequence results are provided in the supplementary material.}
\begin{tabular}{llrrr
}
\toprule
Metric & Method & RPNG & UTMM & FAST-LIVO2 \\
\midrule

\multirow{4}{*}{PSNR $\uparrow$}
& VINGS-Mono~\cite{wu2025vings}  & 11.03 & 11.85 & 10.36 \\
& Splat-SLAM~\cite{sandstrom2024splat} & 17.32 & 13.56 & 13.96 \\
& HI-SLAM2~\cite{zhang2024hislam2}     & 21.12   & 18.84 & 21.49 \\
& \textbf{VIGS-SLAM (Ours)}                        & \textbf{22.21} & \textbf{20.87} & \textbf{23.15} \\
\midrule

\multirow{4}{*}{SSIM $\uparrow$}
& VINGS-Mono~\cite{wu2025vings}  & 0.264 & 0.408 & 0.343\\
& Splat-SLAM~\cite{sandstrom2024splat} & 0.543 & 0.470 & 0.484 \\
& HI-SLAM2~\cite{zhang2024hislam2}     & 0.685   & 0.632 & 0.692 \\
& \textbf{VIGS-SLAM (Ours)}                       & \textbf{0.723} & \textbf{0.687} & \textbf{0.729} \\
\midrule

\multirow{4}{*}{LPIPS $\downarrow$}
& VINGS-Mono~\cite{wu2025vings}  &  0.704 & 0.660 & 0.724\\
& Splat-SLAM~\cite{sandstrom2024splat} & 0.465 & 0.653 & 0.745 \\
& HI-SLAM2~\cite{zhang2024hislam2}     & 0.358   & 0.501 & 0.560 \\
& \textbf{VIGS-SLAM (Ours)}                         & \textbf{0.314} & \textbf{0.441} & \textbf{0.487} \\
\bottomrule
\end{tabular}

\vspace{-1mm}
\label{tab:mapping_avg}
\end{table}

\begin{table*}[t]
\centering
\footnotesize
\caption{\textbf{Tracking Performance on UTMM Dataset~\cite{sun2024mm3dgs}} (ATE RMSE $\downarrow$ [cm]). Best results are highlighted as \colorbox{colorFst}{\bf first},\colorbox{colorSnd}{second}, and\colorbox{colorTrd}{third}. `F' indicates failure. MM3DGS-SLAM~\cite{sun2024mm3dgs} has not open-sourced its non-LiDAR variant; we use the paper’s reported metrics and indicate unreported sequences with `--'.} %
\resizebox{\textwidth}{!}{
\begin{tabular}{lrrrrrrrr
!{\smash{\tikz[baseline]{\draw[densely dashed, gray!80, line width=0.8pt] (0pt,-2pt)--(0pt,8pt);}}}
r}
\toprule
{Method} & \tt{Ego-1} & \tt{Ego-2} & \tt{EgoDrv} & \tt{FastStr} & \tt{SStr-1} & \tt{SStr-2} & \tt{Sq-1} & \tt{Sq-2} & \textbf{{Avg.}} \\
\midrule
\multicolumn{10}{l}{\cellcolor[HTML]{EEEEEE}{\textit{RGB}}} \\
MM3DGS-SLAM (RGB)~\cite{sun2024mm3dgs}     & 4.09 & \multicolumn{1}{r}{--} & 67.20 & 25.78 & \multicolumn{1}{r}{--} & \multicolumn{1}{r}{--} & 59.48 & -- & N/A \\
Splat-SLAM~\cite{sandstrom2024splat}       & \fs 1.38 & \fs 0.62 & \nd3.26 & \fs 0.95 & 5.48 & \fs 0.54 & 103.60 & 71.29 & 23.39 \\
HI-SLAM2~\cite{zhang2024hislam2}           & 2.06 & 3.35 & 4.36 & \nd 0.99 & \rd 0.71 & \rd 0.84 & \rd 27.85 & \rd 24.63 & \nd 8.10 \\
DROID-SLAM~\cite{teed2021droid}            & \rd 2.00 & \rd 3.17 & 30.94 & 1.30 & 0.97 & 0.86 & \nd 14.95 & \fs 9.10 & \rd 7.91 \\
\hdashline
\noalign{\vskip 1pt}
\multicolumn{10}{l}{\cellcolor[HTML]{EEEEEE}{\textit{RGB+IMU}}} \\
ORB-SLAM3~\cite{campos2021orb3}            & 3.64 & \multicolumn{1}{r}{F}  & \rd3.53 & \multicolumn{1}{r}{F}  & \multicolumn{1}{r}{F}  & \multicolumn{1}{r}{F}  & \multicolumn{1}{r}{F}  & F  & N/A \\
MM3DGS-SLAM (RGB+IMU)~\cite{sun2024mm3dgs} & 3.41 & \multicolumn{1}{r}{--} & 68.50 & 16.78 & \multicolumn{1}{r}{--} & \multicolumn{1}{r}{--} & 44.26 & -- & N/A \\
VINS-Mono~\cite{qin2018vins}               & 127.01 & 3.97 & 92.00 & 3.43 & \nd 0.61 & 2.06 & 262.38 & 211.16 & 87.83 \\
DBA-Fusion~\cite{zhou2024dba} & 13.27 & 5.45 & 65.01 & 118.77 & \fs 0.37 & 1.02 & 88.93 & 133.28 & 53.26 \\
OPEN-VINS~\cite{geneva2020openvins}        & 117.27 & 3.43 & 26.49 & 6.29 & 5.58 & 5.98 & 50.53 & 34.39 & 31.50 \\
VINGS-Mono~\cite{wu2025vings} & 5.00 & 7.29 & 12.43 & 13.36 & 0.67 & \nd 0.71 & 36.01 & 25.70 & 12.54 \\
\textbf{VIGS-SLAM (Ours)}                              & \nd 1.81 & \nd 0.93 & \fs 1.45 & \rd 1.20 &  0.81 &  0.93 & \fs 2.17 & \nd 16.61 & \fs 3.24 \\
\bottomrule
\end{tabular}}

\vspace{-2mm}
\label{tab:tracking_utmm}
\end{table*}

\begin{table}[t]
\centering
\scriptsize
\setlength{\tabcolsep}{5pt}
\caption{\textbf{Tracking Performance on FAST-LIVO2 Dataset~\cite{zheng2024fast}} (ATE RMSE $\downarrow$ [cm]). Best results are highlighted as \colorbox{colorFst}{\bf first},\colorbox{colorSnd}{second}, and\colorbox{colorTrd}{third}. `F' indicates failure.}
\begin{tabular}{lrrrrr
!{\smash{\tikz[baseline]{\draw[densely dashed, gray!80, line width=0.8pt] (0pt,-2pt)--(0pt,8pt);}}}
r}
\toprule
Method & \tt{CBD1} & \tt{CBD2} & \tt{HKU} & \tt{Retail} & \tt{SYSU1} & {\textbf{Avg.}} \\
\midrule
\multicolumn{7}{l}{\cellcolor[HTML]{EEEEEE}{\textit{RGB}}} \\ 
Splat-SLAM~\cite{sandstrom2024splat} & \rd 5.52 & \nd 7.74 & \nd 4.44 & 212.01 & 313.56 & 108.65 \\
DROID-SLAM~\cite{teed2021droid}      & 72.48 & \rd 15.67 & 15.10 & \rd 16.81 & \rd 50.94 & \rd 34.20 \\
HI-SLAM2~\cite{zhang2024hislam2}     & \fs 4.38 & 24.30 & \rd 4.87 & \fs 7.20 & \nd 10.36 & \nd 10.22 \\
\hdashline
\noalign{\vskip 1pt}
\multicolumn{7}{l}{\cellcolor[HTML]{EEEEEE}{\textit{RGB+IMU}}} \\ 
OPEN-VINS~\cite{geneva2020openvins} & F & F & F & F & F & N/A \\ 
ORB-SLAM3~\cite{campos2021orb3}     & F & F & F & F & F & N/A \\
VINS-Mono~\cite{qin2018vins}        & F & F & F & 25.64 & F & N/A \\ 
DBA-Fusion~\cite{zhou2024dba} & 22.96 & 127.79 & 56.20 & 26.69 & 337.86 & 114.30\\
VINGS-Mono~\cite{wu2025vings} & 20.54 & 129.55 & 52.42 & 51.67 & 269.83 & 104.80 \\
\textbf{VIGS-SLAM (Ours)}                      & \nd 4.50 & \fs 5.76 & \fs 3.88 & \nd 8.88 & \fs 7.36 & \fs 6.08 \\
\bottomrule
\end{tabular}

\label{tab:tracking_livo2}
\vspace{-4mm}
\end{table}

\begin{figure}[!b]
\vspace{-2mm}
  \centering
  \begin{subfigure}{\linewidth}
    \centering
    \includegraphics[width=0.40\linewidth]{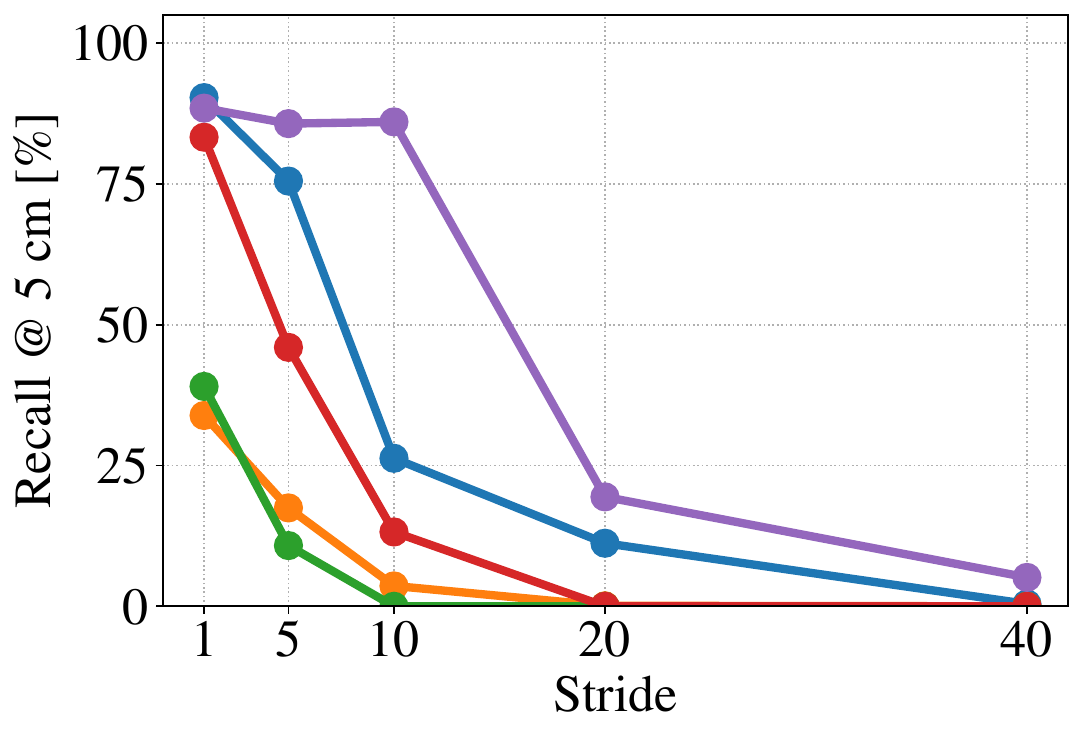}
    \includegraphics[width=0.40\linewidth]{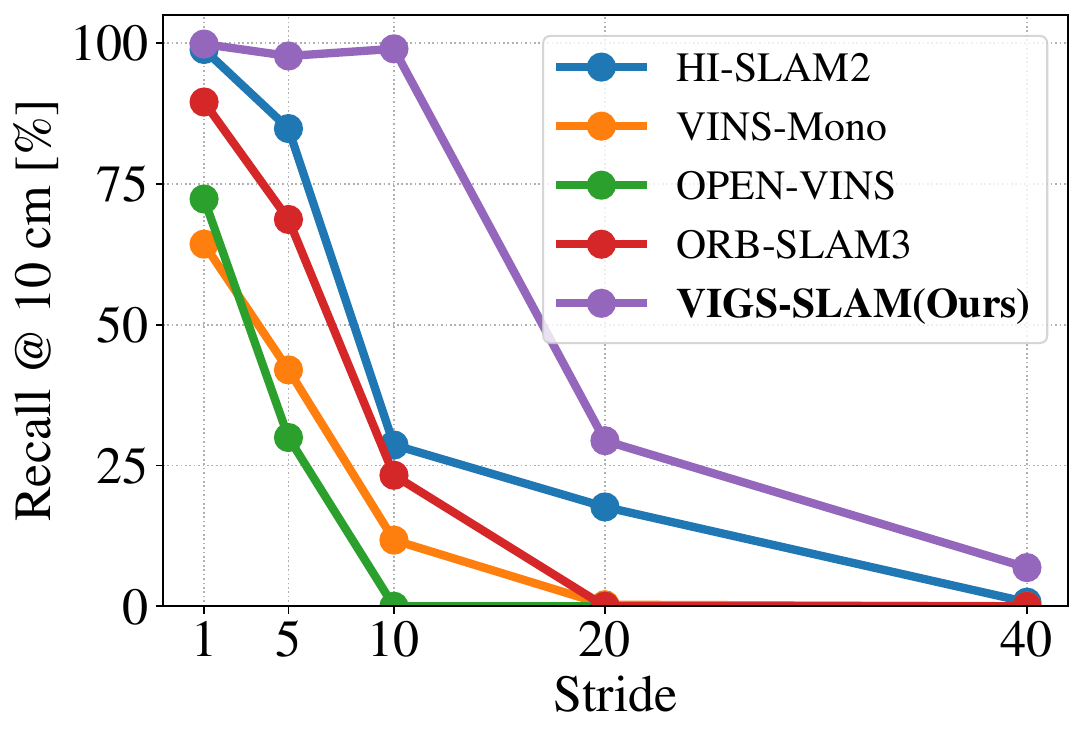}
    \caption{\textbf{Strided EuRoC Dataset~\cite{burri2016euroc}}}
    \label{fig:tracking_euroc_stride_avg}
  \end{subfigure}
  \vspace{0.6em}
  \begin{subfigure}{\linewidth}
    \centering
    \includegraphics[width=0.40\linewidth]{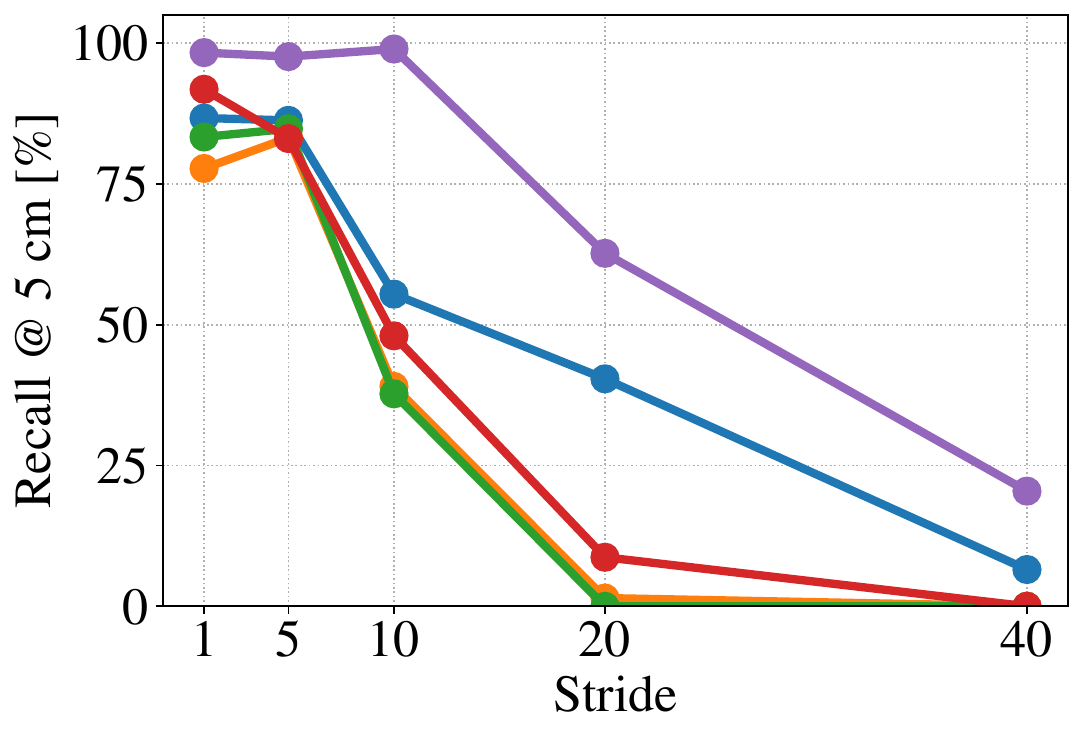}
    \includegraphics[width=0.40\linewidth]{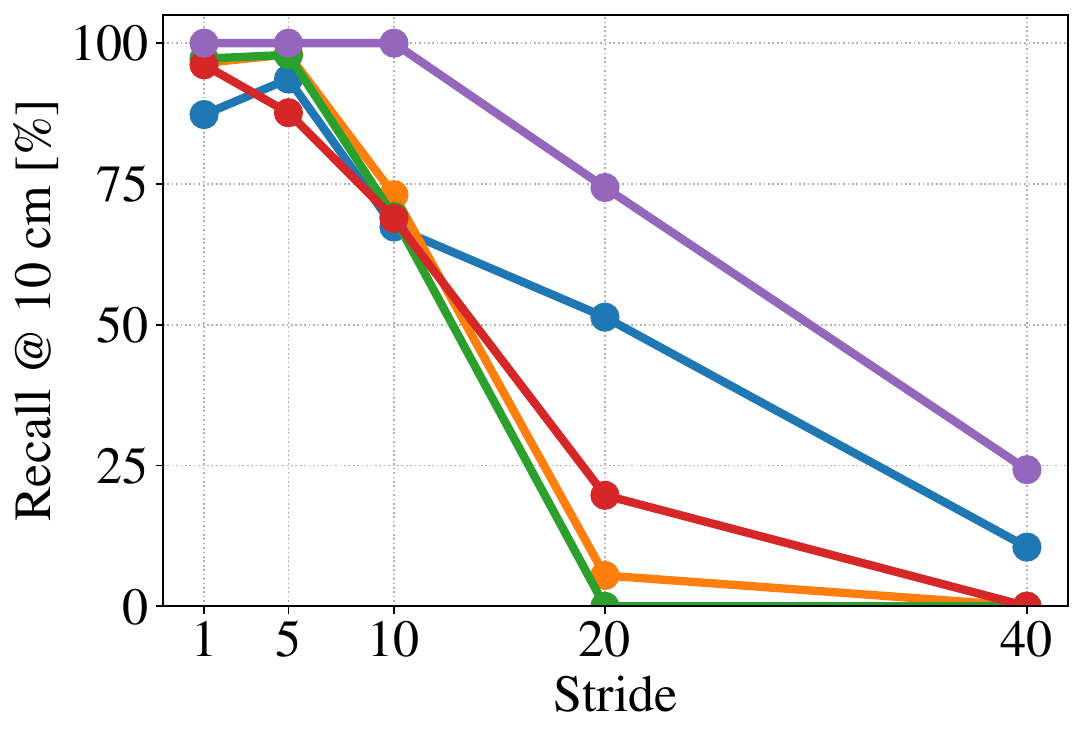}
    \caption{\textbf{Strided RPNG AR Table Dataset~\cite{Chen2023rpng}}}
    \label{fig:tracking_rpng_stride_avg}
  \end{subfigure}
\vspace{-8mm}
  \caption{\textbf{Average Tracking Performance on Strided Datasets.} We plot mean recall at 5 cm and 10 cm thresholds under different stride settings.}
  \label{fig:tracking_stride_avg}
\end{figure}

\paragraph{Baselines}
We compare our VIGS-SLAM with 15 methods.
(a) \textit{Classic Visual SLAM}: DSO~\cite{Engel2017DSO}, SVO~\cite{forster2014svo}; 
(b) \textit{Learning-based Visual SLAM}: TartanVO~\cite{wang2021tartanvo}, DROID-SLAM~\cite{teed2021droid}; 
(c) \textit{Visual Gaussian Splatting SLAM}: Splat-SLAM~\cite{sandstrom2024splat} and HI-SLAM2~\cite{zhang2024hislam2}; 
(d) \textit{classic Visual-Inertial SLAM}: MSCKF~\cite{mourikis2007msckf}, OKVIS~\cite{leutenegger2015okvis}, VINS-Mono~\cite{qin2018vins}, OPEN-VINS~\cite{geneva2020openvins}, ORB-SLAM3~\cite{campos2021orb3}; 
(e) \textit{Learning-based Visual-Inertial SLAM}: DBA-Fusion~\cite{zhou2024dba}; 
(f) \textit{Visual-Inertial Gaussian Splatting SLAM}: MM3DGS-SLAM~\cite{sun2024mm3dgs}, VINGS-Mono~\cite{wu2025vings}; 
(g) \textit{Feed-forward SLAM}: TTT3R~\cite{chen2025ttt3r}. 

To evaluate the online setting, for DROID-SLAM~\cite{teed2021droid}, Splat-SLAM~\cite{sandstrom2024splat}, HI-SLAM2~\cite{zhang2024hislam2}, and our VIGS-SLAM, we report metrics computed before the final global bundle adjustment and the final color refinement (which typically takes over 10 minutes). 
Results with these refinements are provided in the supplementary material.
For brevity, we report all methods on the EuRoC dataset~\cite{burri2016euroc} and evaluate only the stronger and representative baselines in subsequent experiments.
For DBA-Fusion~\cite{zhou2024dba} and its successor VINGS-Mono~\cite{wu2025vings}, we worked closely with the first author of VINGS-Mono~\cite{wu2025vings} and made targeted modifications to improve their performance (see supplementary material for details).
As confirmed by its authors, MM3DGS-SLAM~\cite{sun2024mm3dgs} primarily targets an RGB+LiDAR+IMU setup; the open-sourced code does not fully support a pure visual-inertial setting. It is evaluated only on the UTMM dataset~\cite{sun2024mm3dgs}, and we copy the tracking results from the paper.

\paragraph{Metrics} For camera tracking, we align the estimated trajectory to the ground truth using \texttt{evo}~\cite{grupp2017evo}, and report the Absolute Trajectory Error (ATE) in terms of RMSE~\cite{sturm2012benchmark}.
In addition, we report \emph{Recall}  -- percentage of ground-truth poses whose translation error to the trajectory is below a threshold.
For rendering evaluation, we report PSNR, SSIM, and LPIPS on frames that are not used as keyframes by any method, excluding all views involved in mapping. Consequently, the rendering results reported by MM3DGS-SLAM~\cite{sun2024mm3dgs} are not directly comparable to ours.
In the tables, we use ‘F’ to denote failure, either unable to initialize or significant drift. 

\subsection{Mapping, Tracking, and Rendering}

\paragraph{EuRoC Dataset~\cite{burri2016euroc}}
As shown in \tabref{tab:tracking_euroc}, our approach achieves the best overall ATE, ranking first or second in most sequences. In contrast, purely feedforward methods like TTT3R~\cite{chen2025ttt3r} accumulate large drift over time. 
ORB-SLAM3~\cite{campos2021orb3}, a highly engineered system with robust optimization and loop-closure mechanisms, lags behind our method. We attribute this to its reliance on handcrafted sparse features and non-differentiable components, which limit its robustness. In contrast, our system leverages dense, learning-based visual correspondences and tightly couples inertial constraints into a unified framework.

\paragraph{RPNG AR Table Dataset~\cite{Chen2023rpng}}
The tracking and rendering accuracies are shown in \tabref{tab:tracking_rpng} and \tabref{tab:mapping_avg}, respectively.
The proposed VIGS-SLAM achieves the lowest tracking errors, halving the error of the second-best VINS-Mono \cite{qin2018vins}, while also leading to the highest novel view synthesis scores.
Qualitative results in \figref{fig:rendering_all} show sharper and more faithful high-frequency details (especially in the tablecloth and carpet),
while baselines produce blurred textures. 

\paragraph{UTMM Dataset~\cite{sun2024mm3dgs}}
We additionally evaluate our method on the UTMM dataset, introduced in MM3DGS-SLAM~\cite{sun2024mm3dgs}. 
Since the non-LiDAR variant of MM3DGS-SLAM has not been publicly released, we compare against the results reported in their paper. 
As shown in \tabref{tab:tracking_utmm}, our method achieves substantially more accurate tracking results than baselines. 
Rendering results are shown in \figref{fig:rendering_all} and \tabref{tab:mapping_avg}. This dataset is particularly challenging due to its cluttered backgrounds with fine-grained structures, coupled with foreground objects that introduce strong depth discontinuities and complex occlusions. Nonetheless, our method produces sharp renderings with minimal floating artifacts.

\paragraph{FAST-LIVO2 Dataset~\cite{zheng2024fast}}
We further evaluate on the challenging FAST-LIVO2 dataset, which features outdoor sequences with low frame rates, shaky motion, and reflective surfaces. As shown in \tabref{tab:tracking_livo2}, classical VIO systems exhibit frequent initialization failures despite repeated restarts. 
Our method robustly initializes from the first frame and maintains stable tracking throughout all sequences, achieving the lowest average tracking error. 
Rendering results in \tabref{tab:mapping_avg} and \figref{fig:rendering_all} show sharp and consistent reconstructions with preserved text and edges, benefiting from effective loop-closure handling that maintains a consistent Gaussian map, especially in outdoor sequences with large loops.

\begin{figure}[!b]
\centering
\vspace{-1mm}
\begin{minipage}{0.46\linewidth}
    \centering
    \includegraphics[width=\linewidth]{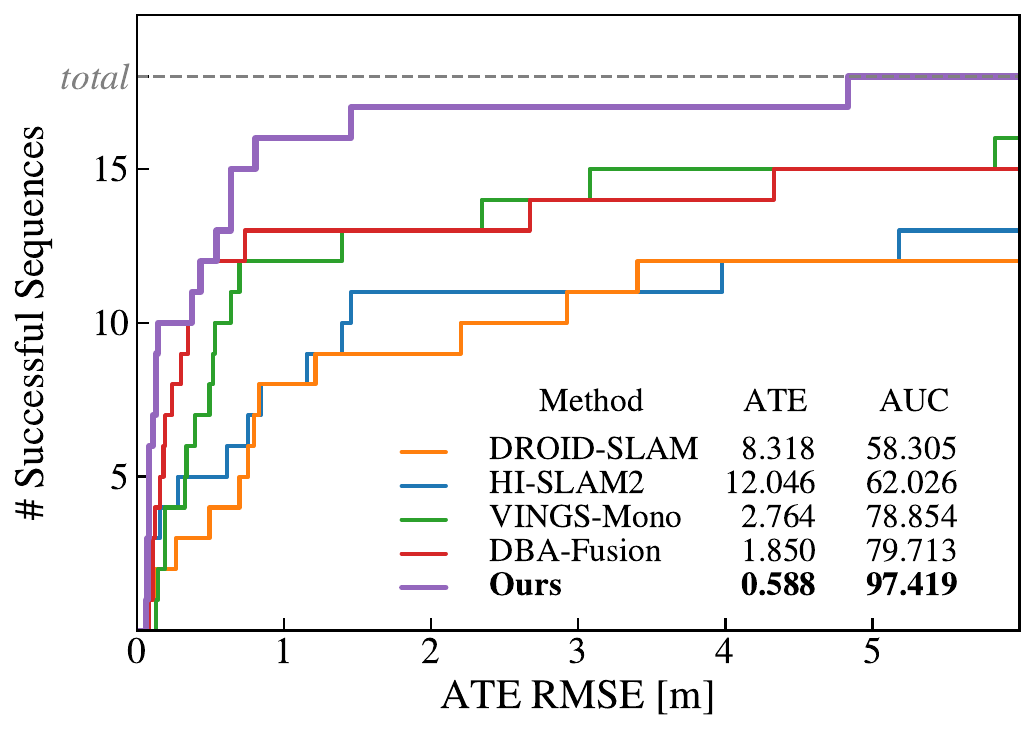}
\end{minipage}
\begin{minipage}{0.52\linewidth}
    \centering
    \begin{tabular}{cc}
        \includegraphics[width=0.48\linewidth]{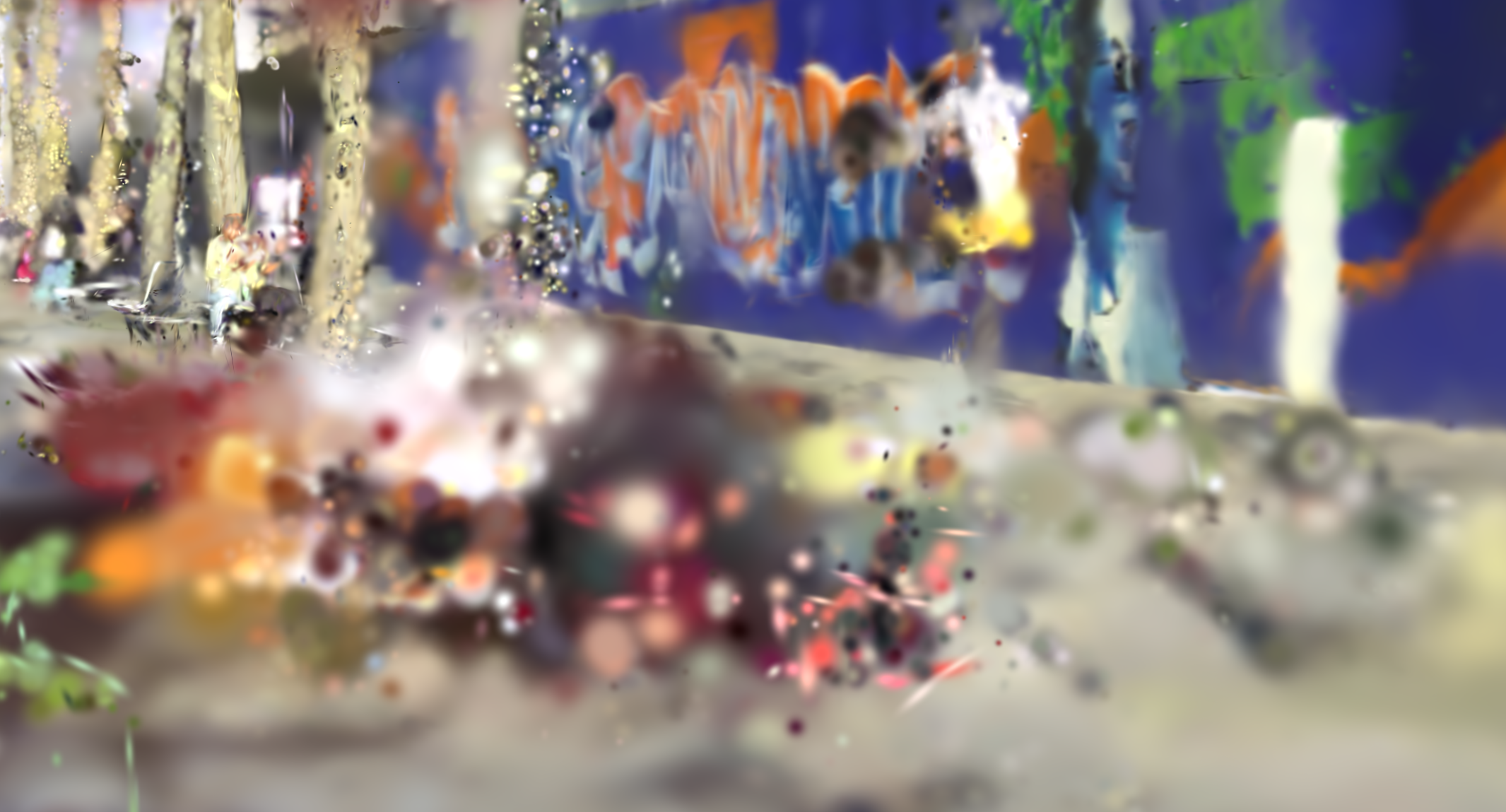} &
        \includegraphics[width=0.48\linewidth]{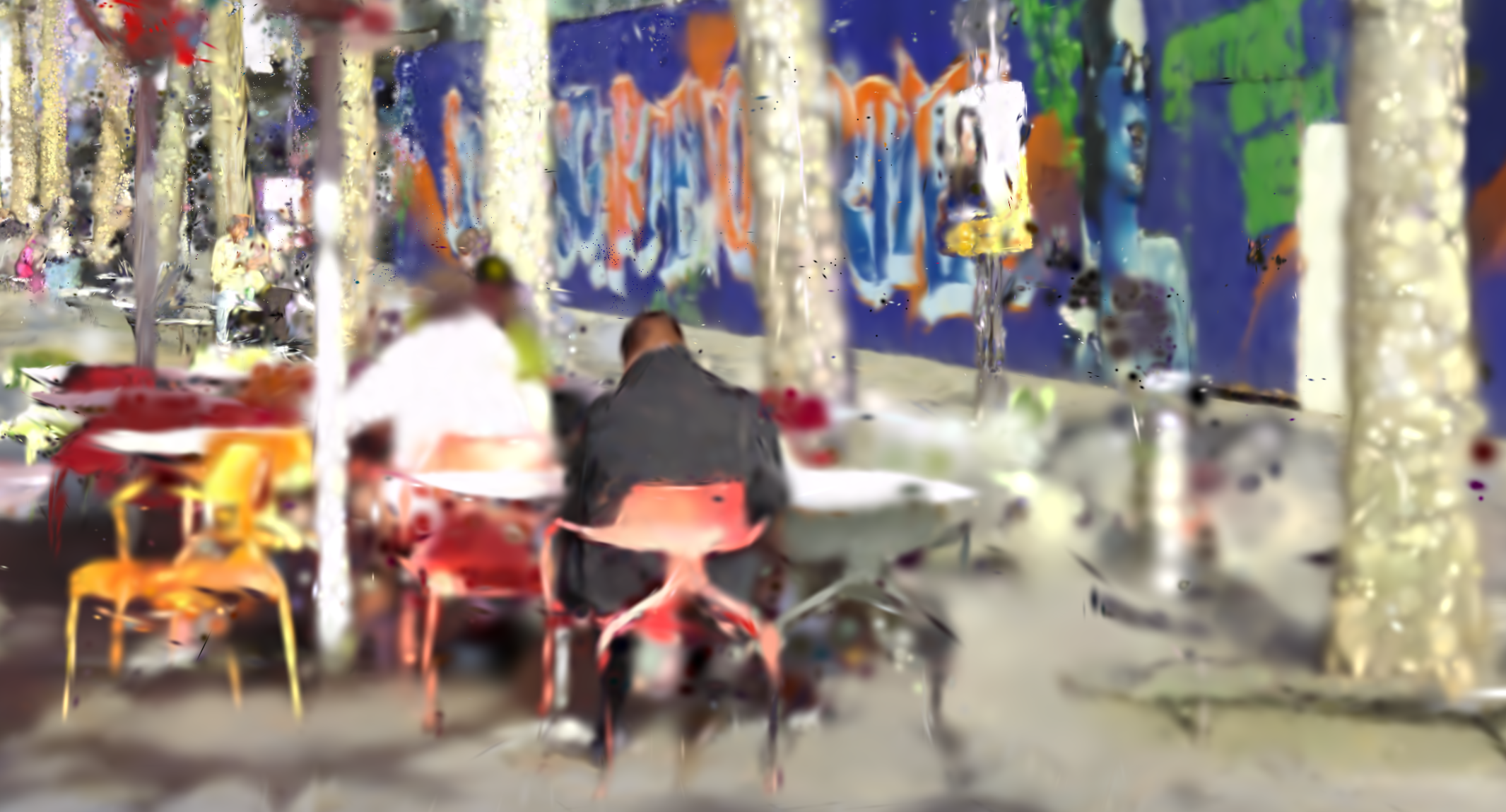} \\
        \includegraphics[width=0.48\linewidth]{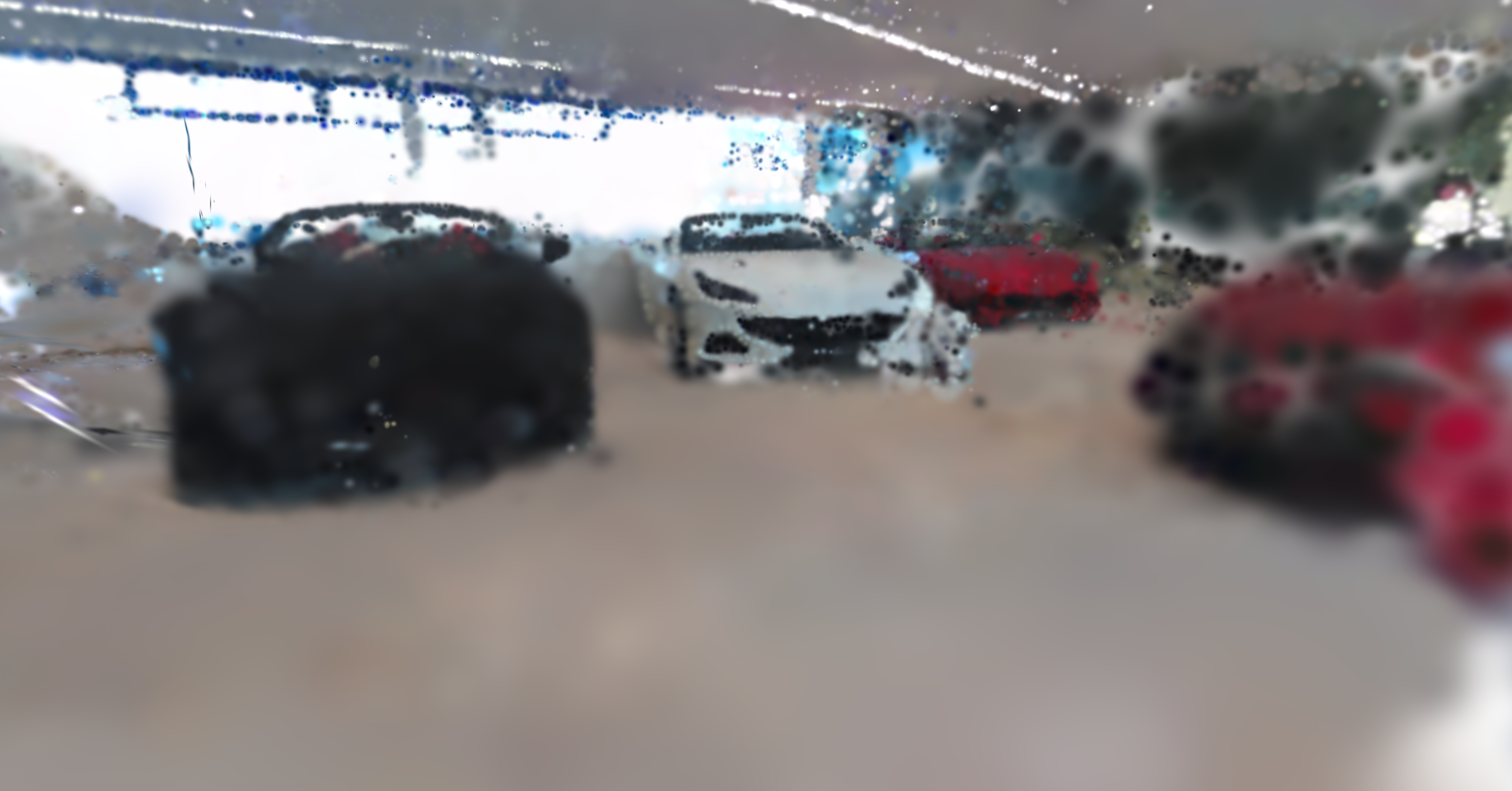} &
        \includegraphics[width=0.48\linewidth]{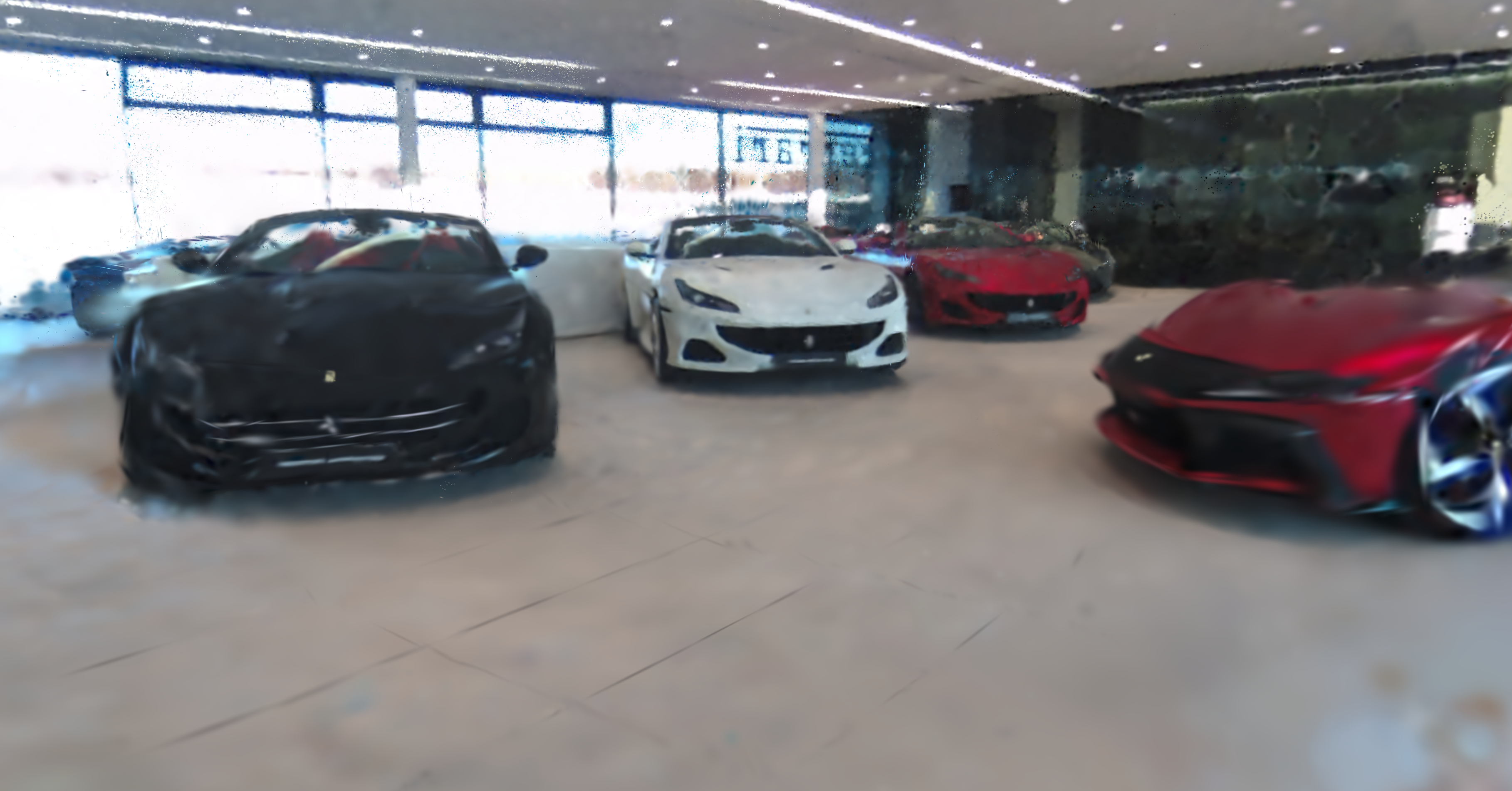} \\
        {\scriptsize{HI-SLAM2}~\cite{zhang2024hislam2}} & {\scriptsize{\textbf{VIGS-SLAM (Ours)}}}\\
    \end{tabular}
\end{minipage}

\caption{\textbf{Evaluation on the Self-Captured Dataset.} Left: cumulative success curve under different ATE RMSE thresholds, together with the average ATE RMSE and the AUC up to 6 meters.
Right: renderings from extrapolated views.}
\label{fig:ate_step_curve}

\end{figure}
\begin{figure*}[!b]
\centering
\scriptsize
\setlength{\tabcolsep}{1pt}
\renewcommand{\arraystretch}{0.5}

\newcommand{\sz}{0.325} %
\setlength{\arrayrulewidth}{0.8pt}
\begin{tabular}{lccc}
&
\hspace{12pt}\includegraphics[width=\dimexpr\sz\linewidth-12pt]{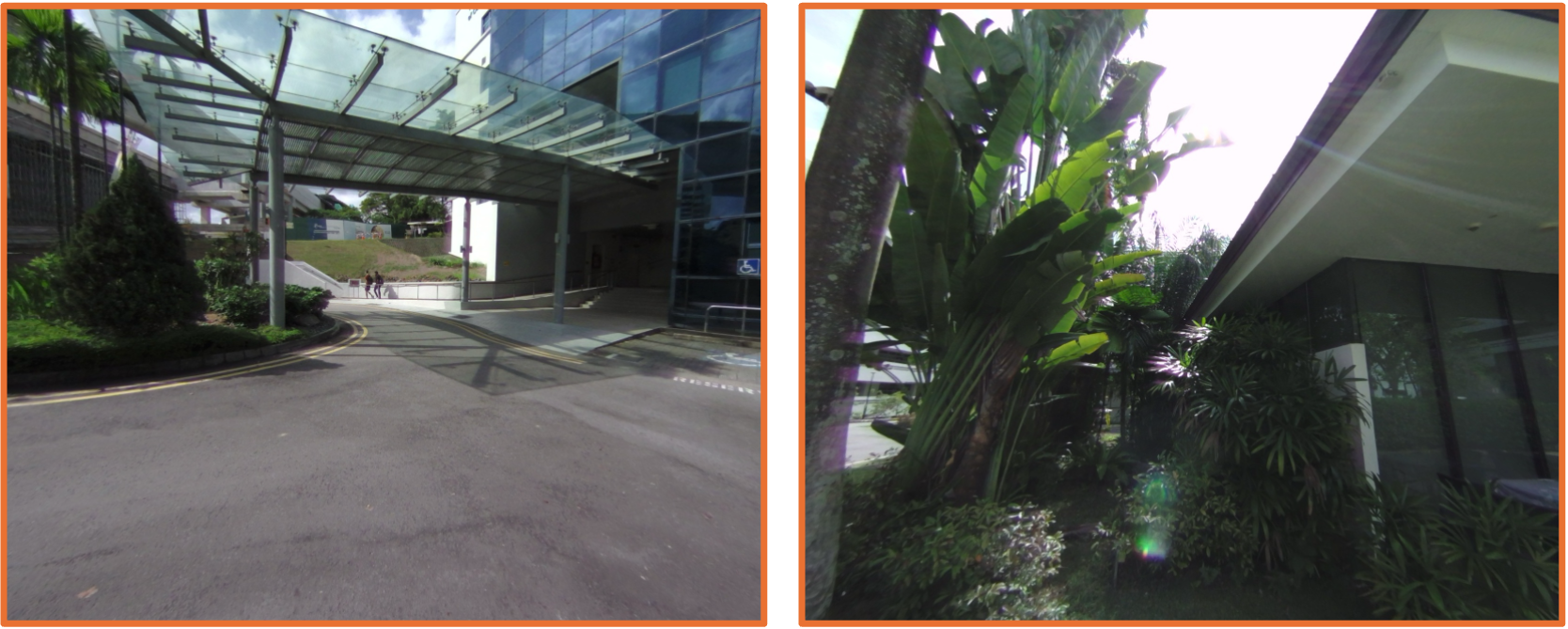} &
\hspace{12pt}\includegraphics[width=\dimexpr\sz\linewidth-12pt]{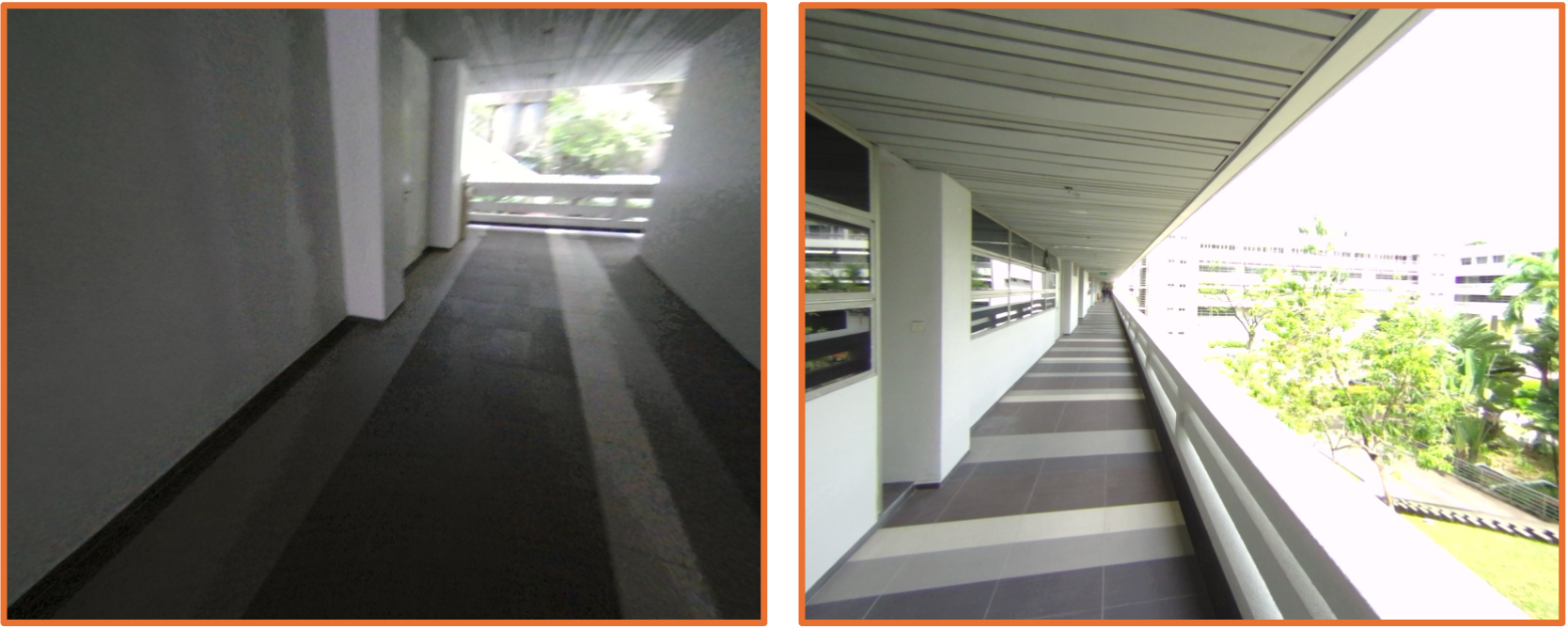} &
\hspace{10pt}\includegraphics[width=\dimexpr\sz\linewidth-10pt]{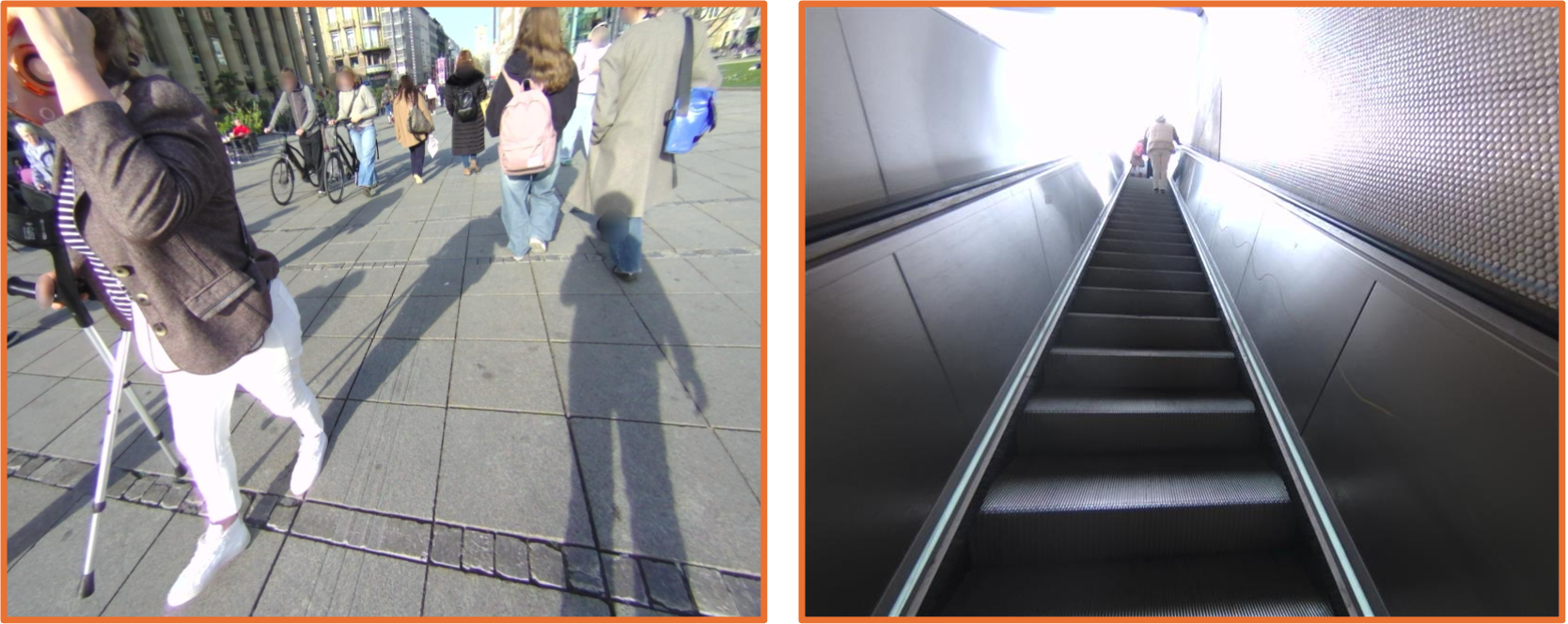} \\
&
\includegraphics[width=\sz\linewidth]{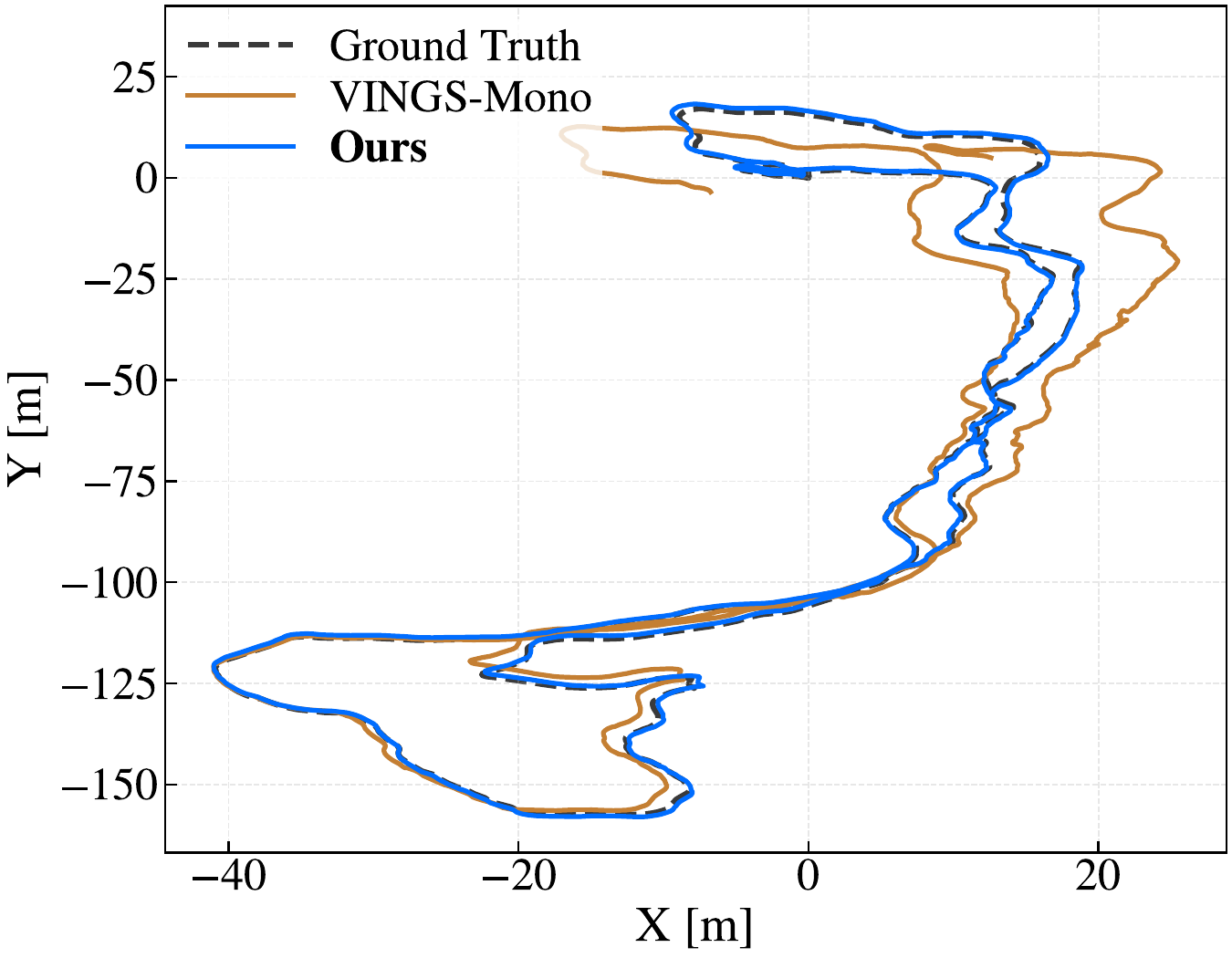} &
\includegraphics[width=\sz\linewidth]{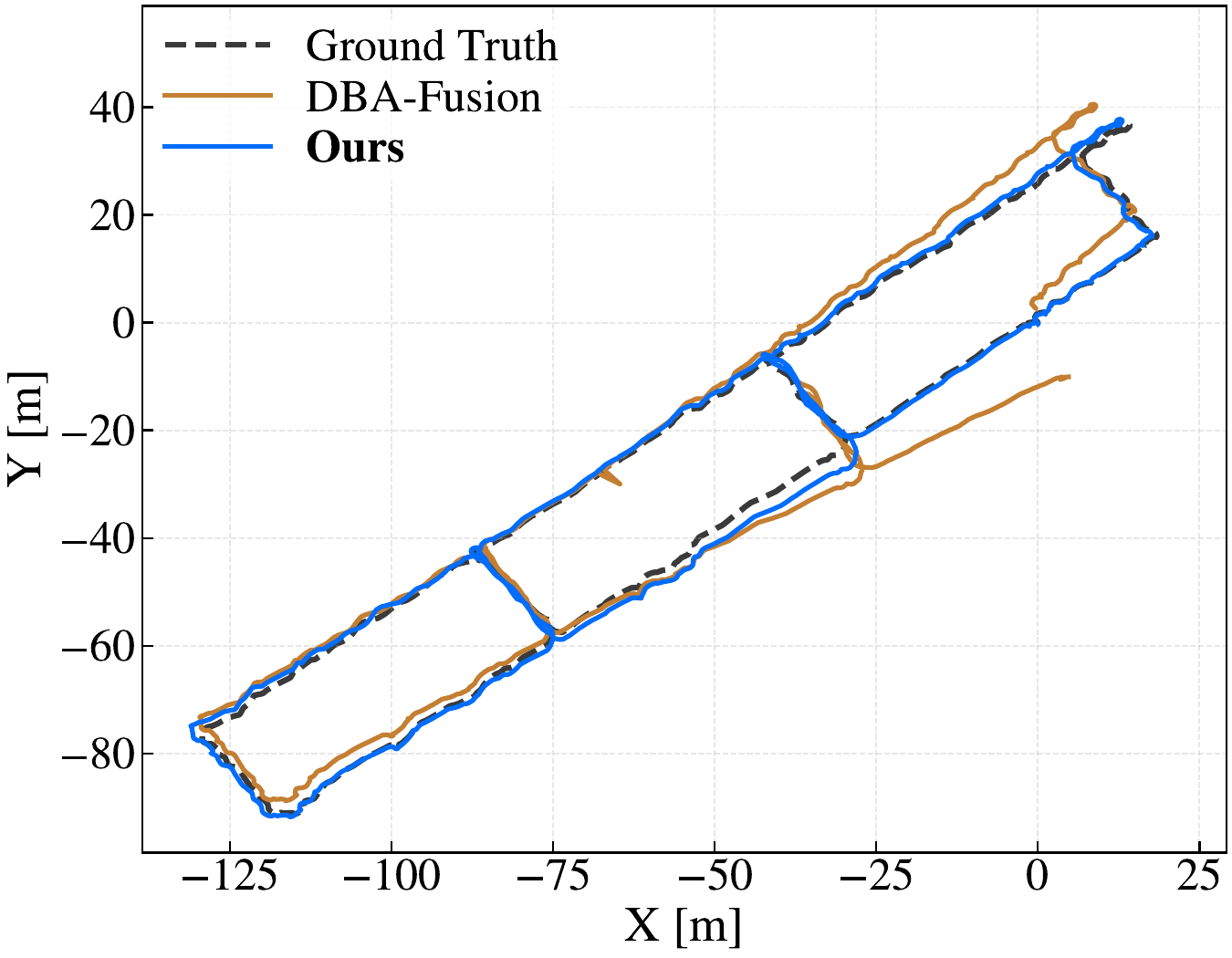} &
\includegraphics[width=\sz\linewidth]{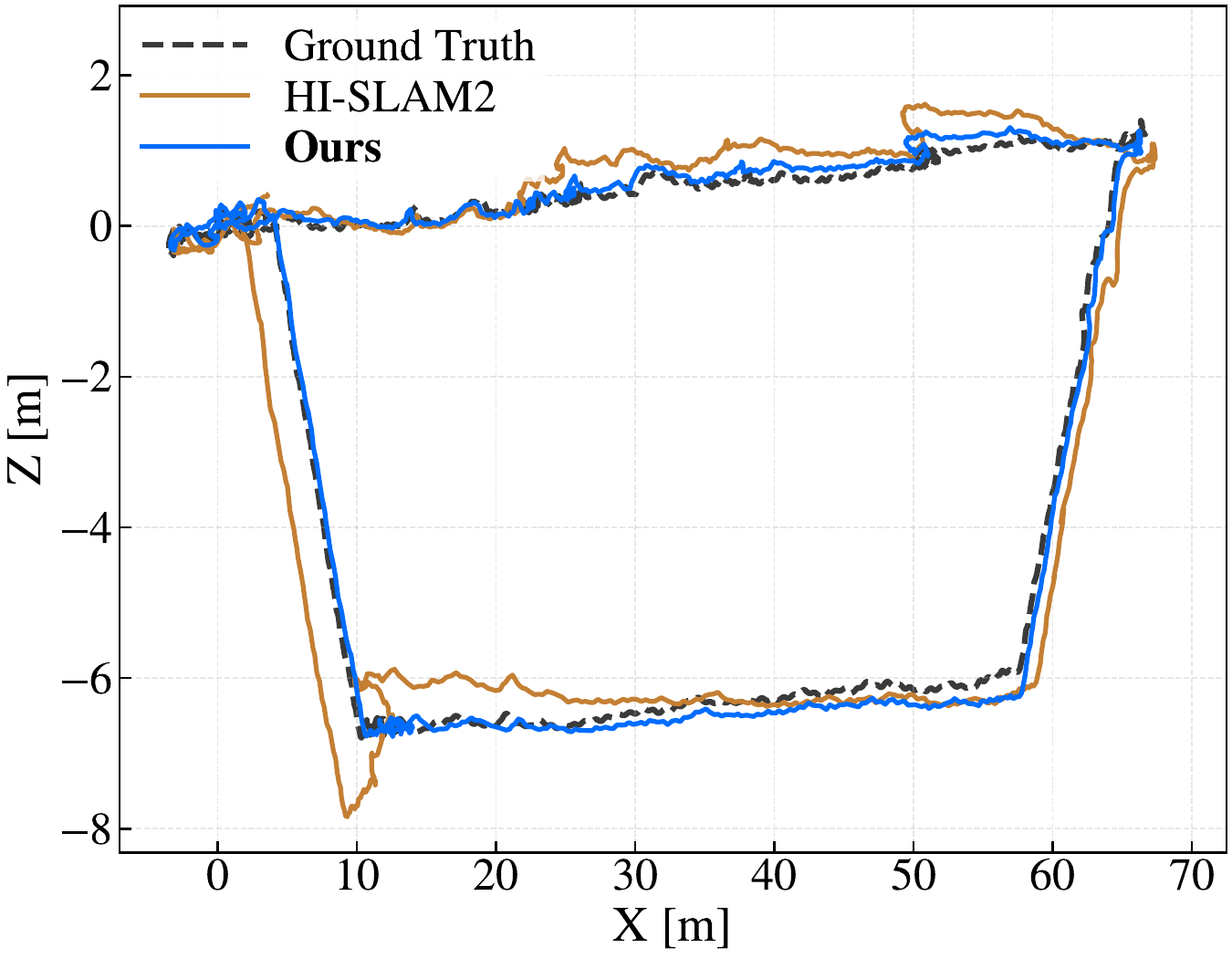} \\
& \hspace{12pt}\texttt{Campus1} & \hspace{10pt}\texttt{Corridor1} & \hspace{10pt}\texttt{Downtown3} \\
\end{tabular}
\vspace{-3mm}
\caption{\textbf{Sample Input Frames and Trajectory Plots on the Self-Captured Dataset.} Representative frames illustrate the various challenges of our dataset.}
\label{fig:traj_odin}
\vspace{-2mm}
\end{figure*}

\subsection{Tracking Robustness}
\label{sec:robust}
\paragraph{Strided Evaluation}
To evaluate robustness under degraded visual input, we create \emph{strided} variants of the EuRoC~\cite{burri2016euroc} and RPNG AR Table~\cite{Chen2023rpng} datasets by temporally subsampling RGB frames with different strides while keeping the original IMU readings, simulating frame drops, limited bandwidth and high-speed motion. 
We report average Recall@5cm and Recall@10cm instead of ATE RMSE, because several methods fail on certain sequences, making it difficult to compute meaningful ATE averages. 
We assign a recall score of zero when a method fails to produce a valid trajectory. 
As shown in \figref{fig:tracking_stride_avg}, our method remains stable and consistently outperforms all baselines, whereas others frequently fail to initialize or lose tracking under large inter-frame gaps.

\paragraph{Self-Captured Dataset} We further capture 18 challenging sequences spanning (i) appearance degradation, including motion blur, exposure changes, sun glare, and low-light conditions; (ii) geometric ambiguity, such as low texture, repetitive patterns, and reflective or transparent surfaces; (iii) interference caused by dynamic objects; (iv) aggressive camera motion with hand-held shake; and (v) long trajectories.
As shown in \figref{fig:ate_step_curve} and \figref{fig:traj_odin}, baseline methods suffer from significant tracking drift, while our method remains accurate and robust despite these severe real-world challenges. Please refer to the supplementary material for more results and dataset details.

\begin{table}[t]
\centering
\scriptsize
\setlength{\tabcolsep}{4.0pt}   %
\renewcommand{\arraystretch}{0.95} %
\caption{\textbf{Tracking Ablation on EuRoC (stride = 10)~\cite{burri2016euroc} Dataset.}}
\begin{tabular}{lrr}
\toprule
        & {\makecell{Avg. ATE RMSE [cm] $\downarrow$}} & {\makecell{Avg. Recall @ 10cm [\%] $\uparrow$}} \\
\midrule
\makebox[1.6em][l]{(a)} w/o IMU Bias Estimation & 338.99 & 0.05 \\
\makebox[1.6em][l]{(b)} w/o IMU Fusion    & 88.54 & 32.50 \\
\makebox[1.6em][l]{(c)} w/o 3-Staged IMU Initialization & 50.65 & 68.32 \\ 
\makebox[1.6em][l]{(d)} w/o KF Pose Initialization with IMU & 40.94 & 64.95 \\
\makebox[1.6em][l]{(e)} w/o Loop Closure & 23.78 & 40.63 \\
\makebox[1.6em][l]{(f)} w/o Per-KF Bias Estimation & 8.83 & 90.01 \\ %
\textbf{Ours (Full System)} & 
\textbf{3.39} &
\textbf{98.99} \\
\bottomrule
\end{tabular}

\vspace{-0mm}
\label{tab:tracking_ablation_avg}
\end{table}

\begin{table}[t]
    \centering
    \scriptsize
    \setlength{\tabcolsep}{12.5pt}
    \renewcommand{\arraystretch}{1}
    \caption{\textbf{Rendering Ablation on FAST-LIVO2 Dataset~\cite{zheng2024fast}}.}
    \begin{tabular}{lccc}
        \toprule
        Method & Avg. PSNR $\uparrow$ & Avg. SSIM $\uparrow$ & Avg. LPIPS $\downarrow$ \\
        \midrule
        w/o Loop Closure GS Update & 22.06 & 0.677 & 0.541 \\
        \textbf{Ours (Full System)}                    & \textbf{23.15} & \textbf{0.729} & \textbf{0.487} \\
        \bottomrule
    \end{tabular}
    \label{tab:mapping_ablation_avg}
    \vspace{-2mm}
\end{table}

\subsection{Ablation Study}
\label{sec:ablation}
\paragraph{Tracking} We report ablation results for 6 design choices in \tabref{tab:tracking_ablation_avg}.
In (a) \textit{w/o IMU Bias Estimation}, the IMU bias is fixed to zero and not optimized.
In (b) \textit{w/o IMU Fusion}, the optimization is constrained solely by visual residuals.
In (c), we remove the inertial-only optimization stage during IMU initialization.
In (d) \textit{w/o KF Pose Initialization with IMU}, we disable the IMU pre-integration-based pose initialization for newly selected keyframes.
In (e), we disable the loop closure.
In (f), we optimize a single global IMU bias shared across all keyframes instead of estimating per-keyframe biases.
\tabref{tab:tracking_ablation_avg} demonstrates that removing any component will degrade tracking accuracy as well as robustness, while our full system achieves the best results.

\paragraph{Mapping} We ablate the loop closure Gaussian update in \tabref{tab:mapping_ablation_avg}. Disabling this update leads to inconsistent Gaussian maps, which in turn leads to rendering performance degradation.

\section{Conclusion}
\label{sec:conclusions}
We present VIGS-SLAM, a novel visual-inertial 3D Gaussian Splatting SLAM system that achieves robust real-time tracking and high-fidelity Gaussian mapping. It tightly couples learning-based dense visual correspondences with inertial constraints within a unified optimization framework. Furthermore, we perform robust IMU initialization and efficient loop closure with consistent Gaussian updates. Extensive evaluations on five challenging datasets show that VIGS-SLAM achieves state-of-the-art performance in both tracking accuracy and novel view synthesis quality, and is among the very few methods that succeed on all sequences without failure. Code and dataset will be made public.

\vspace{1pt} \noindent
\textbf{Limitation.} 
In the current system, the Gaussian map does not directly improve tracking. 
Incorporating a Gaussian re-rendering loss for pose optimization yields only marginal gains, likely because renderings during online optimization are not yet sufficiently sharp to produce reliable gradients for pose refinement.

\clearpage

\setcounter{page}{1}
\title{Supplementary Material for VIGS-SLAM: \\ Visual-Inertial Gaussian Splatting SLAM}
\titlerunning{VIGS-SLAM}

\author{
Zihan Zhu\inst{1} \and
Wei Zhang\inst{2} \and
Moyang Li\inst{1} \and
Norbert Haala\inst{2} \and \\
Marc Pollefeys\inst{1,3} \and
Daniel Barath\inst{1}
}

\authorrunning{Zihan Zhu et al.}

\institute{
ETH Zurich, Zurich, Switzerland \and
University of Stuttgart, Stuttgart, Germany \and
Microsoft
}
\maketitle

\setcounter{figure}{6}
\setcounter{table}{7}
\setcounter{section}{5}
\begin{table*}[t]
\centering
\footnotesize
\renewcommand{\arraystretch}{1.4}
\setlength{\tabcolsep}{2pt}
\caption{\textbf{Dataset Statistics.} Detailed statistics of all datasets used in our evaluation, along with the characteristic challenges.}
\resizebox{\textwidth}{!}{
\begin{tabular}{lccccc}
\toprule
& \textbf{EuRoC}~\cite{burri2016euroc} 
& \textbf{RPNG AR Table}~\cite{Chen2023rpng}
& \textbf{UTMM}~\cite{sun2024mm3dgs}
& \textbf{FAST-LIVO2}~\cite{zheng2024fast} & \textbf{Self-Captured}\\
\midrule

\textbf{Year} 
& 2016 & 2023 & 2024 & 2024 & 2026\\
\midrule
\textbf{\# Sequences} 
& 11 & 8 & 8 & 5 & 18\\
\midrule
\textbf{Avg. \# Frames} 
& 2459 & 4521 & 763 & 1305 & 1949 \\
\midrule
\textbf{RGB / Gray} 
& Gray & RGB & RGB & RGB & RGB\\
\midrule
\textbf{Resolution} 
& $752{\times}480$ 
& $848{\times}480$ 
& $1280{\times}660$ 
& $1280{\times}1024$
& $1600{\times}1296$ \\
\midrule
\textbf{RGB FPS} 
& 20 & 30 & 30 & 10 & 10 \\
\midrule
\textbf{IMU Rate (Hz)} 
& 200 & 400 & 100 & 200 & 400\\
\midrule
\textbf{GT Source} 
& MoCap & MoCap & MoCap & FAST-LIVO2~\cite{zheng2024fast} & CloudMind~\cite{manifold_mindcloud} \\
\midrule
\textbf{Environment} 
& Indoor (industrial) 
& Indoor (tabletop) 
& Indoor (large open hall) 
& Outdoor & Indoor \& Outdoor\\
\midrule
\textbf{Challenges} 
& \makecell[c]{low-texture \\ motion blur \\ illumination changes}
& \makecell[c]{close-range geometry \\ high-frequency texture}
& \makecell[c]{motion blur \\ distant cluttered background}
& \makecell[c]{long trajectory \\ reflective surfaces \\ illumination changes}
& \makecell[c]{low-texture \\ illumination changes \\ motion blur \\ dynamic objects \\ long trajectory \\ sun glare \\ aggressive motion \\ reflective/transparent surfaces \\ repetitive patterns}
\\
\bottomrule
\end{tabular}
}

\label{tab:dataset_stats}
\end{table*}

\begin{abstract}
In the supplementary material, we provide additional details about the following:
\begin{enumerate}
    \item Implementation details of the real-time demo in the supplementary material (\secref{sec:demo}).
    \item More information about the datasets used in evaluation (\secref{sec:dataset}).
    \item Implementation details of Baseline Methods (\secref{sec:baseline_implementation}).
    \item Implementation details of VIGS-SLAM (\secref{sec:implementation}).
    \item Additional results (\secref{sec:add_results}).

\end{enumerate}
We additionally include \textbf{supplementary videos} where we show additional visual results. 
\end{abstract}
\section{Real-time Demo Implementation Details}
\label{sec:demo}
We include a real-time demo in the supplementary videos, where an iPhone 17 Pro streams RGB frames and IMU measurements to a desktop computer equipped with an Intel(R) Core(TM) i7-14700K CPU and an NVIDIA GeForce RTX 5090 GPU. Our VIGS-SLAM system robustly tracks camera motion while simultaneously reconstructing a photorealistic Gaussian map and dense point clouds in real time. The iPhone capture application is implemented in Swift, and data transmission is performed over Wi-Fi. The camera intrinsics remain fixed during capture.

\section{Dataset Details}
\label{sec:dataset}
\paragraph{Dataset Statistics} In \tabref{tab:dataset_stats}, we report comprehensive dataset statistics and outline the specific challenges associated with each dataset. 

\paragraph{Self-Captured Dataset}
For the self-captured dataset described in \secrefn{sec:robust}, we use the Manifold Odin 1~\cite{manifoldtech2025odin1} for data acquisition. RGB frames are captured at 10 FPS, while IMU measurements are recorded at 400 Hz. The camera intrinsics and camera-IMU extrinsics are factory pre-calibrated. The original images are captured with a fisheye lens; we undistort them to a pinhole model before feeding them to all methods. The accompanying offline processing tool, MindCloud~\cite{manifold_mindcloud}, provides ground-truth poses via LiDAR-visual-inertial fusion. In total, we capture 18 sequences across diverse indoor and outdoor environments, with trajectory lengths ranging from 34 to 1079 meters and sequence durations from 47 seconds to 10 minutes. The input sequences are visualized in the supplementary video. We use Deface~\cite{orbhd_deface} to blur human faces and EgoBlur~\cite{raina2023egoblur} to blur vehicle license plates.

\section{Baseline Implementation Details}
\label{sec:baseline_implementation}

For DBA-Fusion~\cite{zhou2024dba} and its successor VINGS-Mono~\cite{wu2025vings}, we worked closely with the first author of VINGS-Mono~\cite{wu2025vings} and applied several targeted modifications to improve their stability and performance.

\subsection{DBA-Fusion~\cite{zhou2024dba}}
On the EuRoC dataset~\cite{burri2016euroc}, the drone often remains static or exhibits only small motions at the beginning of a sequence. In this scenario, DBA-Fusion's vision-only keyframe selection tends to cause severe drift due to long IMU preintegration intervals. To mitigate this issue, we enforce the insertion of a new keyframe at least every 20 frames, even when the frontend motion filter does not trigger keyframe selection. This strategy prevents excessively long IMU preintegration intervals and reduces accumulated noise.

For the UTMM dataset~\cite{sun2024mm3dgs}, some sequences do not reach the default threshold required for IMU initialization throughout the entire sequence, resulting in no IMU initialization being performed. To address this issue, we lower the IMU initialization threshold to 0.15 for those sequences.

\subsection{VINGS-Mono~\cite{wu2025vings}}
For VINGS-Mono, we apply the same modifications as for DBA-Fusion on the EuRoC and UTMM datasets~\cite{burri2016euroc, sun2024mm3dgs}. In addition, we observe that disabling monocular metric depth and loop closure yields more robust results. We suspect this is because the metric depth model does not generalize well to our datasets, and the loop closure module occasionally produces false positives or estimates inaccurate relative poses for loop closure frames. Therefore, we disable these components in all experiments.

\section{Our VIGS-SLAM Implementation Details}
\label{sec:implementation}
Except for the number of keyframes used in IMU initialization $N^{\mathrm{iner}}_{\mathrm{init}}$, all datasets share the same hyperparameters.

\paragraph{Frontend Tracking}
A new keyframe is created when the average optical-flow magnitude exceeds $\tau_{\text{kf}} = 2.4$. To reduce IMU pre-integration drift, we also enforce a keyframe at least every $t_{\text{kf}} = 3$ s. For each new keyframe, we initialize its pose using IMU pre-integration unless the estimated uncertainty is too high. When the covariance trace $\mathrm{tr}(\Sigma_{ij}^{\text{iner}})$ exceeds $\tau_{\Sigma}^{\text{init}} = 10^{-4}$, we revert to initializing the new keyframe with the pose of the previous keyframe.

\paragraph{Loop Closure}
Loop-closure detection runs in parallel with the frontend tracking. We only compare the current keyframe against earlier keyframes that are at least $\tau_{\text{LC-gap}} = 55$ keyframes apart to avoid redundant loop candidates. A loop-closure edge is added when the average optical-flow magnitude falls below $\tau_{\text{LC-flow}} = 22$ and the relative orientation difference is below $\tau_{\text{LC-ang}} = 120^\circ$.

\paragraph{IMU Initialization}
Pure-vision initialization is performed once the number of keyframes reaches $N^{\mathrm{vis}}_{\mathrm{init}} = 10$. Inertial initialization begins when the keyframe count reaches $N^{\mathrm{iner}}_{\mathrm{init}}$, which is set to 20 by default (25 for FAST\text{-}LIVO2 due to rapid motion and 15 for UTMM due to short trajectory).

After completing the vision-only initialization, we begin inertial-only initialization by first recovering the gravity direction $\bm{R}_{\mathrm{wg}}$. Then we aim to recover IMU parameters as well as convert the trajectory to metric scale by introducing a global log-scale parameter $s \in \mathbb{R}$. The positions and velocities are then rewritten as
\begin{align}
\bm{p}_i &= e^{s}\,\bm{p}_i^{\text{vis}}, 
&
\bm{v}_i &= e^{s}\,\bm{v}_i^{\text{vis}}, \\
\bm{p}_j &= e^{s}\,\bm{p}_j^{\text{vis}}, 
&
\bm{v}_j &= e^{s}\,\bm{v}_j^{\text{vis}}.
\end{align}
During this stage, the visual poses
$\bm{T}_i^{\text{vis}} = (\bm{R}_i,\bm{p}_i)$ are kept fixed, and we solve for the gravity direction
$\bm{R}_{\mathrm{wg}}$, IMU bias $\bm{b}_i$, per-keyframe velocity 
$\bm{v}_i$, and the scale parameter $s$.
Finally, we perform full visual–inertial initialization by incorporating the camera poses into the optimization as well.

\paragraph{Final Global Bundle Adjustment}
Following prior works~\cite{teed2021droid,zheng2025wildgs,sandstrom2024splat,zhang2024hislam2}, we optionally perform a final global bundle adjustment step to refine all poses. We adopt the same global frame-graph construction as prior works and optimize only vision residuals, as inertial constraints mainly benefit initialization and coarse pose estimation, but offer limited improvement during the refinement stage. Unless otherwise specified, all reported results reflect the online performance \emph{before} this final refinement.

\paragraph{Final Color Refinement}
Similar to prior works~\cite{zheng2025wildgs,sandstrom2024splat,zhang2024hislam2}, after the final global bundle adjustment, we optionally perform a global refinement of the Gaussian map using all keyframes. Specifically, in each iteration we randomly sample a keyframe from the global keyframe list and minimize the rendering loss in \eqnrefn{eq:render} to update the Gaussian map. Similar to the final global bundle adjustment, all rendering metrics we report are taken \emph{before} this refinement step, unless explicitly stated otherwise.

\begin{table*}[t]
\centering
\footnotesize
\setlength{\tabcolsep}{1.7pt}
\renewcommand{\arraystretch}{1.05}
\caption{\textbf{Runtime Evaluation on the RPNG~\cite{Chen2023rpng} dataset.} We report runtime (FPS) and GPU memory usage for pure tracking and for the full system (tracking + Gaussian mapping).
All results are measured on an Intel(R) Core(TM) i7-14700K CPU and an NVIDIA GeForce RTX 5090 GPU. HI-SLAM2 fails on \texttt{table\_05}, `F' denotes failure, `*' denote taking the average of all other sequences.}
\resizebox{\textwidth}{!}{
\begin{tabular}{llrrrrrrrr
    !{\smash{\tikz[baseline]{\draw[densely dashed, gray!80, line width=0.8pt] (0pt,-2pt)--(0pt,8pt);}}}
r}
\toprule
& & \texttt{table\_01} & \texttt{table\_02} & \texttt{table\_03} & \texttt{table\_04} &
  \texttt{table\_05} & \texttt{table\_06} & \texttt{table\_07} & \texttt{table\_08} & \textbf{Avg.}\\
\midrule
& \#Frames        & 2506 & 2914 & 7006 & 6068 & 6164 & 2767 & 4784 & 8484 & 5087\\
\hdashline

\multirow{3}{*}{ORB-SLAM3}
 & \#Keyframes                & 211  & 389  & 416  & 280  & 250  & 238  & 184  & 227 & 274 \\
 & Tracking [FPS]             & 23.40 & 19.81 & 19.51 & 23.11 & 22.30 & 22.86 & 22.72 & 22.47 & 22.02 \\
 &  + GS Mapping [FPS]& -- & -- & -- & -- & -- & -- & -- & -- & -- \\
\hdashline

\multirow{3}{*}{HI-SLAM2}
 & \#Keyframes                & 252  & 368  & 535  & 424  & F  & 239  & 220  & 564 & *372\\
 & Tracking [FPS]             & 28.04 & 23.50 & 29.58 & 32.08 & F & 27.60 & 50.28 & 22.92 & *30.57 \\
 &  + GS Mapping [FPS]& 9.94 & 8.44 & 11.82 & 12.53 & F & 11.57 & 20.99 & 12.41 & *12.53\\

 \hdashline
  \multirow{5}{*}{\parbox{2.2cm}{ \textbf{VIGS-SLAM\\(Ours)}}}
   & \#Keyframes                & 250 & 349 & 525 & 427 & 341 & 232 & 223 & 579 & 366 \\
   & Tracking [FPS] & 36.73 & 32.47 &  43.30 & 43.30 & 32.44 & 34.02 & 66.37 & 30.03 & 39.83 \\
   & Tracking GPU Mem [GiB] & 7.48 & 7.71 & 8.15 & 8.98 & 8.81 & 8.42 & 7.42 & 9.73 & 8.34 \\
  & +Mapping [FPS] & 9.11 & 7.82 & 10.73 & 11.25 & 13.79 & 11.04 & 20.22 & 12.18 & 12.02 \\
  & +Mapping GPU Mem [GiB] & 7.64 & 7.88 & 8.64 & 9.16 & 8.94 & 8.36 & 7.56 & 9.88 & 8.51  \\
\bottomrule
\end{tabular}
}

\label{tab:rpng_runtime}
\end{table*}

\section{Additional Experiments}
\label{sec:add_results}
\paragraph{Runtime and Memory Analysis}
We report the runtime evaluation on the RPNG dataset in \tabref{tab:rpng_runtime}. All experiments are conducted on a desktop equipped with an Intel(R) Core(TM) i7-14700K CPU and an NVIDIA GeForce RTX 5090 GPU. FPS is computed as the total number of frames divided by the total runtime in seconds. GPU memory reports the peak GPU memory usage, indicating suitability for deployment on common robotic platforms such as Jetson Orin NX and Jetson AGX Orin.

To accelerate runtime, we identified several bottlenecks in the underlying codebases we build upon, DROID-SLAM~\cite{teed2021droid} and HI-SLAM2~\cite{zhang2024hislam2}, such as inefficient Python for-loops, and introduced targeted optimizations to address them. In addition, we deploy TensorRT~\cite{tensorRT} for neural network inference, implement IMU preintegration in C++, and perform Jacobian and Hessian computations using custom CUDA kernels.

\begin{table}[t]
\centering
\setlength{\tabcolsep}{3pt}
\caption{\textbf{Tracking Performance on the Self-Captured Dataset} (ATE RMSE $\downarrow$ [m]). 
Best and second-best results are highlighted as \colorbox{colorFst}{\bf first} and \colorbox{colorSnd}{second}. `F' indicates failure. We also report the trajectory length of each sequence for reference.}
\resizebox{\textwidth}{!}{
\begin{tabular}{lrrrrrr}
\toprule
Sequence
& Seq. Len. [m]
& HI-SLAM2~\cite{zhang2024hislam2}
& DROID-SLAM~\cite{teed2021droid}
& VINGS-Mono~\cite{wu2025vings}
& DBA-Fusion~\cite{zhou2024dba}
& \textbf{VIGS-SLAM (Ours)} \\
\midrule

{\tt Basement1}  & 34.590   & 0.749   & 0.819   & 0.132   & \nd 0.102   & \fs 0.078 \\
{\tt Basement2}  & 79.250   & 1.452   & 1.208   & 1.389   & \fs 0.366   & \nd 0.430 \\
{\tt Basement3}  & 53.710   & 0.834   & 0.749   & 0.321   & \nd 0.240   & \fs 0.076 \\
{\tt Basement4}  & 34.350   & 1.142   & 0.786   & 0.136   & \fs 0.117   & \nd 0.132 \\

{\tt Campus1}    & 499.010  & 19.328  & 21.348  & \nd 5.830 & 7.902      & \fs 0.631 \\
{\tt Campus2}    & 1079.650 & 127.437 & 54.719  & 25.449   & \nd 6.371   & \fs 4.831 \\

{\tt Corridor1}  & 518.050  & 25.269  & 35.621  & 7.148    & \nd 4.318   & \fs 1.454 \\
{\tt Corridor2}  & 504.620  & 16.782  & 9.608   & \nd 2.342 & 2.666      & \fs 0.532 \\

{\tt Downtown1}  & 61.320   & \nd 0.073 & 0.109 & 0.184    & 0.100       & \fs 0.065 \\
{\tt Downtown2}  & 139.890  & 0.154   & 0.482   & 0.487    & \nd 0.148   & \fs 0.131 \\
{\tt Downtown3}  & 278.180  & \nd 3.980 & 7.508 & 3.067    & 8.726       & \fs 0.797 \\

{\tt Ferrari1}   & 253.640  & 1.384   & 2.917   & \nd 0.697 & 0.725      & \fs 0.636 \\
{\tt Ferrari2}   & 91.450   & 0.608   & 7.231   & 0.633    & \nd 0.192   & \fs 0.134 \\

{\tt Graffiti1}  & 50.330   & \nd 0.063 & 0.065 & 0.190    & 0.080       & \fs 0.059 \\
{\tt Graffiti2}  & 70.810   & \nd 0.275 & 3.394 & 0.511    & 0.295       & \fs 0.097 \\
{\tt Graffiti3}  & 63.400   & 5.178   & 2.195   & 0.396    & \fs 0.344   & \nd 0.366 \\

{\tt Motorworld} & 100.990  & \nd 0.075 & 0.695 & 0.327    & 0.430       & \fs 0.072 \\

{\tt Office}     & 96.850   & F       & 0.260   & 0.520    & \nd 0.172   & \fs 0.061 \\

\hdashline
\textbf{Avg.} & 222.783 & 12.046 & 8.318 & 2.764 & \nd 1.850 & \fs 0.588 \\

\bottomrule
\end{tabular}}
\label{tab:tracking_odin}
\end{table}
\paragraph{Per-sequence Tracking Results on the Self-Captured Dataset}
As shown in \tabref{tab:tracking_odin}, pure visual methods such as DROID-SLAM~\cite{teed2021droid} and HI-SLAM2~\cite{zhang2024hislam2} suffer from severe drift under challenging conditions. Although VINGS-Mono~\cite{wu2025vings} and DBA-Fusion~\cite{zhou2024dba} incorporate IMU signals, their interleaved optimization strategies still limit robustness. In contrast, our VIGS-SLAM achieves the best performance, benefiting from the proposed tightly coupled visual-inertial joint optimization, together with robust IMU initialization and effective loop closure.

\paragraph{Simulation of Various Visual Degradation}  %
To evaluate our methods robustness under progressively degraded visual input, we apply controlled photometric corruptions to the RGB input frames. 
\noindent\textbf{Motion Blur} is simulated by convolving each image with a linear motion kernel of fixed length $k$ pixels, where the blur direction is independently sampled per frame from a uniform distribution over $[0, 2\pi)$. Given an input image $I \in [0,1]$, the resulting image is given by
\begin{equation}
I_{\text{blur}} = I * K(k, \theta), \quad \theta \sim \mathcal{U}(0, 2\pi),
\end{equation}
where $K(k, \theta)$ denotes a motion blur kernel with strength $k$ and direction $\theta$. Larger kernel sizes correspond to more severe motion blur.
\textbf{Overexposure} is modeled by amplifying image intensities followed by highlight clipping:
\begin{equation}
I_{\text{over}} = \mathrm{clip}(\beta I, 0, 1),
\end{equation}
where $\beta > 1$ controls the exposure level and progressively saturates bright regions, significantly reducing texture and local contrast.
\textbf{Low-light conditions} are simulated by first reducing image exposure and contrast using a gamma transformation, followed by the injection of realistic sensor noise. We apply
\begin{equation}
I_{\gamma} = I^{\gamma}, \quad \gamma > 1,
\end{equation}
which models reduced illumination and nonlinear camera response under low-light conditions. We then add signal-dependent shot noise and additive Gaussian read noise:
\begin{equation}
I_{\text{low}} =
\mathrm{clip}\!\left(
\frac{\mathrm{Poisson}(s \cdot I_{\gamma})}{s}
+ \mathcal{N}(0,\sigma^2),
\,0,\,1
\right),
\end{equation}
 where $s$ denotes the photon scale and $\sigma$ is the standard deviation of Gaussian noise. 
This formulation jointly reduces visual contrast and amplifies noise in a physically plausible manner.
As shown in \tabref{tab:tracking_euroc_motion_blur}, \tabref{tab:tracking_euroc_overexposure} and \tabref{tab:tracking_euroc_lowlighting}, our method remains robust under various challenges thanks to our tightly coupled vision-inertial fusion. 

\begin{table*}[h]
\vspace{-2mm}
\centering
\footnotesize
\setlength{\tabcolsep}{6.0pt}
\caption{\textbf{Tracking Performance on Motion Blurred  EuRoC Dataset~\cite{burri2016euroc}} (ATE RMSE $\downarrow$ [cm]). }
\resizebox{\textwidth}{!}{
\begin{tabular}{
    lrrrrrrrrrrr
    !{\smash{\tikz[baseline]{\draw[densely dashed, gray!80, line width=0.8pt] (0pt,-2pt)--(0pt,8pt);}}}
    r
}
\toprule
Method & \tt{MH\_01} & \tt{MH\_02} & \tt{MH\_03} & \tt{MH\_04} & \tt{MH\_05} & \tt{V1\_01} & \tt{V1\_02} & \tt{V1\_03} & \tt{V2\_01} & \tt{V2\_02} & \tt{V2\_03} & \textbf{Avg.} \\
\midrule
\multicolumn{13}{l}{\cellcolor[HTML]{EEEEEE}{\textit{Motion Blur Strength $k = 5$}}} \\
HI-SLAM2~\cite{zhang2024hislam2}      
& 1.72 & 1.62 & 2.74 & 6.56 & 7.18 & \textbf{3.53} & 1.51 & 2.53 & 2.94 & 2.68 & 2.23 & 3.20 \\

\textbf{VIGS-SLAM (Ours)}                          
& \textbf{1.32} & \textbf{1.24} & \textbf{2.66} & \textbf{6.48} & \textbf{4.35} & 3.58 & \textbf{1.14} & \textbf{2.29} & \textbf{2.03} & \textbf{1.69} & \textbf{2.08} & \textbf{2.62} \\

\multicolumn{13}{l}{\cellcolor[HTML]{EEEEEE}{\textit{Motion Blur Strength  $k = 10$}}} \\
HI-SLAM2~\cite{zhang2024hislam2}      
& 2.14 & \textbf{2.18} & \textbf{2.79} & 36.63 & 16.24 & 3.63 & 2.03 & \textbf{2.48} & 2.56 & 2.12 & 3.10 & 6.90 \\

\textbf{VIGS-SLAM (Ours)}                         
& \textbf{1.27} & 2.80 & 2.85 & \textbf{6.76} & \textbf{4.40} & \textbf{3.51} & \textbf{1.26} & 2.86 & \textbf{1.57} & \textbf{1.93} & \textbf{2.60} & \textbf{2.89} \\
\bottomrule
\end{tabular}
}

\label{tab:tracking_euroc_motion_blur}
\end{table*}

\begin{table*}[h]
\centering
\footnotesize
\caption{\textbf{Tracking Performance on the EuRoC Dataset~\cite{burri2016euroc} with Over-Exposure} (ATE RMSE $\downarrow$ [cm]).}
\setlength{\tabcolsep}{6.0pt}
\resizebox{\textwidth}{!}{
\begin{tabular}{
    lrrrrrrrrrrr
    !{\smash{\tikz[baseline]{\draw[densely dashed, gray!80, line width=0.8pt] (0pt,-2pt)--(0pt,8pt);}}}
    r
}
\toprule
Method & \tt{MH\_01} & \tt{MH\_02} & \tt{MH\_03} & \tt{MH\_04} & \tt{MH\_05} & \tt{V1\_01} & \tt{V1\_02} & \tt{V1\_03} & \tt{V2\_01} & \tt{V2\_02} & \tt{V2\_03} & \textbf{Avg.} \\
\midrule
\multicolumn{13}{l}{\cellcolor[HTML]{EEEEEE}{\textit{$\beta$ = 2}}} \\
HI-SLAM2~\cite{zhang2024hislam2}     
& 2.29 & \textbf{1.41} & 2.88 & 13.88 & 10.25 & 3.65 & 2.10 & 2.61 & 2.75 & 1.95 & \textbf{2.48} & 4.21 \\

\textbf{VIGS-SLAM (Ours)}                     
& \textbf{1.10} & 2.20 & \textbf{2.62} & \textbf{4.89} & \textbf{5.32} & \textbf{3.60} & \textbf{1.23} & \textbf{2.26} & \textbf{2.45} & \textbf{1.70} & 2.85 & \textbf{2.75} \\

\multicolumn{13}{l}{\cellcolor[HTML]{EEEEEE}{\textit{$\beta$ = 3}}} \\
HI-SLAM2~\cite{zhang2024hislam2}       
& 1.76 & 1.79 & \textbf{2.93} & 260.88 & 11.03 & 4.26 & 37.77 & 3.10 & 3.11 & 3.25 & 3.75 & 30.33 \\

\textbf{VIGS-SLAM (Ours)}    
& \textbf{1.17} & \textbf{1.49} & 3.06 & \textbf{4.81} & \textbf{7.18} & \textbf{3.54} & \textbf{4.41} & \textbf{2.39} & \textbf{2.20} & \textbf{2.49} & \textbf{2.64} & \textbf{3.22} \\

\multicolumn{13}{l}{\cellcolor[HTML]{EEEEEE}{\textit{$\beta$ = 4}}} \\
HI-SLAM2~\cite{zhang2024hislam2}   
& 1.75 & 1.95 & 3.61 & 23.03 & F & 6.26 & 140.40 & 19.24 & 4.36 & 5.49 & 6.18 & 21.23* \\

\textbf{VIGS-SLAM (Ours)} 
& \textbf{1.63} & \textbf{1.24} & \textbf{2.83} & \textbf{11.52} & \textbf{10.14} & \textbf{3.82} & \textbf{5.42} & \textbf{9.79} & \textbf{2.66} & \textbf{3.25} & \textbf{3.72} & \textbf{5.09} \\
\bottomrule
\end{tabular}
}

\label{tab:tracking_euroc_overexposure}
\end{table*}

\begin{table*}[h]
\vspace{-2mm}
\centering
\footnotesize
\caption{\textbf{Tracking Performance on the Low-Lighting  EuRoC Dataset~\cite{burri2016euroc}} (ATE RMSE $\downarrow$ [cm]).  }
\setlength{\tabcolsep}{6.0pt}
\resizebox{\textwidth}{!}{
\begin{tabular}{
    lrrrrrrrrrrr
    !{\smash{\tikz[baseline]{\draw[densely dashed, gray!80, line width=0.8pt] (0pt,-2pt)--(0pt,8pt);}}}
    r
}
\toprule
Method & \tt{MH\_01} & \tt{MH\_02} & \tt{MH\_03} & \tt{MH\_04} & \tt{MH\_05} & \tt{V1\_01} & \tt{V1\_02} & \tt{V1\_03} & \tt{V2\_01} & \tt{V2\_02} & \tt{V2\_03} & \textbf{Avg.} \\
\midrule

\multicolumn{13}{l}{\cellcolor[HTML]{EEEEEE}{\textit{$\gamma=3$, $\sigma_{noise} = 0.05$, $s = 200$ }}} \\
HI-SLAM2~\cite{zhang2024hislam2}   
& 2.11 & 1.48 & 2.69 & 48.62 & 184.65 & \textbf{3.45} & 1.61 & \textbf{4.22} & 2.82 & 2.66 & 3.90 & 23.47 \\

\textbf{VIGS-SLAM (Ours)}  
& \textbf{1.34} & \textbf{1.40} & \textbf{2.58} & \textbf{13.59} & \textbf{4.37} & \textbf{3.45} & \textbf{1.18} & 5.26 & \textbf{1.83} & \textbf{1.77} & \textbf{3.67} & \textbf{3.68} \\
\multicolumn{13}{l}{\cellcolor[HTML]{EEEEEE}{\textit{$\gamma=4$, $\sigma_{noise} = 0.05$, $s = 200$ }}} \\
HI-SLAM2~\cite{zhang2024hislam2}
& 2.75 & \textbf{1.83} & 3.04 & 89.81 & 23.21 & \textbf{3.44} & 1.54 & 32.12 & 2.54 & 2.36 & \textbf{4.17} & 15.16 \\

\textbf{VIGS-SLAM (Ours)}
& \textbf{1.68} & 2.17 & \textbf{2.65} & \textbf{14.84} & \textbf{5.66} & \textbf{3.44} & \textbf{1.24} & \textbf{14.75} & \textbf{1.82} & \textbf{1.82} & 6.15 & \textbf{5.11} \\
\bottomrule
\end{tabular}
}

\label{tab:tracking_euroc_lowlighting}
\end{table*}

\begin{table*}[h]
    \centering
    \setlength{\tabcolsep}{4pt}
    \footnotesize
        \caption{\textbf{Tracking Performance on EuRoC Dataset~\cite{burri2016euroc}} (ATE RMSE $\downarrow$ [cm]).}
    \resizebox{\textwidth}{!}{
    \begin{tabular}{lrrrrrrrrrrr
    !{\smash{\tikz[baseline]{\draw[densely dashed, gray!80, line width=0.8pt] (0pt,-2pt)--(0pt,8pt);}}}
    r}
        \toprule
        Method & \tt{MH\_01} & \tt{MH\_02} & \tt{MH\_03} & \tt{MH\_04} & \tt{MH\_05} &
        \tt{V1\_01} & \tt{V1\_02} & \tt{V1\_03} &
        \tt{V2\_01} & \tt{V2\_02} & \tt{V2\_03} & \textbf{Avg.} \\
        \midrule

        \multicolumn{13}{l}{\cellcolor[HTML]{EEEEEE}{\textit{Without Final BA}}} \\
        
        Splat-SLAM~\cite{sandstrom2024splat}
        & 257.64 & 266.02 & 312.58 & 458.14 & 360.86
        & 168.99 & 166.65 & 128.67
        & 198.84 & 195.85 & 190.86 & 245.01 \\
        
        DROID-SLAM~\cite{teed2021droid}
        & 16.30 & 12.10 & 24.20 & 39.90 & 27.00
        & 10.30 & 16.50 & 15.80
        & 10.20 & 11.50 & 20.40 & 18.60 \\

        HI-SLAM2~\cite{zhang2024hislam2}
        & 2.66 & 1.44 & 2.71 & 6.86 & \textbf{5.07}
        & \textbf{3.55} & 1.32 & \textbf{2.49}
        & 2.56 & 1.77 & \textbf{1.92} & 2.94 \\

        \textbf{VIGS-SLAM (Ours)}
        & \textbf{1.42} & \textbf{1.29} & \textbf{2.55} & \textbf{5.16} & 5.64
        & 3.67 & \textbf{1.15} & 2.68
        & \textbf{2.34} & \textbf{1.53} & 3.27 & \textbf{2.79} \\

        \midrule

        \multicolumn{13}{l}{\cellcolor[HTML]{EEEEEE}{\textit{With Final BA}}} \\
        
        Splat-SLAM~\cite{sandstrom2024splat}
        & 273.08 & 255.52 & 260.95 & 466.85 & 354.35
        & 157.05 & 165.36 & 127.94
        & 216.10 & 197.49 & 192.43 & 242.47 \\
        
        DROID-SLAM~\cite{teed2021droid}
        & 1.30 & 1.40 & 2.20 & \textbf{4.30} & 4.30
        & 3.70 & 1.20 & 2.00
        & 1.70 & 1.30 & 1.40 & 2.20 \\

        HI-SLAM2~\cite{zhang2024hislam2} 
        & 1.18 & 1.23 & 4.70 & 4.79 & \textbf{3.49}
        & \textbf{1.82} & \textbf{1.07} & \textbf{1.49}
        & \textbf{0.92} & 1.32 & 2.33 & 2.21 \\

        \textbf{VIGS-SLAM (Ours)}
        & \textbf{1.10} & \textbf{1.12} & \textbf{2.07} & 4.55 & 4.04
        & 3.58 & 1.10 & 1.77
        & 1.73 & \textbf{1.06} & \textbf{1.26} & \textbf{2.13} \\

        \bottomrule
    \end{tabular}
    }

    \label{tab:tracking_euroc_finalba}
\end{table*}

\begin{table*}[t]
\centering
\setlength{\tabcolsep}{2.4pt}
\footnotesize
\caption{\textbf{Tracking Performance on RPNG AR Table Dataset}~\cite{Chen2023rpng}
(ATE RMSE $\downarrow$ [cm]). “F” indicates failure.}
\resizebox{\textwidth}{!}{
\begin{tabular}{lrrrrrrrr
!{\smash{\tikz[baseline]{\draw[densely dashed, gray!80, line width=0.8pt] (0pt,-2pt)--(0pt,8pt);}}}
r}
\toprule
Method & \tt{table\_01} & \tt{table\_02} & \tt{table\_03} & \tt{table\_04} &
\tt{table\_05} & \tt{table\_06} & \tt{table\_07} & \tt{table\_08} &
\textbf{Avg.} \\
\midrule

\multicolumn{10}{l}{\cellcolor[HTML]{EEEEEE}{\textit{Without Final BA}}} \\

Splat-SLAM~\cite{sandstrom2024splat}
& 22.46 & 41.50 & 37.93 & \textbf{1.09} & \textbf{1.19} & 1.53 & 1.76 & 4.37 & 13.98 \\

DROID-SLAM~\cite{teed2021droid}
& 9.59 & 6.58 & 7.28 & 11.82 & 6.24 & 4.22 & 4.27 & 46.59 & 12.07 \\

HI-SLAM2~\cite{zhang2024hislam2}
& 1.43 & 1.66 & 1.23 & 2.59 & F & 1.47 & \textbf{0.97} & \textbf{2.67} & N/A \\

\textbf{VIGS-SLAM (Ours)}
& \textbf{1.31} & \textbf{1.57} & \textbf{1.22} & 1.75 &
1.28 & \textbf{1.38} & 1.08 & 3.86 & \textbf{1.68} \\

\midrule

\multicolumn{10}{l}{\cellcolor[HTML]{EEEEEE}{\textit{With Final BA}}} \\

Splat-SLAM~\cite{sandstrom2024splat}
& 10.98 & 40.44 & 32.87 & \textbf{0.98} & 1.38 & \textbf{1.21} & 1.20 & 4.27 & 11.67 \\

DROID-SLAM~\cite{teed2021droid}
& \textbf{1.20} & \textbf{1.63} & 1.25 & 1.00 & 4.97 & 1.29 & 0.98 & \textbf{3.86} & 2.02 \\

HI-SLAM2~\cite{zhang2024hislam2}
& 1.26 & 1.65 & \textbf{1.16} & \textbf{0.98} & F & 1.32 & 0.99 & 4.08 & N/A \\

\textbf{VIGS-SLAM (Ours)}
& 1.27 & 1.64 & \textbf{1.16} & 0.99 &
\textbf{1.13} & 1.33 & \textbf{0.97} & 4.09 & \textbf{1.57} \\

\bottomrule
\end{tabular}
}

\label{tab:tracking_rpng_finalba}
\end{table*}

\begin{table*}[t]
\centering
\footnotesize
\caption{\textbf{Tracking Performance on UTMM Dataset}~\cite{sun2024mm3dgs}
(ATE RMSE $\downarrow$ [cm]).}
\resizebox{\textwidth}{!}{
\begin{tabular}{lrrrrrrrr
!{\smash{\tikz[baseline]{\draw[densely dashed, gray!80, line width=0.8pt] (0pt,-2pt)--(0pt,8pt);}}}
r}
\toprule
Method  & \tt{Ego-1} & \tt{Ego-2} & \tt{EgoDrv} & \tt{FastStr} &
\tt{SStr-1} & \tt{SStr-2} & \tt{Sq-1} & \tt{Sq-2} & \textbf{Avg.} \\
\midrule

\multicolumn{10}{l}{\cellcolor[HTML]{EEEEEE}{\textit{Without Final BA}}} \\

Splat-SLAM~\cite{sandstrom2024splat}
& \textbf{1.38} & \textbf{0.62} & 3.26 & \textbf{0.95} & 5.48 & \textbf{0.54} & 103.60 & 71.29 & 23.39 \\

DROID-SLAM~\cite{teed2021droid}
& 2.00 & 3.17 & 30.94 & 1.30 & 0.97 & 0.86 & 14.95 & \textbf{9.10} & 7.91 \\

HI-SLAM2~\cite{zhang2024hislam2}
& 2.06 & 3.35 & 4.36 & 0.99 & \textbf{0.71} & 0.84 & 27.85 & 24.63 & 8.10 \\

\textbf{VIGS-SLAM (Ours)}
& 1.81 & 0.93 & \textbf{1.45} & 1.20 & 0.81 & 0.93 & \textbf{2.17} & 16.61 & \textbf{3.24} \\

\midrule

\multicolumn{10}{l}{\cellcolor[HTML]{EEEEEE}{\textit{With Final BA}}} \\

Splat-SLAM~\cite{sandstrom2024splat}
& 1.57 & 0.72 & 3.15 & \textbf{0.92} & \textbf{0.31} & \textbf{0.48} & 119.19 & 75.65 & 25.25 \\

DROID-SLAM~\cite{teed2021droid}
& 0.61 & 0.52 & 2.99 & 1.04 & 0.90 & 0.80 & 3.75 & 52.02 & 7.83 \\

HI-SLAM2~\cite{zhang2024hislam2}
& 0.57 & 0.65 & 2.70 & 1.41 & 0.71 & 1.29 & 25.25 & 27.34 & 8.46 \\

\textbf{VIGS-SLAM (Ours)}
& \textbf{0.48} & \textbf{0.41} & \textbf{2.13} & 1.02 &
1.05 & 1.01 & \textbf{3.57} & \textbf{19.31} &
\textbf{3.62} \\

\bottomrule
\end{tabular}
}

\label{tab:tracking_utmm_finalba}
\end{table*}

\begin{table*}[t]
\centering
\footnotesize
\caption{\textbf{Tracking Performance on FAST-LIVO2 Dataset}~\cite{zheng2024fast}
(ATE RMSE $\downarrow$ [cm]).}
\begin{tabular}{lrrrrr
!{\smash{\tikz[baseline]{\draw[densely dashed, gray!80, line width=0.8pt] (0pt,-2pt)--(0pt,8pt);}}}
r}
\toprule
Method & \tt{CBD1} & \tt{CBD2} & \tt{HKU} & \tt{Retail} & \tt{SYSU1} & \textbf{Avg.} \\
\midrule

\multicolumn{7}{l}{\cellcolor[HTML]{EEEEEE}{\textit{Without Final BA}}} \\

Splat-SLAM~\cite{sandstrom2024splat}
& 5.52 & 7.74 & 4.44 & 212.01 & 313.56 & 108.65 \\

DROID-SLAM~\cite{teed2021droid}
& 72.48 & 15.67 & 15.10 & 16.81 & 50.94 & 34.20 \\

HI-SLAM2~\cite{zhang2024hislam2}
& \textbf{4.38} & 24.30 & 4.87 & \textbf{7.20} & 10.36 & 10.22 \\

\textbf{VIGS-SLAM (Ours)}
& 4.50 & \textbf{5.76} & \textbf{3.88} & 8.88 & \textbf{7.36} & \textbf{6.08} \\

\midrule

\multicolumn{7}{l}{\cellcolor[HTML]{EEEEEE}{\textit{With Final BA}}} \\

Splat-SLAM~\cite{sandstrom2024splat} 
& 4.44 & 15.56 & 2.95 & 196.71 & 299.82 & 103.89 \\

DROID-SLAM~\cite{teed2021droid}
& 73.51 & 12.30 & 23.98 & 5.25 & 12.12 & 25.43 \\

HI-SLAM2~\cite{zhang2024hislam2} 
& 4.09 & 4.68 & \textbf{2.60} & 4.94 & 8.31 & 4.92 \\

\textbf{VIGS-SLAM (Ours)}
& \textbf{4.02} & \textbf{4.31} & 2.72 & \textbf{4.84} & \textbf{8.23} & \textbf{4.82} \\

\bottomrule
\end{tabular}

\label{tab:tracking_livo2_finalba}
\end{table*}

\paragraph{Final Global Bundle Adjustment}
We report results both before and after the final global bundle adjustment in \tabref{tab:tracking_euroc_finalba}, \tabref{tab:tracking_rpng_finalba}, \tabref{tab:tracking_utmm_finalba}, and \tabref{tab:tracking_livo2_finalba}. Across all datasets, our method achieves the best average performance, both with and without the final global BA refinement.

\begin{table*}[t]
\centering
\footnotesize
\setlength{\tabcolsep}{2.5pt}
\caption{\textbf{Rendering Evaluation on RPNG AR Table Dataset~\cite{Chen2023rpng}}. `F' indicates failure. VINGS-Mono~\cite{wu2025vings} does not do final color refinement.}
\resizebox{\textwidth}{!}{
\begin{tabular}{llrrrrrrrr
!{\smash{\tikz[baseline]{\draw[densely dashed, gray!80, line width=0.8pt] (0pt,-2pt)--(0pt,8pt);}}}
r}
\toprule
Metrics & Method & \tt{table\_01} & \tt{table\_02} & \tt{table\_03} & \tt{table\_04} & \tt{table\_05} & \tt{table\_06} & \tt{table\_07} & \tt{table\_08} & {\textbf{Avg.}} \\
\midrule
\multicolumn{11}{l}{\cellcolor[HTML]{EEEEEE}{\textit{Before Final Color Refinement}}} \\ 
\multirow{4}{*}{PSNR $\uparrow$} 
& VINGS-Mono~\cite{wu2025vings} & 11.34 & 10.73 & 11.13 & 10.86 & 11.06 & 10.88 & 11.24 & 10.99 & 11.03 \\
& Splat-SLAM~\cite{sandstrom2024splat} & 16.12 & 10.96 & 14.16 & 17.43 & 19.92 & 17.63 & 20.76 & \textbf{21.59} & 17.32 \\
& HI-SLAM2~\cite{zhang2024hislam2} & 22.10 & 19.84 & 19.97 & 20.17 & F & 21.82 & 23.77 & 20.19 & *21.12 \\
& \textbf{VIGS-SLAM (Ours)}  & \textbf{23.41} & \textbf{20.84} & \textbf{20.71} & \textbf{21.97} & \textbf{21.44} & \textbf{23.47} & \textbf{24.81} & 21.05 & \textbf{22.21} \\
\hdashline

\multirow{4}{*}{SSIM $\uparrow$} 
& VINGS-Mono~\cite{wu2025vings} & 0.229 & 0.168 & 0.200 & 0.213 & 0.307 & 0.351 & 0.380 & 0.266 & 0.264 \\
& Splat-SLAM~\cite{sandstrom2024splat} & 0.436 & 0.223 & 0.338 & 0.583 & 0.664 & 0.650 & 0.705 & \textbf{0.742} & 0.543 \\
& HI-SLAM2~\cite{zhang2024hislam2} & 0.698 & 0.611 & 0.601 & 0.662 & F & 0.715 & 0.803 & 0.705 & *0.685 \\
& \textbf{VIGS-SLAM (Ours)}  & \textbf{0.750} & \textbf{0.654} & \textbf{0.639} & \textbf{0.742} & \textbf{0.684} & \textbf{0.775} & \textbf{0.821} & 0.720 & \textbf{0.723} \\
\hdashline

\multirow{4}{*}{LPIPS $\downarrow$} 
& VINGS-Mono~\cite{wu2025vings}  & 0.689 & 0.702 & 0.704 & 0.687 & 0.718 & 0.728 & 0.687 & 0.716 & 0.704 \\
& Splat-SLAM~\cite{sandstrom2024splat} & 0.483 & 0.630 & 0.649 & 0.376 & 0.404 & 0.458 & 0.373 & \textbf{0.347} & 0.465 \\
& HI-SLAM2~\cite{zhang2024hislam2} & 0.317 & 0.388 & 0.411 & 0.306 & F & 0.398 & 0.284 & 0.401 & *0.358 \\
& \textbf{VIGS-SLAM (Ours)}  & \textbf{0.289} & \textbf{0.338} & \textbf{0.353} & \textbf{0.247} & \textbf{0.345} & \textbf{0.304} & \textbf{0.252} & 0.383 & \textbf{0.314} \\

\midrule
\multicolumn{11}{l}{\cellcolor[HTML]{EEEEEE}{\textit{After Final Color Refinement}}} \\ 
\noalign{\vskip 1pt}
\multirow{3}{*}{PSNR $\uparrow$} 
& Splat-SLAM~\cite{sandstrom2024splat} & 18.36 & 11.67 & 15.70 & 18.52 & 20.07 & 18.37 & 20.53 & 22.61 & 18.23 \\
& HI-SLAM2~\cite{zhang2024hislam2} & 24.18 & 21.61 & 23.52 & 24.29 & F & 24.99 & \textbf{26.88} & \textbf{28.05} & *24.79 \\
& \textbf{VIGS-SLAM (Ours)} 
& \textbf{25.20} & \textbf{21.89} & \textbf{24.39} & \textbf{26.03} & \textbf{24.76} & \textbf{25.18} & 26.11 & 27.88 & \textbf{25.18} \\
\hdashline

\multirow{3}{*}{SSIM $\uparrow$} 
& Splat-SLAM~\cite{sandstrom2024splat} & 0.556 & 0.234 & 0.381 & 0.679 & 0.708 & 0.720 & 0.745 & 0.791 & 0.602 \\
& HI-SLAM2~\cite{zhang2024hislam2} 
& 0.812 & 0.765 & 0.796 & 0.845 & F & \textbf{0.860} & \textbf{0.893} & \textbf{0.907} & *0.840 \\
& \textbf{VIGS-SLAM (Ours)}
& \textbf{0.837} & \textbf{0.780} & \textbf{0.831} & \textbf{0.884} & \textbf{0.846} & \textbf{0.860} & 0.874 & 0.902 & \textbf{0.852} \\
\hdashline

\multirow{3}{*}{LPIPS $\downarrow$} 
& Splat-SLAM~\cite{sandstrom2024splat}  & 0.365 & 0.548 & 0.541 & 0.294 & 0.292 & 0.329 & 0.289 & 0.243 & 0.363 \\
& HI-SLAM2~\cite{zhang2024hislam2}
& 0.215 & 0.237 & 0.237 & 0.177 & F & 0.211 & \textbf{0.158} & \textbf{0.145} & *0.197 \\
& \textbf{VIGS-SLAM (Ours)} 
& \textbf{0.189} & \textbf{0.210} & \textbf{0.193} & \textbf{0.135} & \textbf{0.187} & \textbf{0.187} & 0.166 & 0.152 & \textbf{0.177} \\
\bottomrule
\end{tabular}
}

\label{tab:rendering_rpng}
\end{table*}

\begin{table*}[t]
\centering
\footnotesize
\caption{\textbf{Rendering Evaluation on UTMM Dataset~\cite{sun2024mm3dgs}}. VINGS-Mono~\cite{wu2025vings} does not do final color refinement.}
\resizebox{\textwidth}{!}{
\begin{tabular}{llrrrrrrrr
!{\smash{\tikz[baseline]{\draw[densely dashed, gray!80, line width=0.8pt] (0pt,-2pt)--(0pt,8pt);}}}
r}
\toprule
Metrics & Method  & \tt{Ego-1} & \tt{Ego-2} & \tt{EgoDrv} & \tt{FastStr} & \tt{SStr-1} & \tt{SStr-2} & \tt{Sq-1} & \tt{Sq-2} & \textbf{{Avg.}} \\

\midrule
\multicolumn{11}{l}{\cellcolor[HTML]{EEEEEE}{\textit{Before Final Color Refinement}}} \\ 
\multirow{4}{*}{PSNR $\uparrow$} 
& VINGS-Mono~\cite{wu2025vings} & 11.16 & 10.33 & 11.47 & 11.52 & 13.36 & 14.52 & 10.78 & 11.66 & 11.85 \\
& Splat-SLAM~\cite{sandstrom2024splat} & 17.90 & 18.65 & 17.95 & \phantom{0}9.74 & 12.09 & \phantom{0}9.97 & 11.32 & 10.89 & 13.56 \\
& HI-SLAM2~\cite{zhang2024hislam2} & 18.87 & 16.69 & 15.56 & 21.31 & \textbf{22.16} & \textbf{22.63} & 17.07 & 16.45 & 18.84 \\
& \textbf{VIGS-SLAM (Ours)}  & \textbf{20.05} & \textbf{20.39} & \textbf{21.54} & \textbf{21.98} & 20.66 & 21.92 & \textbf{19.98} & \textbf{20.42} & \textbf{20.87} \\

\hdashline
\multirow{4}{*}{SSIM $\uparrow$} 
& VINGS-Mono~\cite{wu2025vings}  &  0.399 & 0.366 & 0.384 & 0.407 & 0.456 & 0.496 & 0.354 & 0.404 & 0.408 \\
& Splat-SLAM~\cite{sandstrom2024splat} & 0.651 & 0.663 & 0.588 & 0.278 & 0.466 & 0.278 & 0.418 & 0.416 & 0.470 \\
& HI-SLAM2~\cite{zhang2024hislam2} & 0.675 & 0.612 & 0.526 & 0.682 & \textbf{0.713} & \textbf{0.725} & 0.564 & 0.558 & 0.632 \\
& \textbf{VIGS-SLAM (Ours)}  & \textbf{0.711} & \textbf{0.716} & \textbf{0.696} & \textbf{0.695} & 0.669 & 0.695 & \textbf{0.644} & \textbf{0.668} & \textbf{0.687} \\

\hdashline
\multirow{4}{*}{LPIPS $\downarrow$} 
& VINGS-Mono~\cite{wu2025vings}  &  0.696 & 0.734 & 0.697 & 0.623 & 0.583 & 0.584 & 0.692 & 0.666 & 0.660 \\
& Splat-SLAM~\cite{sandstrom2024splat} & 0.490 & 0.498 & 0.548 & 0.747 & 0.727 & 0.704 & 0.733 & 0.776 & 0.653 \\
& HI-SLAM2~\cite{zhang2024hislam2} & 0.438 & 0.498 & 0.624 & \textbf{0.449} & \textbf{0.435} & \textbf{0.423} & 0.568 & 0.575 & 0.501 \\
& \textbf{VIGS-SLAM (Ours)}  & \textbf{0.394} & \textbf{0.382} & \textbf{0.399} & 0.458 & 0.482 & 0.484 & \textbf{0.470} & \textbf{0.460} & \textbf{0.441} \\

\midrule
\multicolumn{11}{l}{\cellcolor[HTML]{EEEEEE}{\textit{After Final Color Refinement}}} \\ 
\noalign{\vskip 1pt}
\multirow{3}{*}{PSNR $\uparrow$} 
& Splat-SLAM~\cite{sandstrom2024splat} & 16.56 & 17.30 & 16.93 & 10.53 & 14.56 & 9.45 & 11.48 & 10.40 & 13.40 \\
& HI-SLAM2~\cite{zhang2024hislam2} 
& 21.08 & 22.01 & 22.30 & 17.80 & 16.38 & 16.62 & 19.95 & 21.15 & 19.66 \\
& \textbf{VIGS-SLAM (Ours)}  
& \textbf{21.55} & \textbf{22.66} & \textbf{23.47} & \textbf{19.77} & \textbf{17.44} & \textbf{18.54} & \textbf{21.90} & \textbf{22.71} & \textbf{21.00} \\

\hdashline
\multirow{3}{*}{SSIM $\uparrow$} 
& Splat-SLAM~\cite{sandstrom2024splat}  
& 0.622 & 0.641 & 0.568 & 0.255 & 0.505 & 0.229 & 0.412 & 0.392 & 0.453 \\
& HI-SLAM2~\cite{zhang2024hislam2} 
& 0.759 & 0.787 & 0.720 & 0.578 & 0.570 & 0.584 & 0.659 & 0.704 & 0.670 \\
& \textbf{VIGS-SLAM (Ours)}  
& \textbf{0.773} & \textbf{0.792} & \textbf{0.770} & \textbf{0.678} & \textbf{0.613} & \textbf{0.621} & \textbf{0.746} & \textbf{0.761} & \textbf{0.719} \\

\hdashline
\multirow{3}{*}{LPIPS $\downarrow$} 
& Splat-SLAM~\cite{sandstrom2024splat} 
& 0.416 & 0.404 & 0.469 & 0.718 & 0.506 & 0.765 & 0.668 & 0.741 & 0.586 \\
& HI-SLAM2~\cite{zhang2024hislam2} 
& 0.281 & 0.256 & 0.327 & 0.422 & 0.435 & 0.447 & 0.384 & 0.347 & 0.362 \\
& \textbf{VIGS-SLAM (Ours)}   
& \textbf{0.251} & \textbf{0.219} & \textbf{0.276} & \textbf{0.314} & \textbf{0.426} & \textbf{0.402} & \textbf{0.296} & \textbf{0.276} & \textbf{0.308} \\
\bottomrule
\end{tabular}
}

\label{tab:rendering_utmm}
\end{table*}

\begin{table*}[t]
\centering
\footnotesize
\setlength{\tabcolsep}{2.pt}
\caption{\textbf{Rendering Evaluation on FAST-LIVO2 Dataset~\cite{zheng2024fast}}. VINGS-Mono~\cite{wu2025vings} does not do final color refinement.}
\begin{tabular}{llrrrrr
!{\smash{\tikz[baseline]{\draw[densely dashed, gray!80, line width=0.8pt] (0pt,-2pt)--(0pt,8pt);}}}
r}
\toprule
Metrics & Method & \tt{CBD1} & \tt{CBD2} & \tt{HKU} & \tt{Retail} & \tt{SYSU1} & {\textbf{Avg.}} \\
\midrule
\multicolumn{8}{l}{\cellcolor[HTML]{EEEEEE}{\textit{Before Final Color Refinement}}} \\ 
\multirow{4}{*}{PSNR $\uparrow$} 
& VINGS-Mono~\cite{wu2025vings} & 9.87 & 11.07 & 11.88 & 10.83 & 8.15 & 10.36 \\
& Splat-SLAM~\cite{sandstrom2024splat} & 15.84 & 17.99 & 18.34 & \phantom{0}5.39 & 12.22 & 13.96 \\
& HI-SLAM2~\cite{zhang2024hislam2}     & 21.05 & 19.58 & 25.01 & 21.47 & 20.31 & 21.49 \\
& \textbf{VIGS-SLAM (Ours)}                            & \textbf{21.68} & \textbf{22.68} & \textbf{26.38} & \textbf{22.63} & \textbf{22.37} & \textbf{23.15} \\
\hdashline

\multirow{4}{*}{SSIM $\uparrow$} 
& VINGS-Mono~\cite{wu2025vings}  & 0.366 & 0.392 & 0.315 & 0.390 & 0.253 & 0.343 \\
& Splat-SLAM~\cite{sandstrom2024splat} & 0.679 & 0.663 & 0.566 & 0.001 & 0.512 & 0.484 \\
& HI-SLAM2~\cite{zhang2024hislam2}     & 0.749 & 0.680 & 0.694 & 0.658 & 0.679 & 0.692 \\
& \textbf{VIGS-SLAM (Ours)}                          & \textbf{0.764} & \textbf{0.759} & \textbf{0.714} & \textbf{0.709} & \textbf{0.699} & \textbf{0.729} \\
\hdashline

\multirow{4}{*}{LPIPS $\downarrow$}
& VINGS-Mono~\cite{wu2025vings}  & 0.702 & 0.719 & 0.762 & 0.711 & 0.725 & 0.724 \\
& Splat-SLAM~\cite{sandstrom2024splat} & 0.682 & 0.584 & 0.613 & 1.034 & 0.814 & 0.745 \\
& HI-SLAM2~\cite{zhang2024hislam2}     & 0.540 & 0.594 & 0.526 & 0.462 & 0.677 & 0.560 \\
& \textbf{VIGS-SLAM (Ours)}                         & \textbf{0.515} & \textbf{0.455} & \textbf{0.490} & \textbf{0.361} & \textbf{0.611} & \textbf{0.487} \\

\midrule
\multicolumn{8}{l}{\cellcolor[HTML]{EEEEEE}{\textit{After Final Color Refinement}}} \\ 
\noalign{\vskip 1pt}
\multirow{3}{*}{PSNR $\uparrow$} 
& Splat-SLAM~\cite{sandstrom2024splat} & 19.38 & 18.29 & 20.94 & 5.93 & 11.60 & 15.23 \\
& HI-SLAM2~\cite{zhang2024hislam2}     & 22.50 & 25.16 & 29.79 & 26.87 & 23.13 & 25.49 \\
& \textbf{VIGS-SLAM (Ours)}           
& \textbf{22.68} & \textbf{25.42} & \textbf{30.19} & \textbf{27.97} & \textbf{23.97} & \textbf{26.04} \\
\hdashline

\multirow{3}{*}{SSIM $\uparrow$} 
& Splat-SLAM~\cite{sandstrom2024splat}  & 0.724 & 0.661 & 0.623 & 0.070 & 0.460 & 0.508 \\
& HI-SLAM2~\cite{zhang2024hislam2}      
& \textbf{0.820} & \textbf{0.867} & 0.811 & 0.877 & 0.778 & 0.831 \\
& \textbf{VIGS-SLAM (Ours)}             
& 0.808 & \textbf{0.867} & \textbf{0.817} & \textbf{0.892} & \textbf{0.812} & \textbf{0.839} \\
\hdashline

\multirow{3}{*}{LPIPS $\downarrow$} 
& Splat-SLAM~\cite{sandstrom2024splat} & 0.532 & 0.511 & 0.536 & 1.009 & 0.791 & 0.676 \\
& HI-SLAM2~\cite{zhang2024hislam2}     
& \textbf{0.299} & 0.241 & \textbf{0.320} & 0.132 & 0.386 & 0.276 \\
& \textbf{VIGS-SLAM (Ours)}             
& 0.334 & \textbf{0.234} & \textbf{0.320} & \textbf{0.108} & \textbf{0.333} & \textbf{0.266} \\
\bottomrule
\end{tabular}

\label{tab:rendering_livo2}
\end{table*}

\paragraph{Detailed Rendering Results}
We report detailed per-sequence rendering results both before the final color refinement (with average results shown in \tabrefn{tab:mapping_avg} of the main paper) and after refinement. The full results are provided in \tabref{tab:rendering_rpng}, \tabref{tab:rendering_utmm}, and \tabref{tab:rendering_livo2}.

\paragraph{Detailed Results for Strided Evaluation}
We further include detailed per-sequence results for the strided evaluation shown in \figrefn{fig:tracking_stride_avg} in the main paper, as shown in \tabref{tab:tracking_euroc_stride} and \tabref{tab:tracking_rpng_stride}.

While ATE~RMSE is a widely used accuracy metric, it alone can be misleading under strided evaluation. A method may fail to track large portions of a sequence yet still obtain a low ATE by aligning only a short, easy segment. To address this limitation, we additionally report Recall metrics, which quantify how much of the trajectory is successfully tracked within a given error threshold and provide a more comprehensive view of robustness.
Unlike classical VIO methods that select favorable starting points, we initialize from the first $N^{\mathrm{iner}}_{\mathrm{init}}$ keyframes and track the entire sequence from the beginning.
For instance, in the RPNG dataset \texttt{table\_02} at stride $=10$, ORB-SLAM3~\cite{campos2021orb3} achieves a seemingly low ATE of \(0.20\,\mathrm{cm}\) but with only \(10.72\%\) recall@10cm, as it initializes successfully only after roughly 220 frames (given the stride) and tracks a short segment before losing track.
In contrast, our method attains \(1.51\,\mathrm{cm}\) ATE with \(100.00\%\) recall, reflecting both robustness and complete trajectory coverage.

\begin{table*}[htbp]
\centering
\caption{\textbf{Tracking Performance on Strided EuRoC Dataset~\cite{burri2016euroc}} (ATE RMSE $\downarrow$ [cm] and Recall $\uparrow$ [\%]). All baseline results are obtained from the authors' official code, using dataset-specific configurations when available.}
\resizebox{0.8\textwidth}{!}{
\begin{tabular}{llrrrrrrrrrrr
!{\smash{\tikz[baseline]{\draw[densely dashed, gray!80, line width=0.8pt] (0pt,-2pt)--(0pt,8pt);}}}
r}
\toprule
Metrics / Stride & Method & \tt{MH\_01} & \tt{MH\_02} & \tt{MH\_03} & \tt{MH\_04} & \tt{MH\_05} &
        \tt{V1\_01} & \tt{V1\_02} & \tt{V1\_03} &
        \tt{V2\_01} & \tt{V2\_02} & \tt{V2\_03} & \textbf{Avg.} \\
\midrule

\multicolumn{14}{l}{\cellcolor[HTML]{EEEEEE}\textit{Stride = 1}} \\[3pt]

\multirow{5}{*}{ATE RMSE [cm] $\downarrow$}
 & HI-SLAM2~\cite{zhang2024hislam2} & 2.66 & 1.44 & 2.71 & 6.86 & \textbf{5.07} & 3.55 & 1.32 & \textbf{2.49} & 2.56 & 1.77 & \textbf{1.92} & 2.94 \\
 & VINS-Mono~\cite{qin2018vins} & 7.56 & 8.59 & 7.57 & 19.85 & 13.45 & 4.42 & 6.54 & 29.71 & 6.58 & 20.41 & 25.64 & 13.67 \\
 & OPEN-VINS~\cite{geneva2020openvins} & 9.48 & 12.78 & 14.83 & 17.46 & 50.61 & 6.34 & 5.61 & 7.16 & 10.47 & 5.96 & 11.68 & 13.85 \\
 & ORB-SLAM3~\cite{campos2021orb3} & 2.78 & 9.03 & 7.50 & 7.86 & 7.99 & \textbf{3.21} & 1.35 & 3.10 & 4.53 & 2.27 & 1.93 & 4.69 \\
 & \textbf{VIGS-SLAM (Ours)} & \textbf{1.42} & \textbf{1.29} & \textbf{2.55} & \textbf{5.16} & 5.64 & 3.67 & \textbf{1.15} & 2.68 & \textbf{2.34} & \textbf{1.53} & 3.27 & \textbf{2.79} \\
\hdashline

\multirow{5}{*}{Recall @ 5cm [\%] $\uparrow$}
 & HI-SLAM2~\cite{zhang2024hislam2} & 98.23 & \textbf{100.00} & 94.21 & 46.49 & 68.21 & 89.14 & \textbf{100.00} & \textbf{97.65} & \textbf{100.00} & 99.74 & \textbf{100.00} & \textbf{90.33} \\
 & VINS-Mono~\cite{qin2018vins} & 46.54 & 46.37 & 44.81 & 10.73 & 17.93 & 73.70 & 52.52 & 6.71 & 62.36 & 5.75 & 5.30 & 33.88 \\
 & OPEN-VINS~\cite{geneva2020openvins} & 31.72 & 28.11 & 27.76 & 12.87 & 5.98 & 63.61 & 72.99 & 46.62 & 43.10 & 69.42 & 27.23 & 39.04 \\
 & ORB-SLAM3~\cite{campos2021orb3} & 71.24 & 64.69 & 64.78 & \textbf{77.48} & \textbf{71.52} & \textbf{98.74} & 97.65 & 93.73 & 89.09 & 95.00 & 92.38 & 83.30 \\
 & \textbf{VIGS-SLAM (Ours)} & \textbf{100.00} & \textbf{100.00} & \textbf{94.29} & 57.42 & 50.00 & 84.27 & \textbf{100.00} & 96.98 & 98.48 & \textbf{100.00} & 91.64 & 88.46 \\
\hdashline

\multirow{5}{*}{Recall @ 10cm [\%] $\uparrow$}
 & HI-SLAM2~\cite{zhang2024hislam2} & \textbf{100.00} & \textbf{100.00} & \textbf{100.00} & 88.96 & 98.68 & \textbf{100.00} & \textbf{100.00} & \textbf{100.00} & \textbf{100.00} & \textbf{100.00} & \textbf{100.00} & 98.88 \\
 & VINS-Mono~\cite{qin2018vins} & 88.69 & 75.14 & 88.91 & 37.93 & 54.98 & 95.73 & 95.67 & 24.20 & 86.34 & 35.39 & 24.42 & 64.31 \\
 & OPEN-VINS~\cite{geneva2020openvins} & 70.04 & 65.01 & 67.64 & 51.39 & 15.62 & 92.31 & \textbf{100.00} & 92.06 & 72.47 & 96.28 & 72.79 & 72.33 \\
 & ORB-SLAM3~\cite{campos2021orb3} & 76.24 & 75.16 & 79.61 & 83.29 & 86.85 & \textbf{100.00} & 98.40 & 93.99 & 94.10 & 97.43 & \textbf{100.00} & 89.55 \\
 & \textbf{VIGS-SLAM (Ours)} & \textbf{100.00} & \textbf{100.00} & \textbf{100.00} & \textbf{100.00} & \textbf{100.00} & \textbf{100.00} & \textbf{100.00} & \textbf{100.00} & \textbf{100.00} & \textbf{100.00} & 98.25 & \textbf{99.84} \\
\hdashline
\multicolumn{14}{l}{\cellcolor[HTML]{EEEEEE}\textit{Stride = 5}} \\[3pt]

\multirow{5}{*}{ATE RMSE [cm] $\downarrow$}
 & HI-SLAM2~\cite{zhang2024hislam2} & \textbf{1.64} & \textbf{1.58} & \textbf{2.82} & 14.49 & 5.76 & 3.43 & 1.26 & 19.50 & 3.12 & 2.21 & 197.21 & 23.00 \\
 & VINS-Mono~\cite{qin2018vins} & 6.97 & 5.90 & 19.78 & 14.22 & 15.36 & 4.86 & 25.45 & 150.27 & 6.46 & 53.64 & 201.71 & 45.88 \\
 & OPEN-VINS~\cite{geneva2020openvins} & 11.20 & 8.82 & 23.68 & 18.83 & 36.15 & 6.14 & 16.41 & 151.30 & 112.66 & 66.56 & 199.50 & 59.21 \\
 & ORB-SLAM3~\cite{campos2021orb3} & 2.03 & 3.59 & 3.65 & 11.48 & 10.95 & \textbf{3.24} & \textbf{1.08} & \textbf{1.42} & 4.77 & 1.86 & F & N/A \\
 & \textbf{VIGS-SLAM (Ours)} & 2.04 & 3.67 & 3.43 & \textbf{8.55} & \textbf{5.37} & 3.48 & 1.55 & 2.25 & \textbf{2.09} & \textbf{1.29} & \textbf{2.26} & \textbf{3.27} \\
\midrule

\multirow{5}{*}{Recall @ 5cm [\%] $\uparrow$}
 & HI-SLAM2~\cite{zhang2024hislam2} & \textbf{100.00} & \textbf{100.00} & \textbf{94.67} & 9.81 & 56.77 & \textbf{88.39} & \textbf{100.00} & 87.17 & 95.33 & 98.42 & 0.00 & 75.51 \\
 & VINS-Mono~\cite{qin2018vins} & 26.96 & 29.45 & 5.63 & 9.56 & 6.64 & 57.41 & 3.18 & 0.12 & 52.48 & 0.62 & 0.10 & 17.47 \\
 & OPEN-VINS~\cite{geneva2020openvins} & 22.92 & 19.88 & 2.33 & 14.66 & 1.12 & 47.99 & 7.30 & 0.22 & 0.26 & 1.66 & 0.27 & 10.78 \\
 & ORB-SLAM3~\cite{campos2021orb3} & 77.13 & 53.30 & 48.09 & 3.93 & 11.57 & 75.96 & 44.55 & 52.25 & 80.12 & 58.86 & 0.00 & 45.98 \\
 & \textbf{VIGS-SLAM (Ours)} & 96.51 & 86.15 & 90.36 & \textbf{20.57} & \textbf{66.48} & 86.27 & \textbf{100.00} & \textbf{100.00} & \textbf{100.00} & \textbf{100.00} & \textbf{96.45} & \textbf{85.71} \\
\hdashline

\multirow{5}{*}{Recall @ 10cm [\%] $\uparrow$}
 & HI-SLAM2~\cite{zhang2024hislam2} & \textbf{100.00} & \textbf{100.00} & \textbf{100.00} & 37.85 & 95.31 & \textbf{100.00} & \textbf{100.00} & 99.67 & \textbf{100.00} & \textbf{100.00} & 0.35 & 84.83 \\
 & VINS-Mono~\cite{qin2018vins} & 69.01 & 67.63 & 36.69 & 49.38 & 35.62 & 92.72 & 16.05 & 0.60 & 88.65 & 4.36 & 1.06 & 41.98 \\
 & OPEN-VINS~\cite{geneva2020openvins} & 65.98 & 71.09 & 14.90 & 29.37 & 8.19 & 92.57 & 34.67 & 0.55 & 0.94 & 10.77 & 0.65 & 29.97 \\
 & ORB-SLAM3~\cite{campos2021orb3} & 97.38 & 85.51 & 74.10 & 23.82 & 47.62 & 96.56 & 70.90 & 81.22 & 93.13 & 85.35 & 0.00 & 68.69 \\
 & \textbf{VIGS-SLAM (Ours)} & \textbf{100.00} & 96.97 & 98.93 & \textbf{82.86} & \textbf{96.15} & \textbf{100.00} & \textbf{100.00} & \textbf{100.00} & \textbf{100.00} & \textbf{100.00} & \textbf{100.00} & \textbf{97.72} \\
\midrule
\multicolumn{14}{l}{\cellcolor[HTML]{EEEEEE}\textit{Stride = 10}} \\[3pt]

\multirow{5}{*}{ATE RMSE [cm] $\downarrow$}
 & HI-SLAM2~\cite{zhang2024hislam2} & 6.23 & 39.40 & 269.95 & 239.54 & 321.23 & 3.54 & 167.69 & 156.13 & 2.66 & 191.08 & 191.29 & 144.43 \\
 & VINS-Mono~\cite{qin2018vins} & 23.49 & 25.15 & 100.74 & F & 77.16 & 14.16 & F & 165.96 & 15.62 & F & 216.58 & N/A \\
 & OPEN-VINS~\cite{geneva2020openvins} & F & F & F & F & F & F & F & F & F & F & F & N/A \\
 & ORB-SLAM3~\cite{campos2021orb3} & F & \textbf{2.23} & \textbf{3.14} & \textbf{2.21} & 11.39 & \textbf{3.14} & F & F & 4.95 & \textbf{0.02} & F & N/A \\
 & \textbf{VIGS-SLAM (Ours)} & \textbf{3.32} & 3.51 & 3.85 & 5.72 & \textbf{5.11} & 3.49 & \textbf{3.49} & \textbf{2.61} & \textbf{2.53} & 1.00 & \textbf{2.68} & \textbf{3.39} \\
\hdashline

\multirow{5}{*}{Recall @ 5cm [\%] $\uparrow$}
 & HI-SLAM2~\cite{zhang2024hislam2} & \textbf{98.17} & 2.38 & 0.00 & 0.00 & 0.00 & \textbf{86.54} & 0.00 & 1.19 & 99.26 & 0.97 & 0.60 & \textbf{26.28} \\
 & VINS-Mono~\cite{qin2018vins} & 3.00 & 4.20 & 0.17 & 0.00 & 0.00 & 7.12 & 0.00 & 0.06 & 14.19 & 0.00 & 0.20 & 2.63 \\
 & OPEN-VINS~\cite{geneva2020openvins} & 0.00 & 0.00 & 0.00 & 0.00 & 0.00 & 0.00 & 0.00 & 0.00 & 0.00 & 0.00 & 0.00 & 0.00 \\
 & ORB-SLAM3~\cite{campos2021orb3} & 0.00 & 18.75 & 18.30 & 7.93 & 3.66 & 41.90 & 0.00 & 0.00 & 52.20 & 2.43 & 0.00 & 13.20 \\
 & \textbf{VIGS-SLAM (Ours)} & 90.45 & \textbf{86.36} & \textbf{85.71} & \textbf{58.91} & \textbf{65.28} & 86.43 & \textbf{83.55} & \textbf{95.51} & \textbf{99.29} & \textbf{100.00} & \textbf{94.67} & 86.02 \\
\hdashline

\multirow{5}{*}{Recall @ 10cm [\%] $\uparrow$}
 & HI-SLAM2~\cite{zhang2024hislam2} & 98.62 & 7.14 & 0.00 & 0.00 & 0.00 & \textbf{100.00} & 2.04 & 1.19 & \textbf{100.00} & 2.43 & 3.61 & 28.64 \\
 & VINS-Mono~\cite{qin2018vins} & 19.14 & 24.21 & 1.75 & 0.00 & 0.26 & 29.95 & 0.00 & 0.33 & 53.16 & 0.00 & 0.52 & 11.75 \\
 & OPEN-VINS~\cite{geneva2020openvins} & 0.00 & 0.00 & 0.00 & 0.00 & 0.00 & 0.00 & 0.00 & 0.00 & 0.00 & 0.00 & 0.00 & 0.00 \\
 & ORB-SLAM3~\cite{campos2021orb3} & 0.00 & 35.93 & 31.91 & 12.14 & \textbf{14.12} & 73.83 & 0.00 & 0.00 & 85.47 & 2.61 & 0.00 & 23.27 \\
 & \textbf{VIGS-SLAM (Ours)} & \textbf{100.00} & \textbf{98.86} & \textbf{99.49} & \textbf{93.80} & \textbf{97.92} & \textbf{100.00} & \textbf{100.00} & \textbf{100.00} & \textbf{100.00} & \textbf{100.00} & \textbf{98.82} & \textbf{98.99} \\
\midrule
\multicolumn{14}{l}{\cellcolor[HTML]{EEEEEE}\textit{Stride = 20}} \\[3pt]

\multirow{5}{*}{ATE RMSE [cm] $\downarrow$}
 & HI-SLAM2~\cite{zhang2024hislam2} & \textbf{1.58} & \textbf{11.65} & 356.45 & 415.73 & 459.36 & 140.07 & \textbf{176.20} & \textbf{145.07} & \textbf{124.11} & 201.46 & \textbf{198.88} & 202.78 \\
 & VINS-Mono~\cite{qin2018vins} & 3.58 & F & F & F & F & F & F & F & F & F & F & N/A \\
 & OPEN-VINS~\cite{geneva2020openvins} & F & F & F & F & F & F & F & F & F & F & F & N/A \\
 & ORB-SLAM3~\cite{campos2021orb3} & F & F & F & F & F & F & F & F & F & F & F & N/A \\
 & \textbf{VIGS-SLAM (Ours)} & 11.00 & 446.48 & \textbf{7.90} & \textbf{15.75} & \textbf{21.12} & \textbf{3.58} & 178.79 & 152.87 & 200.96 & \textbf{191.40} & 200.88 & \textbf{130.07} \\
\hdashline

\multirow{5}{*}{Recall @ 5cm [\%] $\uparrow$}
 & HI-SLAM2~\cite{zhang2024hislam2} & \textbf{100.00} & \textbf{19.67} & 0.00 & 0.00 & 0.00 & 0.00 & \textbf{1.28} & 0.00 & 0.00 & 0.00 & \textbf{2.27} & 11.20 \\
 & VINS-Mono~\cite{qin2018vins} & 1.02 & 0.00 & 0.00 & 0.00 & 0.00 & 0.00 & 0.00 & 0.00 & 0.00 & 0.00 & 0.00 & 0.09 \\
 & OPEN-VINS~\cite{geneva2020openvins} & 0.00 & 0.00 & 0.00 & 0.00 & 0.00 & 0.00 & 0.00 & 0.00 & 0.00 & 0.00 & 0.00 & 0.00 \\
 & ORB-SLAM3~\cite{campos2021orb3} & 0.00 & 0.00 & 0.00 & 0.00 & 0.00 & 0.00 & 0.00 & 0.00 & 0.00 & 0.00 & 0.00 & 0.00 \\
 & \textbf{VIGS-SLAM (Ours)} & 46.72 & 0.00 & \textbf{56.14} & \textbf{10.71} & \textbf{9.47} & \textbf{90.70} & 0.00 & 0.00 & 0.00 & 0.00 & 0.00 & \textbf{19.43} \\
\hdashline

\multirow{5}{*}{Recall @ 10cm [\%] $\uparrow$}
 & HI-SLAM2~\cite{zhang2024hislam2} & \textbf{100.00} & \textbf{80.33} & 0.00 & 0.00 & 0.00 & 4.69 & \textbf{1.28} & \textbf{2.04} & \textbf{3.41} & 0.00 & \textbf{2.27} & 17.64 \\
 & VINS-Mono~\cite{qin2018vins} & 3.11 & 0.00 & 0.00 & 0.00 & 0.00 & 0.00 & 0.00 & 0.00 & 0.00 & 0.00 & 0.00 & 0.28 \\
 & OPEN-VINS~\cite{geneva2020openvins} & 0.00 & 0.00 & 0.00 & 0.00 & 0.00 & 0.00 & 0.00 & 0.00 & 0.00 & 0.00 & 0.00 & 0.00 \\
 & ORB-SLAM3~\cite{campos2021orb3} & 0.00 & 0.00 & 0.00 & 0.00 & 0.00 & 0.00 & 0.00 & 0.00 & 0.00 & 0.00 & 0.00 & 0.00 \\
 & \textbf{VIGS-SLAM (Ours)} & 67.88 & 1.69 & \textbf{89.47} & \textbf{29.76} & \textbf{32.63} & \textbf{100.00} & 0.00 & 0.00 & 0.00 & \textbf{1.98} & 0.00 & \textbf{29.40} \\
\midrule
\multicolumn{14}{l}{\cellcolor[HTML]{EEEEEE}\textit{Stride = 40}} \\[3pt]

\multirow{5}{*}{ATE RMSE [cm] $\downarrow$}
 & HI-SLAM2~\cite{zhang2024hislam2} & 300.60 & \textbf{289.42} & \textbf{351.79} & \textbf{642.30} & \textbf{662.75} & 177.83 & 177.01 & \textbf{153.13} & 201.51 & 203.35 & \textbf{184.52} & 304.02 \\
 & VINS-Mono~\cite{qin2018vins} & F & F & F & F & F & F & F & F & F & F & F & N/A \\
 & OPEN-VINS~\cite{geneva2020openvins} & F & F & F & F & F & F & F & F & F & F & F & N/A \\
 & ORB-SLAM3~\cite{campos2021orb3} & F & F & F & F & F & F & F & F & F & F & F & N/A \\
 & \textbf{VIGS-SLAM (Ours)} & \textbf{10.27} & 461.22 & 362.04 & 689.67 & 693.66 & \textbf{177.80} & \textbf{176.79} & 154.44 & \textbf{196.70} & \textbf{191.02} & 195.09 & \textbf{300.79} \\
\hdashline

\multirow{5}{*}{Recall @ 5cm [\%] $\uparrow$}
 & HI-SLAM2~\cite{zhang2024hislam2} & 0.00 & 0.00 & 0.00 & 0.00 & 0.00 & 0.00 & 0.00 & 0.00 & 0.00 & 0.00 & \textbf{4.65} & 0.42 \\
 & VINS-Mono~\cite{qin2018vins} & 0.00 & 0.00 & 0.00 & 0.00 & 0.00 & 0.00 & 0.00 & 0.00 & 0.00 & 0.00 & 0.00 & 0.00 \\
 & OPEN-VINS~\cite{geneva2020openvins} & 0.00 & 0.00 & 0.00 & 0.00 & 0.00 & 0.00 & 0.00 & 0.00 & 0.00 & 0.00 & 0.00 & 0.00 \\
 & ORB-SLAM3~\cite{campos2021orb3} & 0.00 & 0.00 & 0.00 & 0.00 & 0.00 & 0.00 & 0.00 & 0.00 & 0.00 & 0.00 & 0.00 & 0.00 \\
 & \textbf{VIGS-SLAM (Ours)} & \textbf{56.47} & 0.00 & 0.00 & 0.00 & 0.00 & 0.00 & 0.00 & 0.00 & 0.00 & 0.00 & 0.00 & \textbf{5.13} \\
\hdashline

\multirow{5}{*}{Recall @ 10cm [\%] $\uparrow$}
 & HI-SLAM2~\cite{zhang2024hislam2} & 1.37 & 0.00 & 0.00 & 0.00 & 0.00 & 0.00 & 0.00 & \textbf{2.00} & 0.00 & 0.00 & \textbf{4.65} & 0.73 \\
 & VINS-Mono~\cite{qin2018vins} & 0.00 & 0.00 & 0.00 & 0.00 & 0.00 & 0.00 & 0.00 & 0.00 & 0.00 & 0.00 & 0.00 & 0.00 \\
 & OPEN-VINS~\cite{geneva2020openvins} & 0.00 & 0.00 & 0.00 & 0.00 & 0.00 & 0.00 & 0.00 & 0.00 & 0.00 & 0.00 & 0.00 & 0.00 \\
 & ORB-SLAM3~\cite{campos2021orb3} & 0.00 & 0.00 & 0.00 & 0.00 & 0.00 & 0.00 & 0.00 & 0.00 & 0.00 & 0.00 & 0.00 & 0.00 \\
 & \textbf{VIGS-SLAM (Ours)} & \textbf{74.12} & 0.00 & 0.00 & 0.00 & 0.00 & \textbf{1.41} & 0.00 & 0.00 & 0.00 & 0.00 & 0.00 & \textbf{6.87} \\
\bottomrule

\end{tabular}}

\label{tab:tracking_euroc_stride}
\end{table*}

\begin{table*}[htbp]
\centering
\caption{\textbf{Tracking Performance on Strided RPNG AR Table Dataset~\cite{Chen2023rpng}} (ATE RMSE $\downarrow$ [cm] and Recall $\uparrow$ [\%]). All baseline results are obtained from the authors' official code, using dataset-specific configurations when available.}
\resizebox{0.8\textwidth}{!}{
\begin{tabular}{lcrrrrrrrr
!{\smash{\tikz[baseline]{\draw[densely dashed, gray!80, line width=0.8pt] (0pt,-2pt)--(0pt,8pt);}}}
r
}
\toprule
Metrics / Stride & Method &\tt{table\_01} & \tt{table\_02} & \tt{table\_03} & \tt{table\_04} & \tt{table\_05} & \tt{table\_06} & \tt{table\_07} & \tt{table\_08} & {\textbf{Avg.}} \\
\midrule

\multicolumn{11}{l}{\cellcolor[HTML]{EEEEEE}\textit{Stride = 1}} \\

\multirow{5}{*}{ATE RMSE [cm] $\downarrow$}
 & HI-SLAM2~\cite{zhang2024hislam2} & 1.43 & 1.66 & 1.23 & 2.59 & F & 1.47 & \textbf{0.97} & \textbf{2.67} & N/A \\
 & VINS-Mono~\cite{qin2018vins} & 2.72 & 5.98 & 3.30 & 4.01 & 2.18 & 1.87 & 2.05 & 5.54 & 3.46 \\
 & OPEN-VINS~\cite{geneva2020openvins} & 4.29 & 3.03 & 3.11 & 6.20 & 3.85 & 4.45 & 6.58 & 9.20 & 5.09 \\
 & ORB-SLAM3~\cite{campos2021orb3} & 2.52 & 15.79 & 1.57 & \textbf{1.22} & 7.34 & 1.49 & 1.24 & 3.43 & 4.33 \\
 & \textbf{VIGS-SLAM (Ours)} & \textbf{1.31} & \textbf{1.57} & \textbf{1.22} & 1.75 & \textbf{1.28} & \textbf{1.38} & 1.08 & 3.86 & \textbf{1.68} \\
\hdashline

\multirow{5}{*}{Recall @ 5cm [\%] $\uparrow$}
 & HI-SLAM2~\cite{zhang2024hislam2} & \textbf{100.00} & \textbf{100.00} & \textbf{100.00} & 95.02 & 0.00 & \textbf{100.00} & \textbf{100.00} & \textbf{98.58} & 86.70 \\
 & VINS-Mono~\cite{qin2018vins} & 72.71 & 58.95 & 68.89 & 76.74 & 90.88 & 85.15 & 87.47 & 81.21 & 77.75 \\
 & OPEN-VINS~\cite{geneva2020openvins} & 89.82 & 93.64 & 95.62 & 76.80 & 93.96 & 90.20 & 77.56 & 49.09 & 83.34 \\
 & ORB-SLAM3~\cite{campos2021orb3} & 93.52 & 82.02 & 97.68 & 96.49 & 73.02 & 99.83 & 95.61 & 96.32 & 91.81 \\
 & \textbf{VIGS-SLAM (Ours)} & \textbf{100.00} & \textbf{100.00} & \textbf{100.00} & \textbf{98.44} & \textbf{100.00} & \textbf{100.00} & \textbf{100.00} & 88.14 & \textbf{98.32} \\
\hdashline

\multirow{5}{*}{Recall @ 10cm [\%] $\uparrow$}
 & HI-SLAM2~\cite{zhang2024hislam2} & \textbf{100.00} & \textbf{100.00} & \textbf{100.00} & 98.58 & 0.00 & \textbf{100.00} & \textbf{100.00} & \textbf{100.00} & 87.32 \\
 & VINS-Mono~\cite{qin2018vins} & 95.74 & 95.92 & 96.56 & 95.45 & 99.66 & 98.22 & 95.01 & 95.28 & 96.48 \\
 & OPEN-VINS~\cite{geneva2020openvins} & \textbf{100.00} & \textbf{100.00} & \textbf{100.00} & 98.36 & 99.78 & 93.74 & 98.51 & 87.58 & 97.25 \\
 & ORB-SLAM3~\cite{campos2021orb3} & 99.98 & 91.69 & 98.08 & 96.56 & 84.12 & \textbf{100.00} & \textbf{100.00} & 98.82 & 96.16 \\
 & \textbf{VIGS-SLAM (Ours)} & \textbf{100.00} & \textbf{100.00} & \textbf{100.00} & \textbf{100.00} & \textbf{100.00} & \textbf{100.00} & \textbf{100.00} & \textbf{100.00} & \textbf{100.00} \\
\midrule
\multicolumn{11}{l}{\cellcolor[HTML]{EEEEEE}\textit{Stride = 5}} \\

\multirow{5}{*}{ATE RMSE [cm] $\downarrow$}
 & HI-SLAM2~\cite{zhang2024hislam2} & 2.91 & 17.33 & 6.26 & 3.83 & 1.76 & 1.52 & 1.14 & \textbf{2.61} & 4.67 \\
 & VINS-Mono~\cite{qin2018vins} & 1.93 & 6.02 & 5.07 & 3.12 & 2.07 & 2.74 & 1.17 & 4.02 & 3.27 \\
 & OPEN-VINS~\cite{geneva2020openvins} & 2.27 & 2.35 & 2.44 & 3.08 & 2.81 & 3.88 & 3.06 & 6.69 & 3.32 \\
 & ORB-SLAM3~\cite{campos2021orb3} & 4.30 & 2.84 & 7.77 & 4.80 & 4.19 & 8.55 & 4.04 & 5.77 & 5.28 \\
 & \textbf{VIGS-SLAM (Ours)} & \textbf{1.26} & \textbf{1.56} & \textbf{1.19} & \textbf{2.50} & \textbf{1.26} & \textbf{1.41} & \textbf{1.03} & 3.55 & \textbf{1.72} \\
\hdashline

\multirow{5}{*}{Recall @ 5cm [\%] $\uparrow$}
 & HI-SLAM2~\cite{zhang2024hislam2} & 96.35 & 12.30 & 94.51 & 89.37 & 99.38 & \textbf{100.00} & \textbf{100.00} & \textbf{98.45} & 86.29 \\
 & VINS-Mono~\cite{qin2018vins} & 75.52 & 70.75 & 83.50 & 85.14 & 93.14 & 84.46 & 91.49 & 81.70 & 83.21 \\
 & OPEN-VINS~\cite{geneva2020openvins} & 74.68 & 90.60 & 95.70 & 93.29 & 93.47 & 78.07 & 90.40 & 62.21 & 84.80 \\
 & ORB-SLAM3~\cite{campos2021orb3} & 12.25 & 91.38 & 97.72 & \textbf{95.46} & 95.58 & 91.15 & 91.11 & 89.86 & 83.06 \\
 & \textbf{VIGS-SLAM (Ours)} & \textbf{100.00} & \textbf{100.00} & \textbf{100.00} & 92.87 & \textbf{100.00} & \textbf{100.00} & \textbf{100.00} & 88.02 & \textbf{97.61} \\
\hdashline

\multirow{5}{*}{Recall @ 10cm [\%] $\uparrow$}
 & HI-SLAM2~\cite{zhang2024hislam2} & 98.63 & 54.57 & 99.37 & 96.46 & \textbf{100.00} & \textbf{100.00} & \textbf{100.00} & \textbf{100.00} & 93.63 \\
 & VINS-Mono~\cite{qin2018vins} & 98.32 & 97.39 & 96.60 & 94.94 & 99.53 & 98.93 & 98.99 & 99.48 & 98.02 \\
 & OPEN-VINS~\cite{geneva2020openvins} & 99.63 & 99.76 & 99.91 & 96.21 & 99.74 & 99.49 & 95.09 & 93.54 & 97.92 \\
 & ORB-SLAM3~\cite{campos2021orb3} & 13.03 & \textbf{100.00} & 98.37 & 96.58 & \textbf{100.00} & 99.95 & 96.74 & 96.48 & 87.64 \\
 & \textbf{VIGS-SLAM (Ours)} & \textbf{100.00} & \textbf{100.00} & \textbf{100.00} & \textbf{100.00} & \textbf{100.00} & \textbf{100.00} & \textbf{100.00} & \textbf{100.00} & \textbf{100.00} \\
\midrule

\multicolumn{11}{l}{\cellcolor[HTML]{EEEEEE}\textit{Stride = 10}} \\

\multirow{5}{*}{ATE RMSE [cm] $\downarrow$}
 & HI-SLAM2~\cite{zhang2024hislam2} & 11.78 & 76.90 & 45.69 & 8.65 & 1.56 & 2.07 & 1.04 & F & N/A \\
 & VINS-Mono~\cite{qin2018vins} & F & 14.53 & 5.30 & 6.06 & 2.94 & 11.33 & 1.34 & 21.81 & N/A \\
 & OPEN-VINS~\cite{geneva2020openvins} & 4.22 & 120.15 & 5.04 & 8.11 & 4.03 & 178.25 & 3.22 & 8.32 & 41.42 \\
 & ORB-SLAM3~\cite{campos2021orb3} & 1.57 & \textbf{0.20} & 1.19 & \textbf{0.93} & 1.32 & \textbf{1.04} & \textbf{0.94} & \textbf{2.79} & \textbf{1.25} \\
 & \textbf{VIGS-SLAM (Ours)} & \textbf{1.20} & 1.51 & \textbf{1.18} & 1.62 & \textbf{1.29} & 1.32 & 1.04 & 3.24 & 1.55 \\
\hdashline

\multirow{5}{*}{Recall @ 5cm [\%] $\uparrow$}
 & HI-SLAM2~\cite{zhang2024hislam2} & 41.12 & 5.58 & 13.29 & 84.84 & 99.66 & 98.98 & \textbf{100.00} & 0.00 & 55.43 \\
 & VINS-Mono~\cite{qin2018vins} & 0.00 & 17.37 & 60.68 & 48.59 & 72.17 & 26.00 & 69.32 & 18.15 & 39.04 \\
 & OPEN-VINS~\cite{geneva2020openvins} & 40.64 & 0.00 & 64.62 & 30.07 & 71.14 & 0.00 & 65.91 & 29.44 & 37.73 \\
 & ORB-SLAM3~\cite{campos2021orb3} & 13.97 & 4.55 & 70.24 & 62.94 & 78.09 & 21.09 & 72.21 & 61.11 & 48.03 \\
 & \textbf{VIGS-SLAM (Ours)} & \textbf{100.00} & \textbf{100.00} & \textbf{100.00} & \textbf{97.91} & \textbf{100.00} & \textbf{100.00} & \textbf{100.00} & \textbf{93.75} & \textbf{98.96} \\
\hdashline

\multirow{5}{*}{Recall @ 10cm [\%] $\uparrow$}
 & HI-SLAM2~\cite{zhang2024hislam2} & 89.72 & 32.27 & 50.90 & 95.74 & \textbf{100.00} & \textbf{100.00} & \textbf{100.00} & 0.00 & 67.36 \\
 & VINS-Mono~\cite{qin2018vins} & 0.00 & 63.63 & 94.67 & 96.15 & 93.34 & 80.68 & 97.78 & 58.43 & 73.09 \\
 & OPEN-VINS~\cite{geneva2020openvins} & 86.59 & 0.46 & 99.14 & 87.43 & 98.35 & 0.00 & 98.93 & 82.81 & 69.21 \\
 & ORB-SLAM3~\cite{campos2021orb3} & 15.75 & 10.72 & 94.25 & 89.13 & 98.01 & 49.72 & 99.94 & 93.48 & 68.88 \\
 & \textbf{VIGS-SLAM (Ours)} & \textbf{100.00} & \textbf{100.00} & \textbf{100.00} & \textbf{100.00} & \textbf{100.00} & \textbf{100.00} & \textbf{100.00} & \textbf{100.00} & \textbf{100.00} \\
\midrule

\multicolumn{11}{l}{\cellcolor[HTML]{EEEEEE}\textit{Stride = 20}} \\

\multirow{5}{*}{ATE RMSE [cm] $\downarrow$}
 & HI-SLAM2~\cite{zhang2024hislam2} & 28.59 & 87.66 & 114.75 & 29.30 & 5.57 & 4.56 & \textbf{1.01} & 127.72 & 49.90 \\
 & VINS-Mono~\cite{qin2018vins} & F & F & F & F & F & F & 14.78 & 147.92 & N/A \\
 & OPEN-VINS~\cite{geneva2020openvins} & F & F & F & F & F & F & F & F & N/A \\
 & ORB-SLAM3~\cite{campos2021orb3} & F & \textbf{0.53} & F & \textbf{0.28} & F & \textbf{0.23} & 1.03 & \textbf{3.15} & N/A \\
 & \textbf{VIGS-SLAM (Ours)} & \textbf{8.96} & 123.66 & \textbf{0.95} & 2.36 & \textbf{1.26} & 1.55 & 1.18 & 160.08 & \textbf{37.50} \\
\hdashline

\multirow{5}{*}{Recall @ 5cm [\%] $\uparrow$}
 & HI-SLAM2~\cite{zhang2024hislam2} & \textbf{8.41} & \textbf{4.07} & 7.19 & 13.21 & 98.44 & 91.80 & \textbf{100.00} & 0.00 & 40.39 \\
 & VINS-Mono~\cite{qin2018vins} & 0.00 & 0.00 & 0.00 & 0.00 & 0.00 & 0.00 & 11.35 & 0.34 & 1.46 \\
 & OPEN-VINS~\cite{geneva2020openvins} & 0.00 & 0.00 & 0.00 & 0.00 & 0.00 & 0.00 & 0.00 & 0.00 & 0.00 \\
 & ORB-SLAM3~\cite{campos2021orb3} & 0.00 & 3.82 & 0.00 & 1.59 & 0.00 & 4.31 & 38.19 & \textbf{21.73} & 8.71 \\
 & \textbf{VIGS-SLAM (Ours)} & 1.77 & 0.00 & \textbf{100.00} & \textbf{99.63} & \textbf{100.00} & \textbf{100.00} & \textbf{100.00} & 0.00 & \textbf{62.68} \\
\hdashline

\multirow{5}{*}{Recall @ 10cm [\%] $\uparrow$}
 & HI-SLAM2~\cite{zhang2024hislam2} & 34.58 & \textbf{11.38} & 4.63 & 50.94 & 99.61 & 96.72 & \textbf{100.00} & 12.94 & 51.35 \\
 & VINS-Mono~\cite{qin2018vins} & 0.00 & 0.00 & 0.00 & 0.00 & 0.00 & 0.00 & 42.93 & 0.81 & 5.47 \\
 & OPEN-VINS~\cite{geneva2020openvins} & 0.00 & 0.00 & 0.00 & 0.00 & 0.00 & 0.00 & 0.00 & 0.00 & 0.00 \\
 & ORB-SLAM3~\cite{campos2021orb3} & 0.00 & 10.35 & 0.00 & 2.05 & 0.00 & 11.31 & 80.63 & \textbf{53.37} & 19.71 \\
 & \textbf{VIGS-SLAM (Ours)} & \textbf{93.81} & 0.86 & \textbf{100.00} & \textbf{100.00} & \textbf{100.00} & \textbf{100.00} & \textbf{100.00} & 0.32 & \textbf{74.37} \\
\midrule

\multicolumn{11}{l}{\cellcolor[HTML]{EEEEEE}\textit{Stride = 40}} \\

\multirow{5}{*}{ATE RMSE [cm] $\downarrow$}
 & HI-SLAM2~\cite{zhang2024hislam2} & \textbf{158.63} & \textbf{120.37} & \textbf{145.92} & \textbf{145.39} & 22.82 & 172.33 & 52.77 & \textbf{160.83} & 122.38 \\
 & VINS-Mono~\cite{qin2018vins} & F & F & F & F & F & F & F & F & N/A \\
 & OPEN-VINS~\cite{geneva2020openvins} & F & F & F & F & F & F & F & F & N/A \\
 & ORB-SLAM3~\cite{campos2021orb3} & F & F & F & F & F & F & F & F & N/A \\
 & \textbf{VIGS-SLAM (Ours)} & 162.20 & 122.94 & 147.07 & 148.91 & \textbf{7.20} & \textbf{166.17} & \textbf{1.05} & 162.41 & \textbf{114.74} \\
\hdashline

\multirow{5}{*}{Recall @ 5cm [\%] $\uparrow$}
 & HI-SLAM2~\cite{zhang2024hislam2} & 0.00 & \textbf{1.59} & 0.00 & 0.00 & 46.21 & 0.00 & 4.46 & 0.00 & 6.53 \\
 & VINS-Mono~\cite{qin2018vins} & 0.00 & 0.00 & 0.00 & 0.00 & 0.00 & 0.00 & 0.00 & 0.00 & 0.00 \\
 & OPEN-VINS~\cite{geneva2020openvins} & 0.00 & 0.00 & 0.00 & 0.00 & 0.00 & 0.00 & 0.00 & 0.00 & 0.00 \\
 & ORB-SLAM3~\cite{campos2021orb3} & 0.00 & 0.00 & 0.00 & 0.00 & 0.00 & 0.00 & 0.00 & 0.00 & 0.00 \\
 & \textbf{VIGS-SLAM (Ours)} & 0.00 & 0.0 & 0.0 & 0.0 & \textbf{62.07} & \textbf{1.56} & \textbf{100.00} & 0.00 & \textbf{20.45} \\
\hdashline

\multirow{5}{*}{Recall @ 10cm [\%] $\uparrow$}
 & HI-SLAM2~\cite{zhang2024hislam2} & 0.00 & \textbf{1.59} & \textbf{0.60} & 0.00 & 64.83 & 0.00 & 16.07 & \textbf{0.99} & 10.51 \\
 & VINS-Mono~\cite{qin2018vins} & 0.00 & 0.00 & 0.00 & 0.00 & 0.00 & 0.00 & 0.00 & 0.00 & 0.00 \\
 & OPEN-VINS~\cite{geneva2020openvins} & 0.00 & 0.00 & 0.00 & 0.00 & 0.00 & 0.00 & 0.00 & 0.00 & 0.00 \\
 & ORB-SLAM3~\cite{campos2021orb3} & 0.00 & 0.00 & 0.00 & 0.00 & 0.00 & 0.00 & 0.00 & 0.00 & 0.00 \\
 & \textbf{VIGS-SLAM (Ours)} & 0.00 & 1.49 & 0.00 & 0.00 & \textbf{91.03} & \textbf{1.56} & \textbf{100.00} & 0.00 & \textbf{24.26} \\
\bottomrule

\end{tabular}}

\label{tab:tracking_rpng_stride}
\end{table*}

\bibliographystyle{splncs04}
\bibliography{main}
\end{document}